\documentclass[final,3p,times,twocolumn,authoryear]{elsarticle}

\usepackage{amssymb}
\usepackage{amsmath}
\usepackage{multirow}
\usepackage{booktabs}
\usepackage{graphicx}
\usepackage{subcaption}
\usepackage{hyperref}
\usepackage{rotating} 

\journal{Photos}

\begin{document}

\begin{frontmatter}



\title{EuroMineNet: A Multitemporal Sentinel-2 Benchmark for Spatiotemporal Mining Footprint Analysis in the European Union (2015–2024)} 


\author[tum,hzdr]{Weikang Yu} 
\author[hzdr]{Vincent Nwazelibe}
\author[swjt]{Xianping Ma}
\author[whu]{Xiaokang Zhang\corref{cor1}}
\author[hzdr]{Richard Gloaguen}
\author[tum]{Xiao Xiang Zhu}
\author[hzdr,lancaster]{Pedram Ghamisi}

\cortext[cor1]{Corresponding Author}

\affiliation[tum]{organization={Technical University of Munich},
            addressline={}, 
            city={Munich},
            postcode={80331}, 
            state={},
            country={Germany}}
\affiliation[hzdr]{organization={Helmholtz-Zentrum Dresden-Rossendorf},
            addressline={}, 
            city={Freiberg},
            postcode={09599}, 
            state={},
            country={Germany}}
\affiliation[swjt]{organization={Southwest Jiaotong University},
            addressline={}, 
            city={Chengdu},
            postcode={611756}, 
            state={},
            country={China}}
\affiliation[whu]{organization={Wuhan University},
            addressline={}, 
            city={Wuhan},
            postcode={430072}, 
            state={},
            country={China}}
\affiliation[lancaster]{organization={Lancaster University},
            addressline={}, 
            city={Lancaster},
            postcode={LA1 4YW}, 
            state={},
            country={United Kingdom}}

\begin{abstract}
Mining activities are essential for industrial and economic development but remain a leading source of environmental degradation, contributing to deforestation, soil erosion, and water contamination. Sustainable resource management and environmental governance require consistent, long-term monitoring of mining-induced land surface changes, yet existing datasets are often limited in temporal depth or geographic scope. To address this gap, we present EuroMineNet, the first comprehensive multitemporal benchmark for mining footprint mapping and monitoring based on Sentinel-2 multispectral imagery. Spanning 133 mining sites across the European Union, EuroMineNet provides annual observations and expert-verified annotations from 2015 to 2024, enabling GeoAI-based models to analyze environmental dynamics at continental scale. It supports two sustainability-driven tasks: (1) multitemporal mining footprint mapping for consistent annual land-use delineation, evaluated with a novel Change-Aware Temporal IoU (CA-TIoU) metric, and (2) cross-temporal change detection to capture both gradual and abrupt surface transformations. Benchmarking 20 state-of-the-art deep learning models reveals that while GeoAI methods effectively identify long-term environmental changes, challenges remain in detecting short-term dynamics critical for timely mitigation. By advancing temporally consistent and explainable mining monitoring, EuroMineNet contributes to sustainable land-use management, environmental resilience, and the broader goal of applying GeoAI for social and environmental good. We release the codes and datasets by aligning with FAIR and the open science paradigm at \href{https://github.com/EricYu97/EuroMineNet}{https://github.com/EricYu97/EuroMineNet}.

\end{abstract}


\end{frontmatter}



\section{Introduction}
\label{sec1}
Mining plays a critical role in supporting industrial development and the energy transition, yet is also among the most significant drivers of land surface transformation \citep{lebre2020social, giljum2025metal}. Mining presents a paradoxical situation for the energy transition. While it provides essential minerals such as lithium, cobalt, and copper, which are crucial for renewable technologies and batteries, it often results in substantial environmental degradation \citep{sengupta2021environmental}, including deforestation \citep{sonter2017mining}, water pollution \citep{liu2021review}, and habitat destruction \citep{siqueira2020exploring}. Furthermore, mining itself is energy-intensive and frequently relies on fossil fuels \citep{azadi2020transparency}, thereby undermining the very emissions reductions it seeks to support. This creates a conundrum: extracting resources to construct a sustainable future can inadvertently perpetuate ecological harm and social conflicts. Negative impacts are particularly pronounced in regions with high resource demand and limited environmental oversight. In this context, systematic and timely monitoring of mining activities has become increasingly important, which is not only necessary to ensure regulatory compliance and sustainable resource management but also to understand the broader environmental implications of mineral extraction. 

Remote sensing, with its synoptic view, repeatability, and scalability, offers a powerful means to observe and analyze the spatial and temporal dynamics of mining across large areas \citep{yu2018monitoring}. Over the last decade, rapid advancements in remote sensing techniques and the increasing availability of Earth observation data, characterized by improved spatial, spectral, and temporal resolution, have greatly enhanced our ability to monitor environmental changes at multiple scales \citep{zhang2012application, ghamisi2021potential, ghamisi2025responsible}. These developments have facilitated a wide range of mining-related applications, including delineation of mining extents \citep{werner2020global}, detection of land cover change \citep{sonter2014processes}, assessment of environmental impacts \citep{charou2010using}, and enforcement of land use regulations \citep{dube2024assessment}. Multispectral imagery has proven especially effective for capturing the spectral characteristics of disturbed surfaces \citep{cohen2018landtrendr, yang2018identification}, while time-series analysis and machine learning approaches have enhanced the ability to identify subtle or progressive changes over time \citep{fu2024remote}. However, the progress of mining monitoring research remains constrained by the lack of standardized, multitemporal benchmarks that allow for robust model development and evaluation.

Despite recent progress, current mining monitoring studies face several key limitations that restrict their effectiveness for comprehensive, long-term analysis. First, large-scale mapping efforts have demonstrated the feasibility of identifying mining footprints across hundreds of sites globally \citep{yu2024minenetcd}, but these are typically based on bitemporal observations, which fail to capture the continuous and often subtle evolution of mining activities over time. Second, existing studies tend to focus either on globally distributed but sparsely sampled sites \citep{saputra2025multi} or on small-scale regional areas \citep{xie2025integrated}, limiting their ability to balance spatial coverage with temporal depth and contextual consistency. Third, although some recent research has explored multiclass mapping of mining-related features \citep{saputra2025multi}, these datasets are often limited to single-date imagery or small study areas, which restricts their usefulness for temporal change analysis. These limitations highlight the urgent need for a benchmark that combines dense multitemporal coverage with a well-defined and manageable geographic scope, supporting the development and evaluation of models for both mining footprint mapping and long-term change detection.

To address the existing research gaps, we introduce EuroMineNet, a comprehensive multitemporal benchmark for mining footprint mapping and continuous monitoring of mining activity dynamics across the European Union. The EU presents a strategically important, environmentally diverse, yet geographically coherent region that is well-suited for focused, cross-country mining monitoring studies. EuroMineNet leverages Sentinel-2 multispectral imagery spanning a full decade (2015–2024), made possible by the Sentinel-2 mission’s launch in 2015, which has provided consistent, high-resolution, and freely accessible Earth observation data since then. With yearly annotations for each mining site, the benchmark enables dynamic tracking of mining footprint changes on an annual basis. With this unique decade-long data record, EuroMineNet enables the development and benchmarking of Earth observation methods capable of capturing the evolving nature of mining activities. This continuous temporal resolution, combined with the focused spatial scale, makes EuroMineNet an ideal resource for advancing robust and interpretable mining monitoring approaches at regional and continental scales.

The contributions of this paper are organized as follows:
\begin{itemize}
    \item We propose \textbf{EuroMineNet}, the first multitemporal mining footprint mapping and monitoring dataset based on multispectral remote sensing data. EuroMineNet consists of 51330 image patches that cover 133 mining sites across the European Union (EU), providing per-year Sentinel-2 observations and accurate mining footprint annotations over the past decade, enabling both static and dynamic monitoring of mining activities.
    
    \item We formalize \textbf{multitemporal mining footprint mapping}, aiming to generate consistent, year-by-year binary maps (mine vs. non-mine) from a decade of multispectral data for both long-term trend and short-term variation analysis. We further propose two Change-Aware Temporal IoU (CA-TIoU) metrics to assess temporal consistency while accounting for actual land cover changes, promoting stable yet change-sensitive footprint mapping.
    
    \item We define the task of \textbf{cross-temporal change detection}, which targets the identification of mining-induced changes at arbitrary temporal intervals, ranging from short-term to long-term. Through extensive evaluation, we highlight the challenge that existing change detection models struggle to consistently detect dynamic short-term changes while maintaining accuracy across different temporal scales.
\end{itemize}

\section{Related Work}
\subsection{Remote Sensing for Mining Footprint Monitoring and Analysis}
Remote sensing has become a fundamental tool for monitoring mining activities due to its capability to provide consistent and large-scale observations \citep{tang2023global, maus2020global,maus2022update}. Mining operations often cause significant landscape alterations, including land clearance, soil disruption, and waste deposition \citep{sengupta2021environmental, jain2015environmental}, which are detectable through spectral, spatial, and textural features in satellite imagery \citep{charou2010using, zhang2012application, padmanaban2017remote, firozjaei2021historical}. In the literature, remote sensing data have been widely used to generate land-use and land-cover (LULC) maps or compute environmental spectral indices \cite{wang2023rseife, wang2024mapping}, which serve as indicators of ecological conditions and whose variations are analyzed to assess environmental impacts \citep{dehkordi2024soil}.

Traditional approaches relied on visual interpretation of Earth observation data and threshold-based spectral indices.
For example, \cite{sun2024impacts} extracted phenological indices from Sentinel-2-based vegetation index time series and quantified mining impacts by analyzing changes in phenological differences across spatial gradients, applying this method to the Bainaimiao copper mining footprint in Inner Mongolia, China. 
\cite{firozjaei2021historical} developed a homogeneity distance classification algorithm to evaluate the historical impacts of mining activities on surface biophysical characteristics, and also applied the CA-Markov model to predict the future changes in the pattern of vegetation cover and land surface temperature. 
\cite{zhang2023assessing} quantifies the vegetation restoration process of dumping sites in mining footprints by analyzing the spatio-temporal change of the Fractional Vegetation Cover (FVC) based on Normalized Difference Vegetation Index (NDVI) and Digital Elevation Model (DEM) data derived from remote sensing imagery. 
However, these approaches not only require professional domain knowledge to interpret the data but also require a manual process of such data, which not only requires extensive efforts in data analysis but also limits them to only the scope of case studies with a regional scale. 

The advent of higher-resolution optical sensors, big data, and advances in image analysis has driven the adoption of deep learning and object-based methods for mining monitoring. These methods enable accurate pixel-level LULC mapping from remote sensing imagery \citep{xie2020semantic, kumar2023development, li2025machine, saputra2025multi, chen2022open} and facilitate the detection of spatiotemporal dynamics from multitemporal Earth observation data \citep{yu2024minenetcd, camalan2022change, jablonska2024minecam, li2022change}. LULC mapping approaches focus on distinguishing classes relevant to environmental impacts, such as waste disposal and water bodies. For example, \cite{saputra2025multi} applied four deep learning-based segmentation models to map mining and non-mining land cover across 37 global mining sites using multispectral imagery. Meanwhile, change detection methods aim to capture mining-induced changes over time. For instance, \cite{yu2024minenetcd} introduced a global mining change detection benchmark and a fast Fourier transform-based change detection algorithm to capture the mining activities from bitemporal optical imagery across 100 mining sites worldwide.

Despite these advances, existing methods remain limited to single-temporal mapping or bitemporal change detection, overlooking the continuous and rapid nature of mining processes \citep{zhang2021continuous}. Furthermore, the recent availability of rich multispectral data, enabled by missions like Sentinel-2, has not been fully exploited for long-term mining monitoring. As a result, modeling dynamic mining activities over extended periods remains a significant challenge. In addition, while previous studies typically focus on either small-scale mapping in a few case studies \citep{wang2024mapping} or large-scale mapping in sparsely distributed global sites \citep{saputra2025multi, yu2024minenetcd}, they rarely address a comprehensive assessment within a union- or country-level region under a unified administrative and regulatory framework, which would be highly beneficial for policy-making, compliance monitoring, and sustainable resource management.


\subsection{Change Detection in Earth Observation}

The increasing availability of dense satellite image time series has significantly advanced spatiotemporal analysis for monitoring gradual and abrupt land-use and land-cover (LULC) changes. Change detection in remote sensing aims to automate this process by generating pixel-level change maps from bitemporal or multitemporal imagery \citep{peng2025deep, cheng2024change, wu2024unet}. These techniques have been widely applied in diverse domains such as natural disaster assessment \citep{zhang2023cross, saleh2024dam}, forestry \citep{pelletier2024inter}, agriculture \citep{sun2024identifying}, and urban expansion \citep{chen2022egde, ning2024multi}.

Traditional change detection methods, including post-classification comparison \citep{wu2017post}, image differencing \citep{bruzzone2002automatic}, and change vector analysis \citep{hu2018automatic, he2011detecting}, rely on handcrafted features and threshold-based decision rules. While effective for small-scale case studies, these methods suffer from limited generalization capability and often require substantial manual intervention, making them unsuitable for large-scale or long-term monitoring. In the past decade, the emergence of artificial intelligence (AI) has revolutionized change detection, shifting towards deep learning-based approaches. Most state-of-the-art methods adopt an encoder-decoder architecture, often leveraging Siamese networks to extract spatiotemporal features from bitemporal inputs, which are then fused to produce pixel-level change maps. These models demonstrate strong generalization across different sensors and regions while maintaining high accuracy. Among them, UNet and its variants are the most widely used \citep{wu2024unet, daudt2018fully}. Many of these UNet-based methods introduce feature fusion mechanisms through skip connections to improve spatiotemporal representation learning \citep{pan2023new, li2022transunetcd}.

Recent advances in self-attention have further driven the development of transformer-based change detection models, which capture long-range dependencies in spatial and temporal domains. For example, \cite{chen2021remote} introduced the Bitemporal Image Transformer (BIT), which represents change information using a compact set of semantic tokens for efficient context modeling. Similarly, \cite{zheng2022changemask} proposed ChangeMask, a multi-task encoder–transformer–decoder network that incorporates semantic-change relationships and temporal symmetry as inductive biases. \cite{yu2024maskcd} extended this paradigm by employing a detection transformer (DETR)-based decoder to generate category-aware change masks, improving localization accuracy and robustness.

Despite these advances, most studies remain focused on bitemporal change detection, which limits their ability to capture continuous and dynamic land-cover transitions over extended periods. In contrast, this work addresses the challenge of multitemporal change detection, introducing a new challenging scenario of cross-temporal analysis, where models must detect changes across multiple time intervals, leveraging long-term Earth observation data for dynamic monitoring.

\subsection{Remote Sensing Benchmark Datasets for Spatiotemporal Monitoring}
Benchmark datasets play a critical role in advancing remote sensing research by providing standardized evaluation protocols and facilitating the development of robust methods. With the growing availability of Earth observation (EO) data, numerous benchmarks have been introduced for spatiotemporal monitoring of land-cover changes in domains such as urban development, agriculture, and mining. However, existing change detection benchmarks exhibit two major limitations.

First, most datasets rely exclusively on optical imagery \citep{peng2025deep}, overlooking the rich spectral information provided by multispectral sensors, which is crucial for detecting subtle LULC variations and monitoring environmental indicators. For example, the Sentinel-2 Multitemporal Cities Pairs (S2MTCP) dataset \citep{leenstra2021self} and the Onera Satellite Change Detection (OSCD) dataset \citep{daudt2018urban} utilize Sentinel-2 multispectral imagery to capture urbanization-related changes. However, their spatial coverage and temporal diversity remain limited, restricting their use for training generalized models that perform well across large-scale or heterogeneous regions.

Second, the majority of benchmarks focus on bitemporal change detection, involving only two images captured at different times \citep{peng2025deep}. This design fails to represent real-world scenarios where short-term and incremental changes need to be captured for timely decision-making. In practice, anthropogenic changes, such as mining expansion or urban sprawl, occur progressively rather than instantaneously. Continuous monitoring with multiple time points offers a more comprehensive understanding of spatiotemporal dynamics, providing critical insights into how environmental impacts evolve over time.

These limitations are particularly evident in the context of mining footprint monitoring. Existing mining-related studies typically rely on two time points to assess changes, yielding only coarse insights into the evolution of mining activities. The absence of benchmarks with dense temporal coverage and geographically coherent regions significantly hinders the development and evaluation of robust models for dynamic mining monitoring. To bridge this gap, future benchmarks must combine temporal depth with spatial consistency, enabling both static and dynamic change detection approaches. Such datasets would allow for the implementation of continuous monitoring strategies, leveraging annual or intra-annual observations to capture the progressive nature of mining operations and their environmental impacts.

\section{Methodology}
\subsection{Study Area}
This study focuses on the mining developments in the European Union (EU). The EU represents a politically and economically integrated region comprising 27 member states, characterized by diverse climatic zones, topography, and land-use patterns. As a major global economic bloc, the EU is a significant consumer and producer of mineral resources, with mining activities concentrated in countries such as Poland, Germany, Sweden, Spain, and Finland. Mining in Europe continues to be of paramount importance in securing strategic commodities such as lithium, cobalt, copper, and rare earth elements. These elements are indispensable for the development of renewable energy technologies, electric vehicles, and digital industries. Although Europe’s production capacity is relatively limited compared to global leaders, domestic mining mitigates the dependency on imports, particularly from geopolitically sensitive regions. This approach aligns with the European Union’s objective of enhancing supply chain resilience and sustainability. However, environmental regulations and social acceptance present significant challenges, influencing Europe’s strategy for harmonizing resource extraction with its green transition aspirations. The EU has implemented stringent environmental policies under frameworks like the European Green Deal and the Raw Materials Initiative to ensure sustainable resource extraction and land rehabilitation. These regulatory measures, combined with the heterogeneous distribution of mining sites across various ecosystems—from boreal forests in the north to Mediterranean landscapes in the south—make the EU an ideal study area for assessing the spatiotemporal dynamics of mining and its environmental impacts under unified administrative and policy frameworks.

\begin{figure*}[!htbp]
    \centering
    \includegraphics[width=\linewidth]{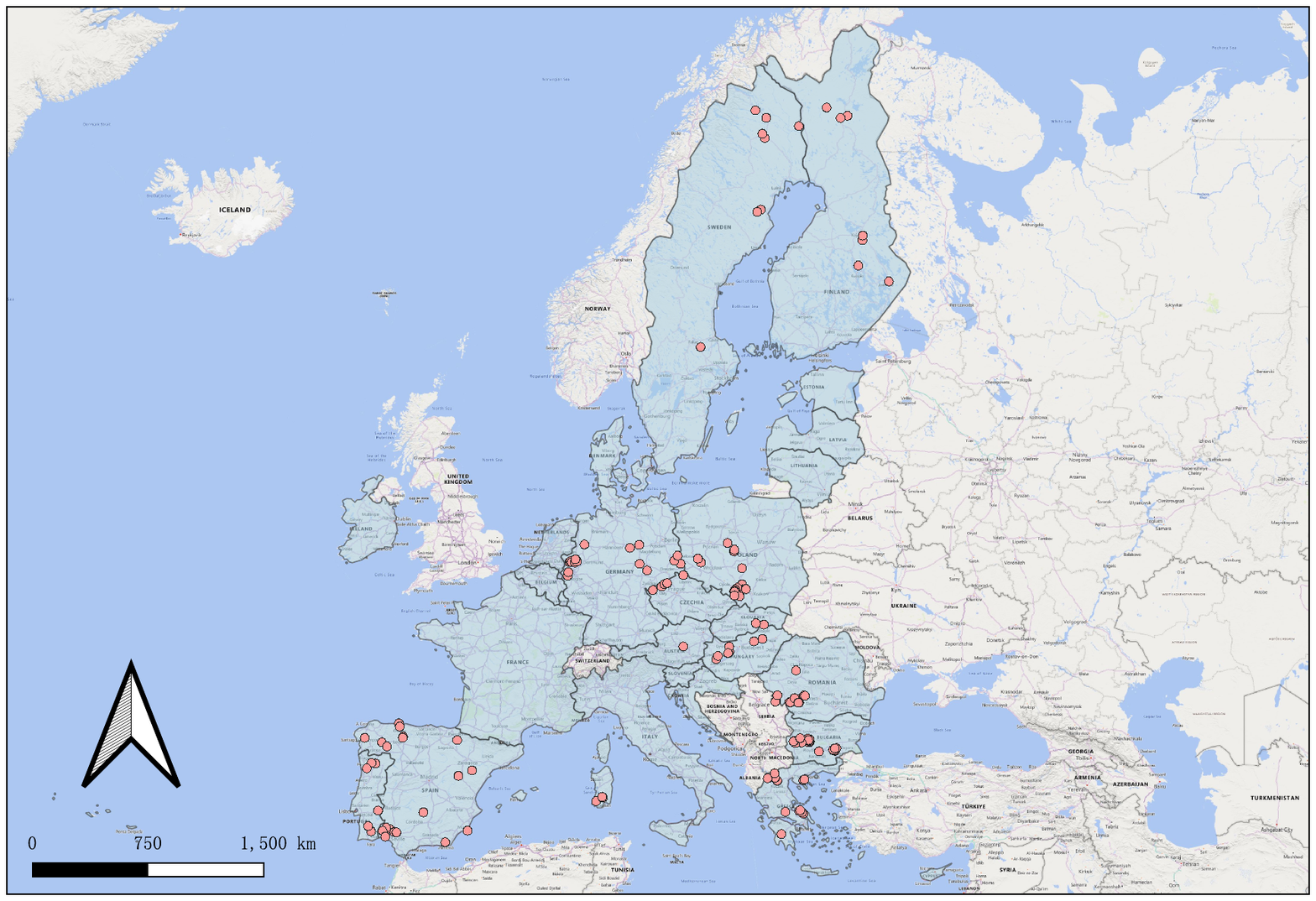}

    \caption{Geospatial distribution of the 133 mining sites investigated in this study. The study area of the European Union is highlighted.}
    \label{fig:mining_sites_map}
\end{figure*}

From the mining sites recorded in the global-scale dataset of mining footprints \citep{maus2020global}, we selected 133 of the most representative mining sites from 14 countries across the EU region, as shown in the Table. \ref{tab:minelist} and Fig.\ref{fig:mining_sites_map}. These mining sites vary from different commodities, which can be roughly categorized into four production types: metallic mine extract metal-bearing ores such as iron, copper, and gold; non-metallic mine extract non-metallic materials used in construction, agriculture, or industry, such as potash and quartz. Open-pit coal mines extract mainly lignite for use as a fuel and industrial material; quarries extract bulk materials used in construction, such as road aggregates. Overall, the 133 mining sites selected consist of 50 metallic, 56 coal, 8 non-metallic, and 19 large quarries, and the proportion of such mines across different countries varies. For all the mining sites investigated, we mapped a total area of $11324.5 \mathrm{km}^{2}$.
\begin{table*}[h]
\centering
\caption{Listed countries and mining sites in the study area (i.e., EU). The percentage of the changed area of the mining sites from each country over the last decade is demonstrated after the data on the covered area in 2024.}
\label{tab:minelist}
\resizebox{\linewidth}{!}{%
\begin{tabular}{l|cccc}
\toprule
EU Country & Number of Sites & Production Type (Amount) & Covered Area in 2015 ($km^{2}$)& Covered Area in 2024 ($km^{2}$) \\ \midrule
Austria     & 1  & Metallic (1)                                      & 9.28   & 9.69 (+4.42\%)   \\
Bulgaria    & 18 & Metallic (8), Coal (8), Quarry (2)               & 180.53 & 202.50 (+12.17\%) \\
Czechia     & 4  & Metallic (1), Coal (3)                           & 186.08 & 192.46 (+3.43\%)  \\
Finland     & 7  & Metallic (4), Non-metallic (2), Quarry (1)      & 106.77 & 151.27 (+41.63\%) \\
Germany     & 16 & Metallic (1), Non-metallic (2), Coal (11), Quarry (2) & 588.44 & 632.49 (+7.49\%)  \\
Greece      & 13 & Metallic (9), Coal (4)                           & 213.68 & 203.15 (-4.92\%)  \\
Hungary     & 7  & Coal (4), Non-metallic (1), Quarry (2)          & 37.13  & 41.52 (+11.81\%) \\
Italy       & 3  & Metallic (2), Quarry (1)                         & 5.16   & 4.26 (-17.44\%)  \\
Poland      & 17 & Metallic (2), Coal (12), Quarry (3)             & 255.03 & 274.63 (+7.67\%)  \\
Portugal    & 6  & Metallic (2), Quarry (4)                         & 21.43  & 22.14 (+3.31\%)   \\
Romania     & 12 & Metallic (3), Coal (9)                           & 92.09  & 79.19 (-14.00\%) \\
Slovakia    & 2  & Metallic (1), Quarry (1)                         & 3.14   & 2.89 (-7.96\%)   \\
Spain       & 19 & Metallic (8), Coal (5), Non-metallic (3), Quarry (3) & 160.43 & 168.09 (+4.77\%)  \\
Sweden      & 8  & Metallic (8)                                     & 229.68 & 263.38 (+14.66\%) \\ \midrule
Total       & 133 & Metallic (50), Coal (56), Non-metallic (8), Quarry (19) &    2088.89     &  2247.65 (+7.60\%)       \\
\bottomrule
\end{tabular}}
\end{table*}

\subsection{Data Acquisition and Annotation}
We utilized Sentinel-2 satellite imagery to acquire Earth observation data for mining sites across the European Union. Sentinel-2 offers high-resolution multispectral data with a revisit frequency of 5 days, providing consistent and dense temporal coverage ideal for monitoring land surface changes. We selected Level-2A surface reflectance products, which include atmospheric correction and are suitable for downstream analysis. For each mining site, we collected a multitemporal image sequence spanning multiple years, prioritizing cloud-free observations during peak vegetation seasons to ensure reliable interpretation of surface features.

To ensure consistent spatial resolution across spectral bands, we selected 10 bands from Sentinel-2, excluding the three 60m/pixel bands (B1, B9, and B10) due to their low spatial detail. All remaining bands, including those originally at 20m/pixel resolution, were resampled to 10m/pixel to enable unified processing and analysis.

The four main areas of a mine are:
\begin{itemize}
    \item Mine Site (Pit): The core area for ore extraction, including pits and stopes.
    \item Processing Plant: Facilities for crushing, grinding, and processing ore to separate valuable minerals.
    \item Waste Management Areas: Tailings storage and waste rock dumps to contain byproducts and minimize environmental impact.
    \item Infrastructure and Support: Access roads, power, water, administrative buildings, and worker accommodations. 
\end{itemize}
To construct pixel-level annotations, we manually delineated mining-related land cover classes, such as active extraction zones, waste deposits, and tailings ponds, using high-resolution reference imagery and visual interpretation techniques. Change annotations were generated by comparing temporal snapshots and marking regions exhibiting mining-induced transformations, such as expansion of extraction areas or development of new infrastructure. All annotations were validated by cross-referencing auxiliary data sources, including OpenStreetMap, national mining inventories, and industry reports, to ensure consistency and accuracy. This comprehensive dataset enables robust training and evaluation of spatiotemporal models for monitoring mining dynamics.

\begin{table*}[]
\centering
\caption{Statistics for mining footprint evolution in the study area from 2015 to 2024. The current area, expanded area, decreased area, and total changes are calculated on all the mining sites in this study and demonstrated in the unit of $km^{2}$.}
\label{tab:subsequent_years}
\small
\resizebox{\linewidth}{!}{%
\begin{tabular}{l|cccccccccc}
\toprule
\multirow{2}{*}{Statistics ($km^{2}$)} & \multicolumn{10}{c}{Years} \\
 & 2015 & 2016 & 2017 & 2018 & 2019 & 2020 & 2021 & 2022 & 2023 & 2024 \\ \midrule
Current Area & 2088.89 & 2122.90 & 2139.36 & 2165.45 & 2186.11 & 2207.60 & 2210.13 & 2212.78 & 2219.96 & 2247.65  \\
Percentage of Mining Footprint &18.45\%	&18.75\%	&18.89\%	&19.12\%	&19.30\%	&19.49\%	&19.52\%	&19.54\%	&19.60\%	&19.85\%
 \\ \midrule
Expanded Area & - & 69.43&	69.35&	67.49&	69.62&	59.66&	44.79&	39.48&	52.86&	51.40  \\
 (Percentage) & - & (66.2\%)	&(56.7\%)	&(62.0\%)	&(58.7\%)	&(61.0\%)	&(51.5\%)	&(51.7\%)	&(53.6\%)	&(68.4\%) \\
Decreased Area & - & 35.42&	52.89&	41.39&	48.97&	38.17&	42.25&	36.83&	45.67&	23.71 \\
 (Percentage) & - & (33.8\%)	& (43.3\%)	& (38.0\%)	& (41.3\%)	& (39.0\%)	& (48.5\%)	& (48.3\%)	& (46.4\%)	& (31.6\%) \\
Total Changes & - & 104.84&	122.25&	108.88&	118.59&	97.83&	87.04&	76.31&	98.53&	75.11  \\ \bottomrule
\end{tabular}%
}
\end{table*}

\begin{table*}[]
\centering
\caption{Statistics for mining footprint changes in the study area from 2015 to 2024, with regard to different interval years. The  expanded area, decreased area, and accumulated changes are calculated on all the mining sites in this study. The percentage of expanded area and decreased area for each year is demonstrated in the brackets below the area ($km^{2}$) data.}
\label{tab:year-interval}
\resizebox{\linewidth}{!}{%
\begin{tabular}{l|ccccccccc}
\toprule
\multirow{2}{*}{Statistics ($km^{2}$)} & \multicolumn{9}{c}{Interval Year(s)} \\
 & \multicolumn{1}{c}{1} & \multicolumn{1}{c}{2} & \multicolumn{1}{c}{3} & \multicolumn{1}{c}{4} & \multicolumn{1}{c}{5} & \multicolumn{1}{c}{6} & \multicolumn{1}{c}{7} & \multicolumn{1}{c}{8} & \multicolumn{1}{c}{9} \\ \midrule
Scenes & \multicolumn{1}{c}{9} & \multicolumn{1}{c}{8} & \multicolumn{1}{c}{7} & \multicolumn{1}{c}{6} & \multicolumn{1}{c}{5} & \multicolumn{1}{c}{4} & \multicolumn{1}{c}{3} & \multicolumn{1}{c}{2} & \multicolumn{1}{c}{1} \\ \midrule
Expanded Area & 58.28 & 104.68 & 147.59 & 189.74 & 230.79 & 268.93 & 304.70 & 343.45 & 381.28 \\
 (Percentage)& (58.9\%) & (59.0\%) & (59.5\%) & (59.8\%) & (60.4\%) & (60.5\%) & (61.0\%) & (61.5\%) & (63.2\%) \\
Decreased Area & 40.61 & 72.63 & 100.43 & 127.29 & 151.54 & 175.25 & 194.67 & 215.26 & 222.24 \\
  (Percentage) & (41.1\%) & (41.0\%) & (40.5\%) & (40.2\%) & (39.6\%) & (39.5\%) & (39.0\%) & (38.5\%) & (36.8\%) \\ \bottomrule
Accumulated Changes & 98.88 & 177.31 & 248.02 & 317.03 & 382.33 & 444.18 & 499.37 & 558.71 & 603.52 \\ 
Percentage of Changed Area & 0.87\%	& 1.56\%	& 2.18\%	& 2.79\%	& 3.36\%	& 3.90\%	& 4.39\%	& 4.91\%	& 5.30\% \\ \bottomrule
\end{tabular}
}

\end{table*}

\begin{figure*}[htbp]
    \centering
    \begin{subfigure}[t]{0.30\textwidth}
        \includegraphics[width=\linewidth]{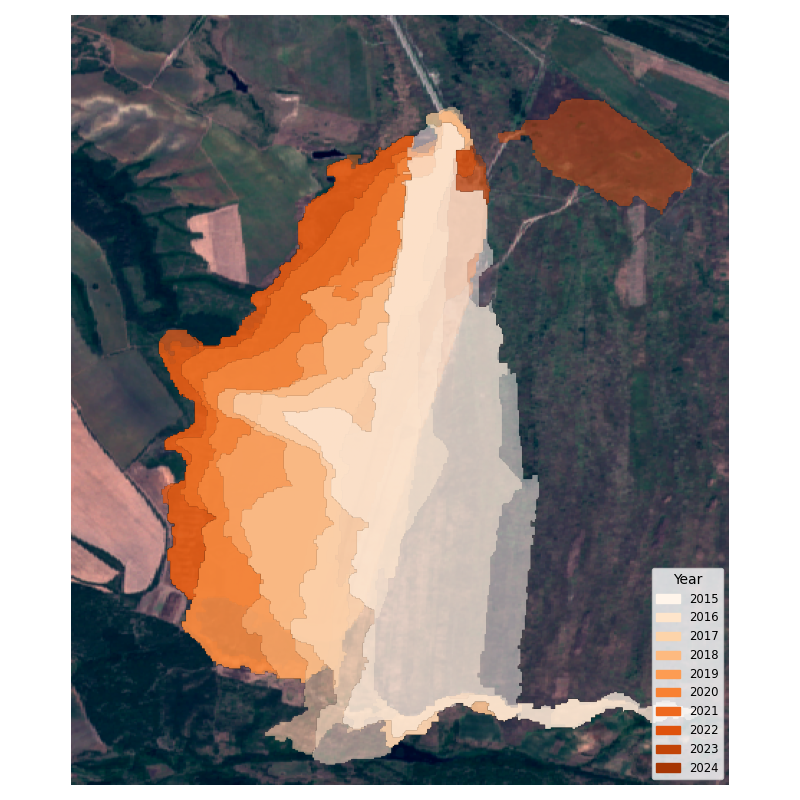}
        \caption{Maritsa Iztok Complex, Bulgaria}
    \end{subfigure}
    \hfill
    \begin{subfigure}[t]{0.30\textwidth}
        \includegraphics[width=\linewidth]{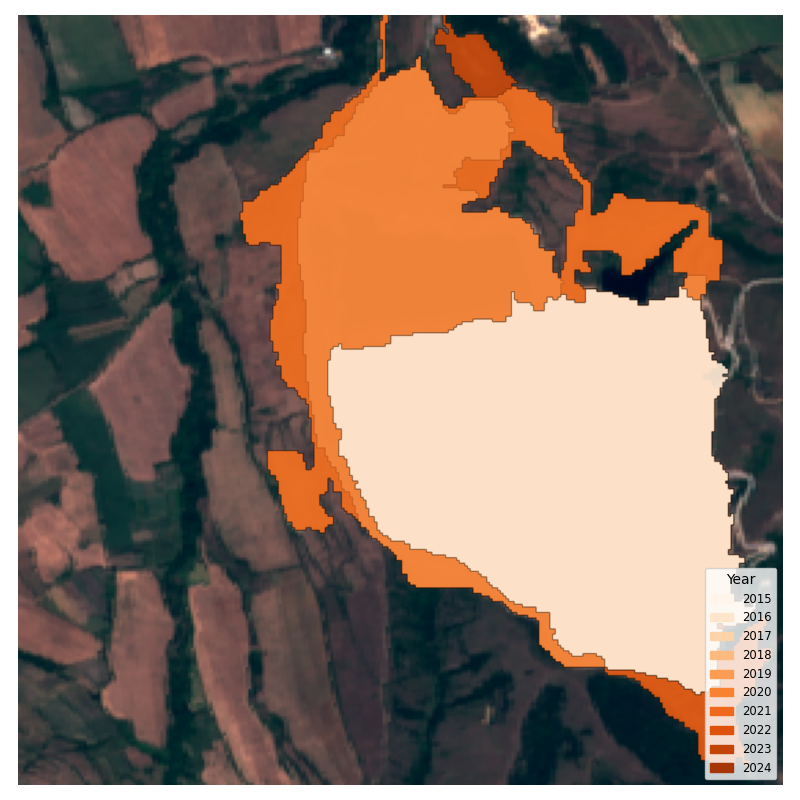}
        \caption{Chelopech Au-Cu Mine, Bulgaria}
    \end{subfigure}
    \hfill
    \begin{subfigure}[t]{0.30\textwidth}
        \includegraphics[width=\linewidth]{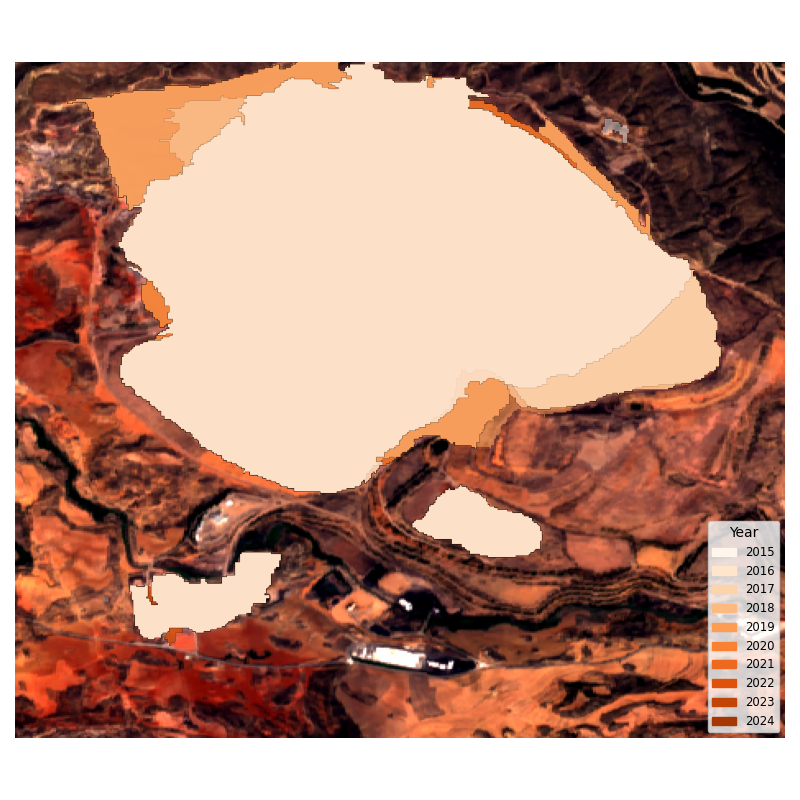}
        \caption{Ariño Coal Mine, Spain} 
    \end{subfigure}

    \begin{subfigure}[t]{0.30\textwidth}
        \includegraphics[width=\linewidth]{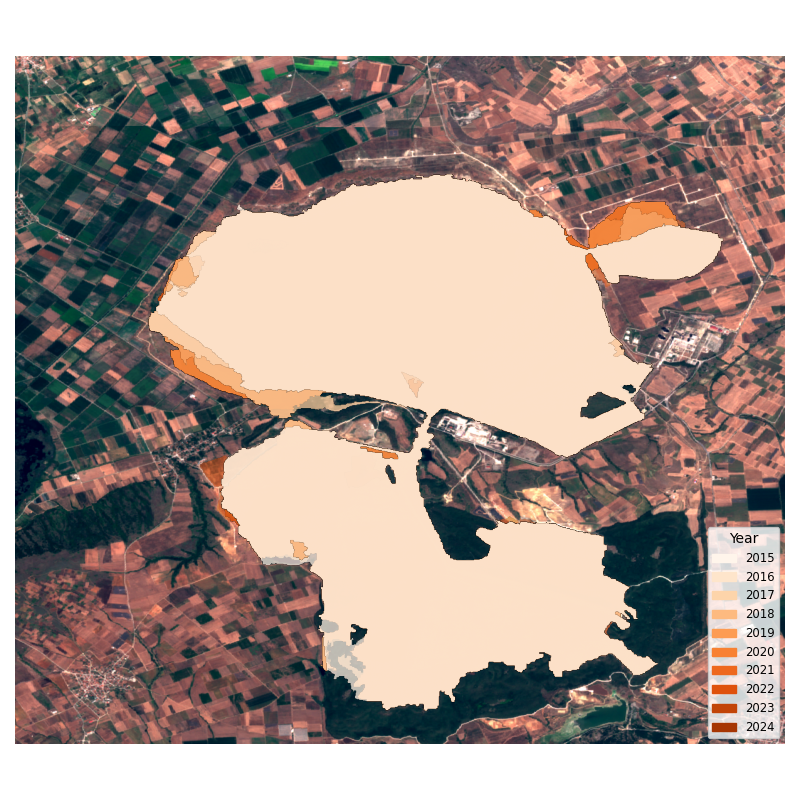}
        \caption{Amyntaio Coal Mine, Greece}
    \end{subfigure}
    \hfill
    \begin{subfigure}[t]{0.30\textwidth}
        \includegraphics[width=\linewidth]{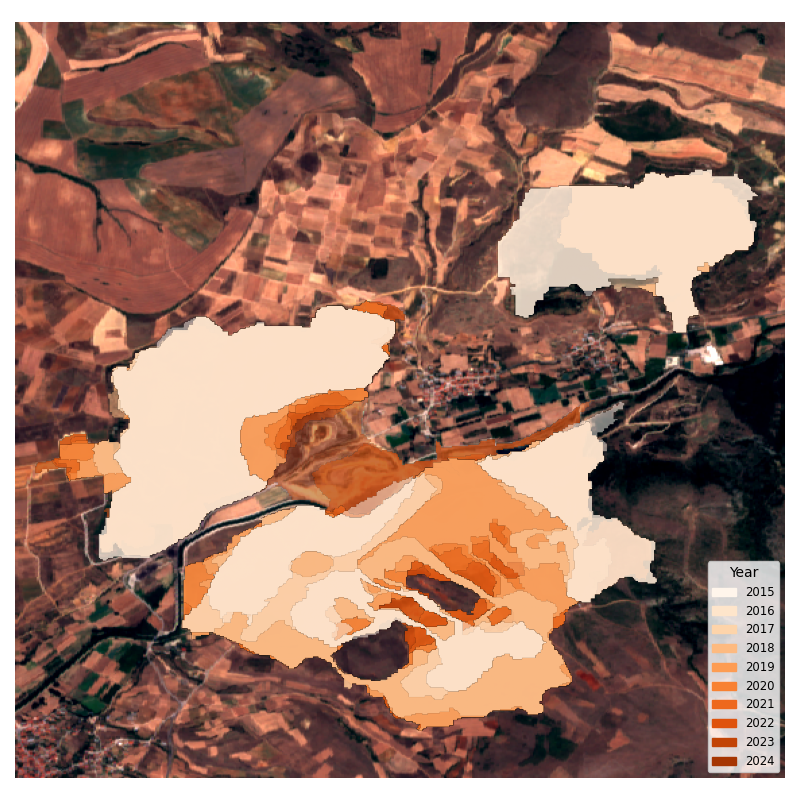}
        \caption{Achlada Coal Mine, Greece} 
    \end{subfigure}
    \hfill
    \begin{subfigure}[t]{0.30\textwidth}
        \includegraphics[width=\linewidth]{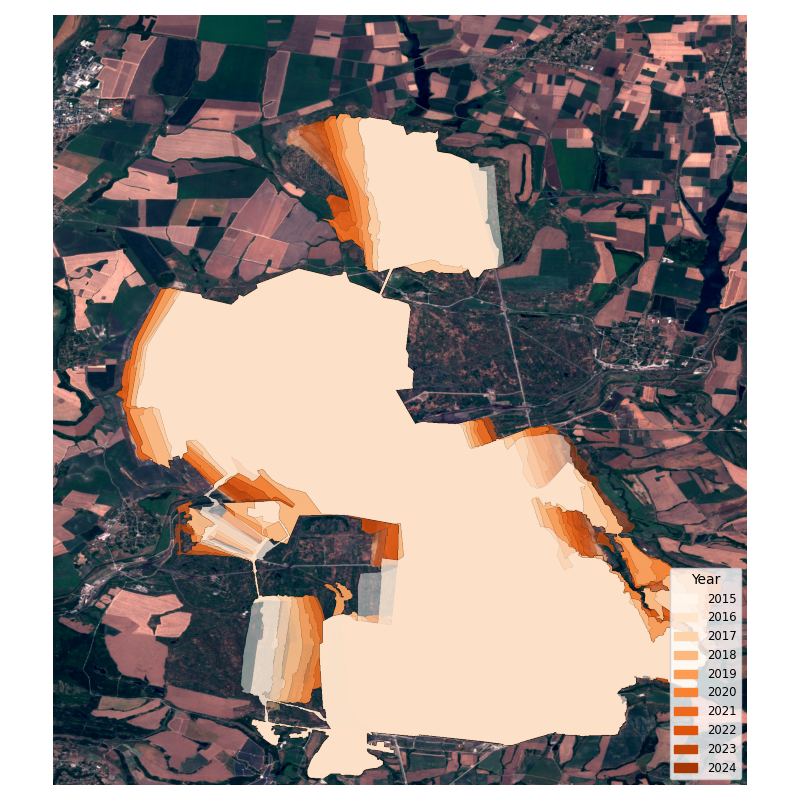}
        \caption{Maritsa Coal Mine, Bulgaria}
    \end{subfigure}

    \begin{subfigure}[t]{0.30\textwidth}
        \includegraphics[width=\linewidth]{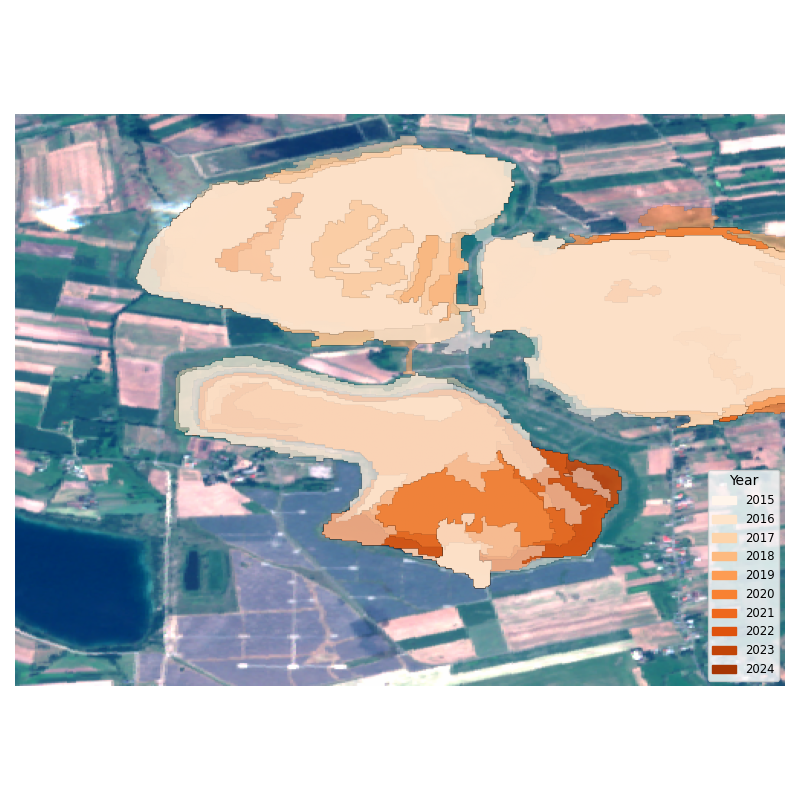}
        \caption{Koźmin Mine, Poland} 
    \end{subfigure}
    \hfill
    \begin{subfigure}[t]{0.30\textwidth}
        \includegraphics[width=\linewidth]{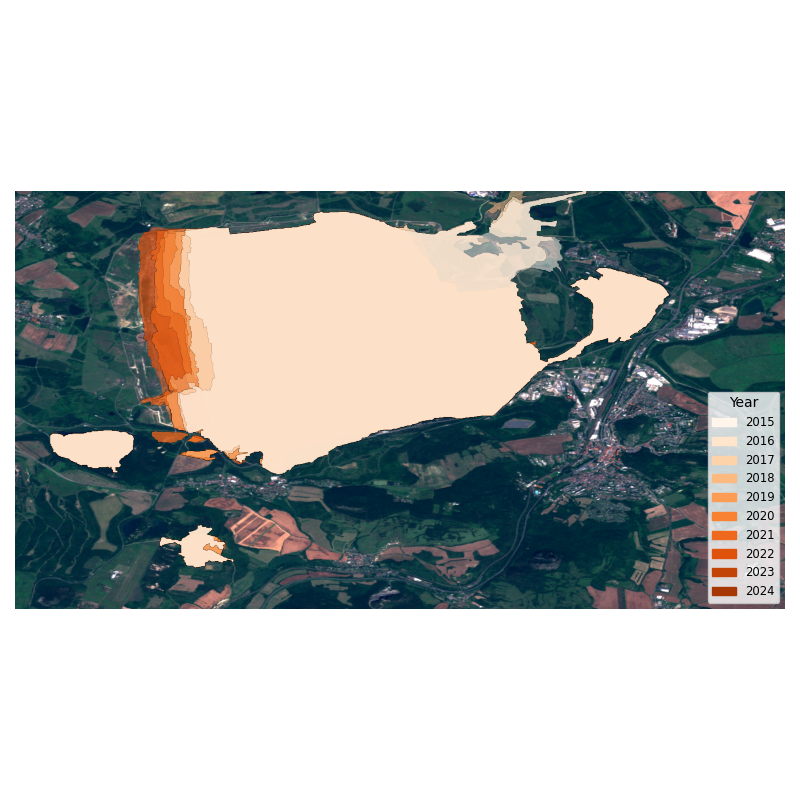}
        \caption{Bilina Coal Mine, Czechia}
    \end{subfigure}
    \hfill
    \begin{subfigure}[t]{0.30\textwidth}
        \includegraphics[width=\linewidth]{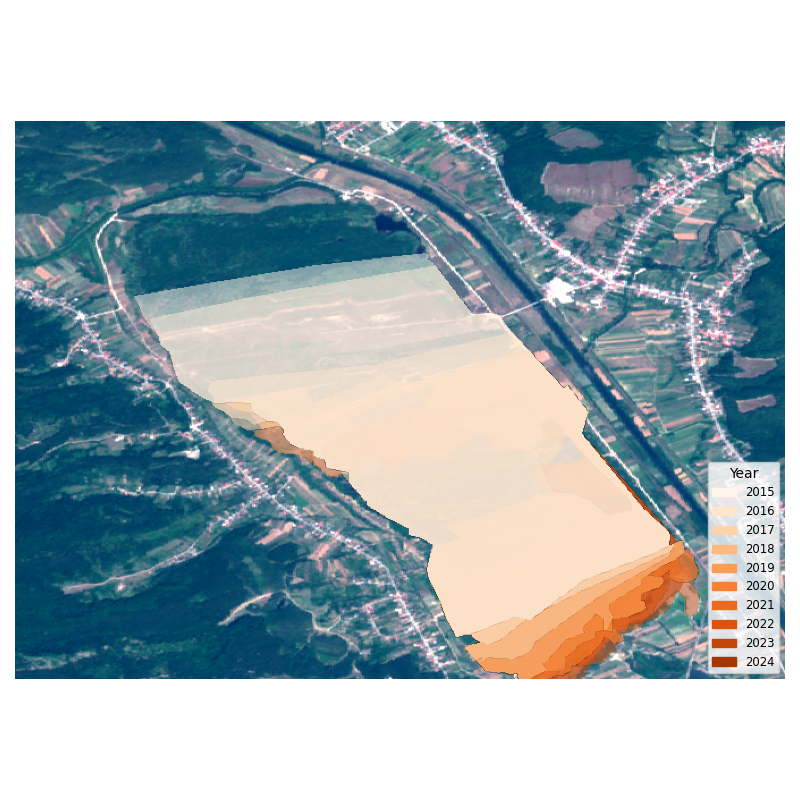}
        \caption{Rosia Coal Mine, Romania} 
    \end{subfigure}

    \begin{subfigure}[t]{0.30\textwidth}
        \includegraphics[width=\linewidth]{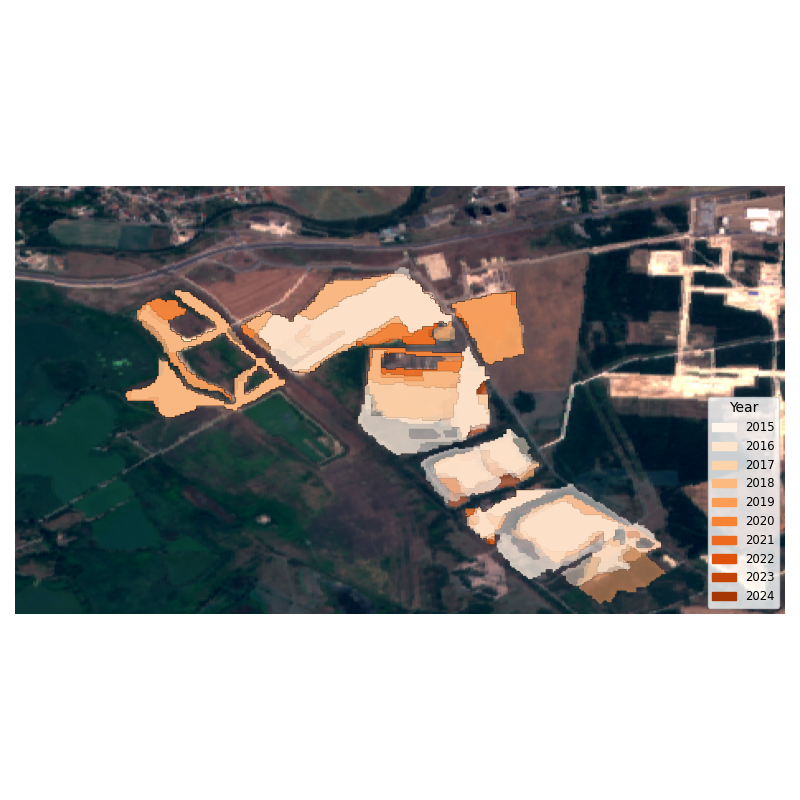}
        \caption{Várpalota Coal Mine, Hungary}
    \end{subfigure}
    \hfill
    \begin{subfigure}[t]{0.30\textwidth}
        \includegraphics[width=\linewidth]{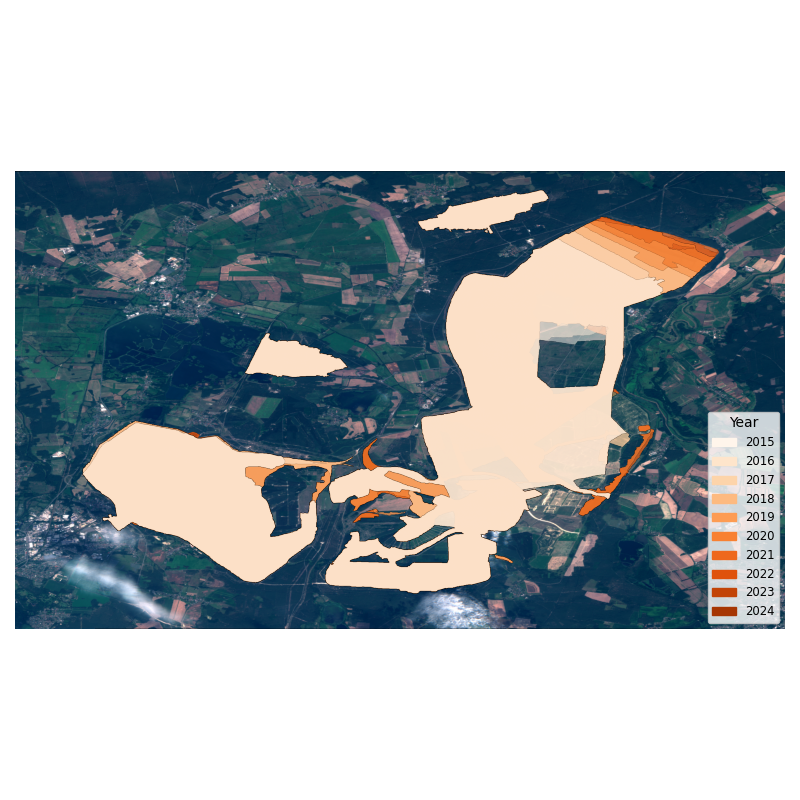}
        \caption{Jänschwalde Coal Mine, Germany}
    \end{subfigure}
    \hfill
    \begin{subfigure}[t]{0.30\textwidth}
        \includegraphics[width=\linewidth]{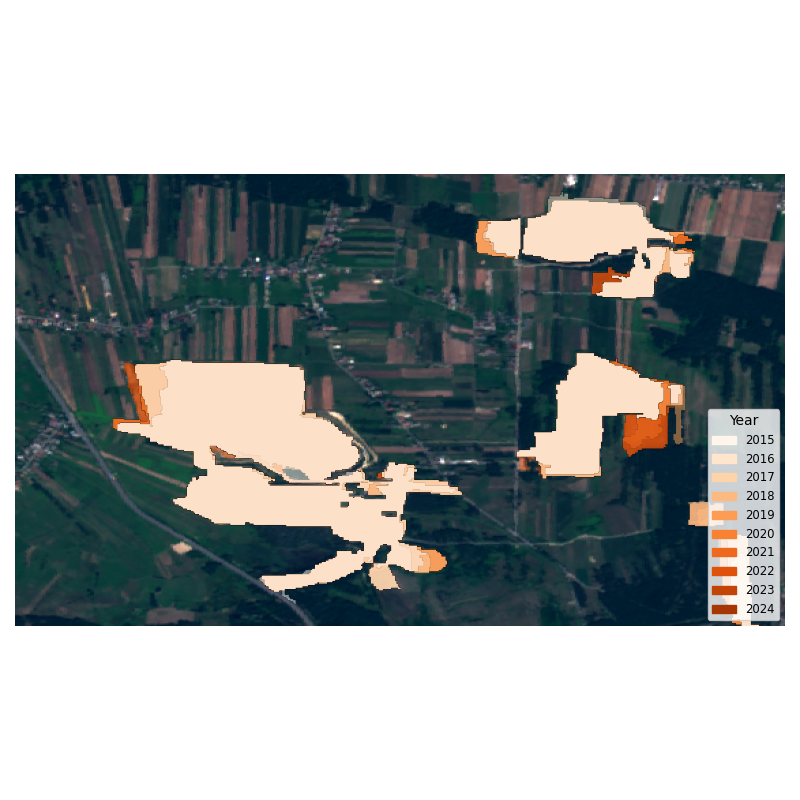}
        \caption{Nowa Wioska Dolostone Quarry, Poland}
    \end{subfigure}

    \caption{Samples of EuroMineNet benchmark. The evolution of each mining site is demonstrated in a color map from 2015 to 2024.}
    \label{fig:3x3_subfigs}
\end{figure*}

\begin{figure*}[htbp]
    \centering

    \begin{subfigure}[t]{0.18\textwidth}
        \includegraphics[width=\linewidth]{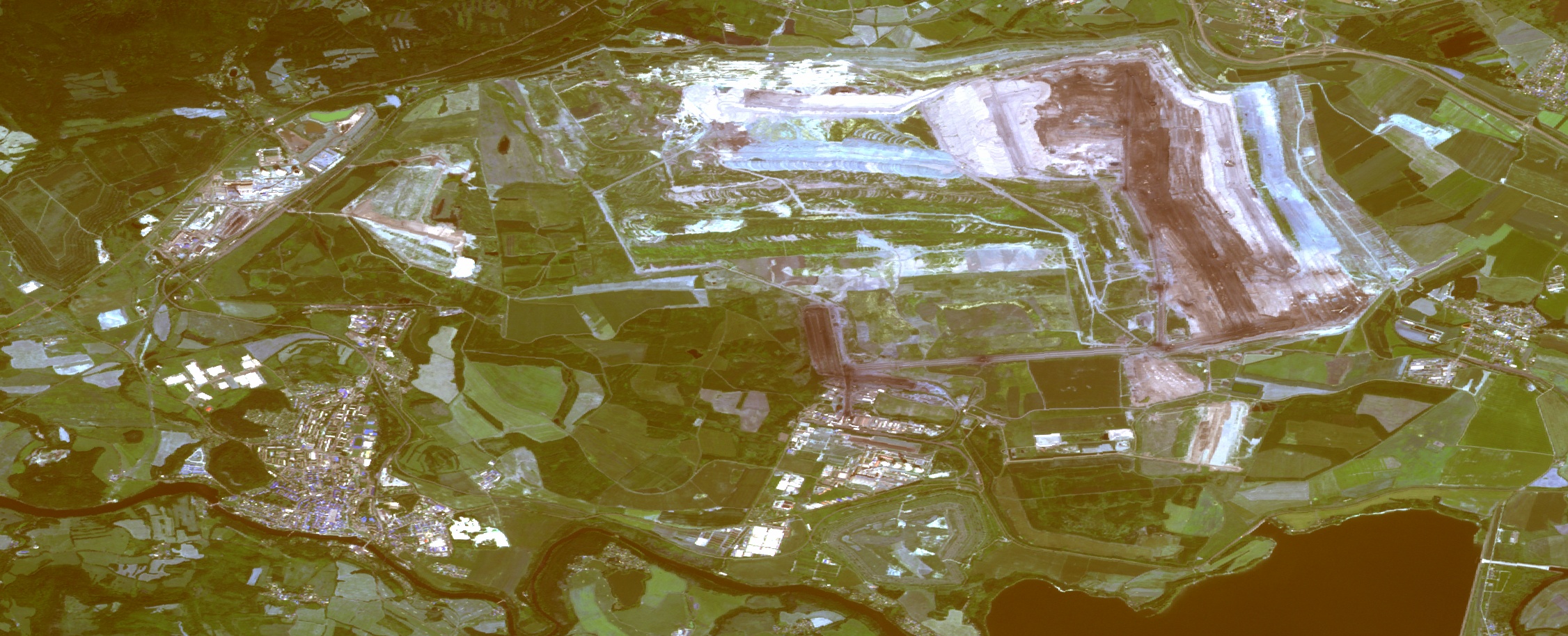}
    \end{subfigure}
    \begin{subfigure}[t]{0.18\textwidth}
        \includegraphics[width=\linewidth]{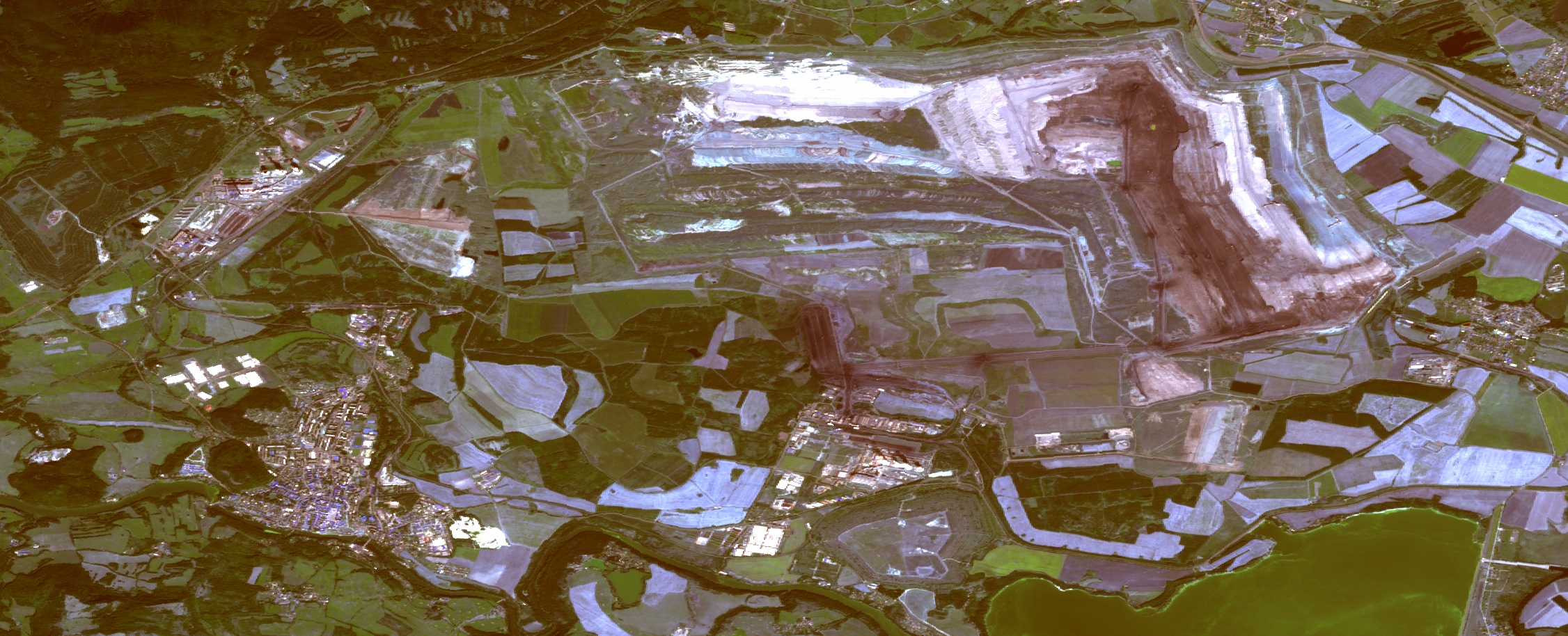}
    \end{subfigure}
    \begin{subfigure}[t]{0.18\textwidth}
        \includegraphics[width=\linewidth]{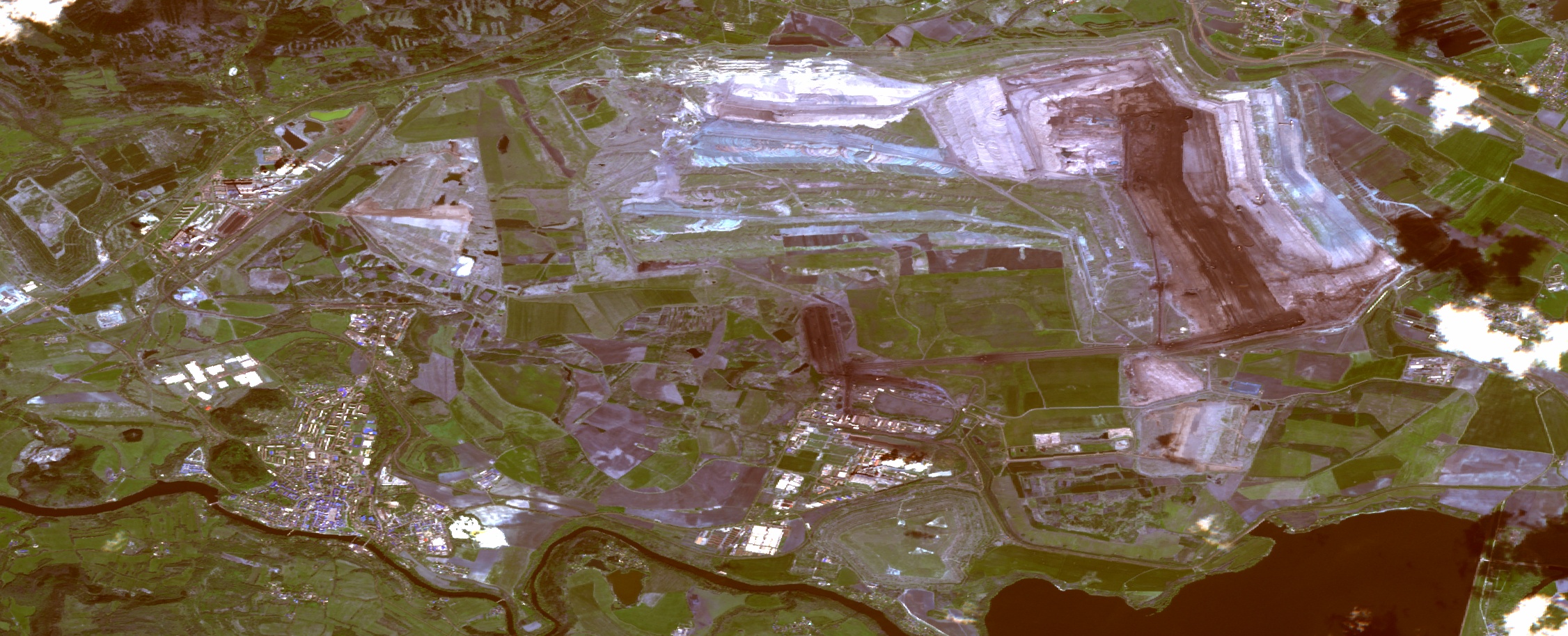}
    \end{subfigure}
    \begin{subfigure}[t]{0.18\textwidth}
        \includegraphics[width=\linewidth]{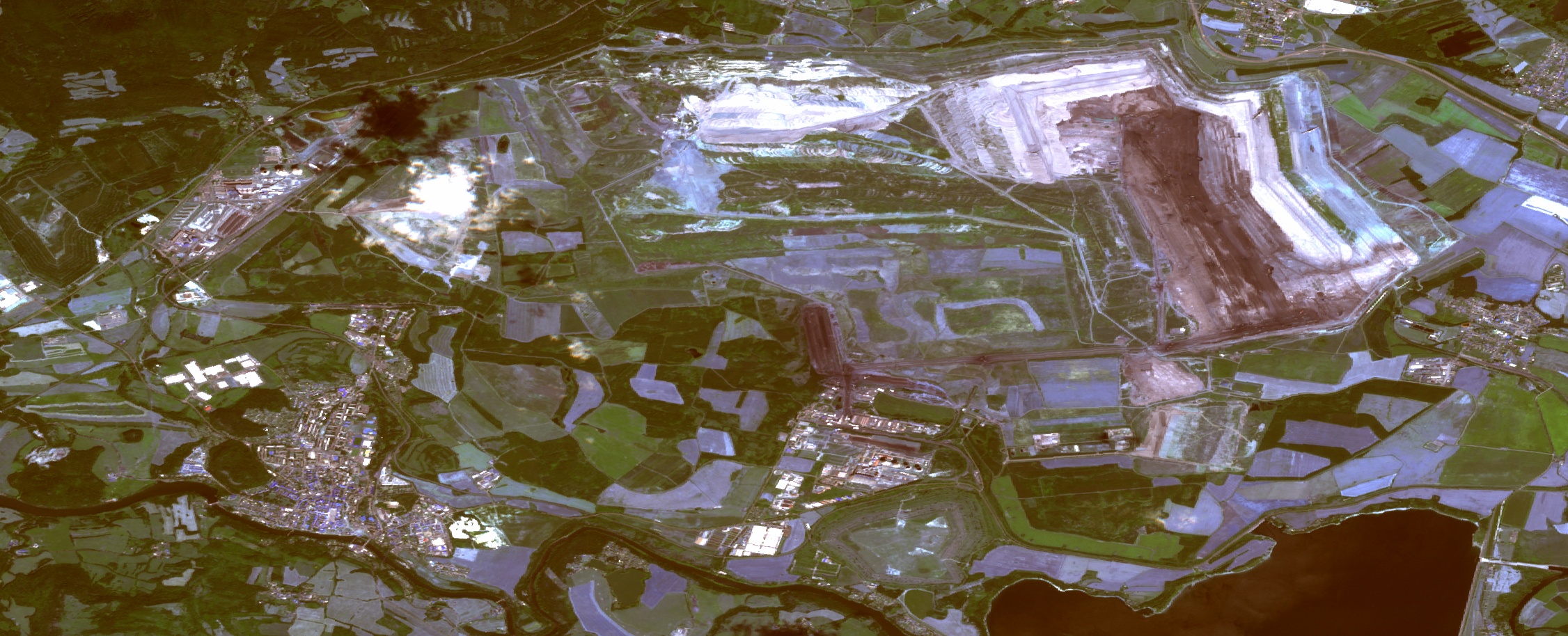}
    \end{subfigure}
    \begin{subfigure}[t]{0.18\textwidth}
        \includegraphics[width=\linewidth]{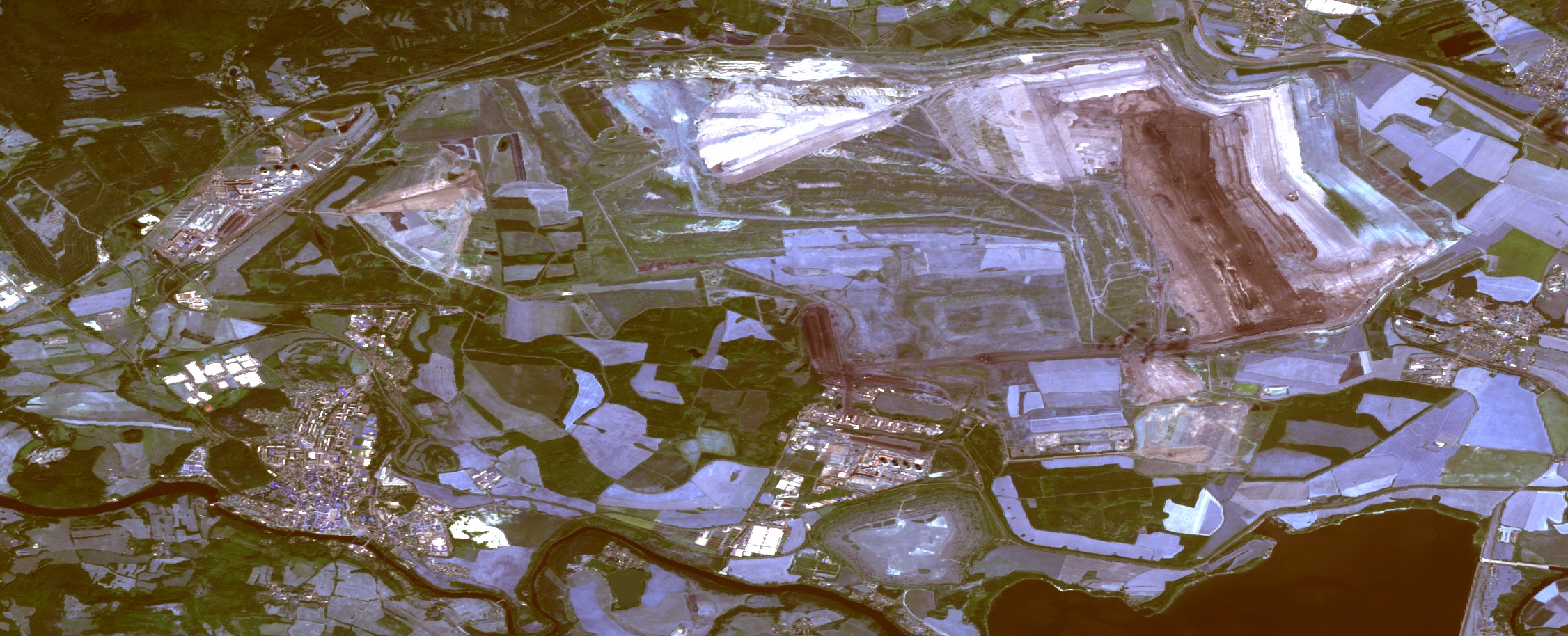}
    \end{subfigure}

    \vspace{1ex}

    \begin{subfigure}[t]{0.18\textwidth}
        \includegraphics[width=\linewidth]{figures/multitemporal_examples/Czechia_2/Year2015.jpg}
    \end{subfigure}
    \begin{subfigure}[t]{0.18\textwidth}
        \includegraphics[width=\linewidth]{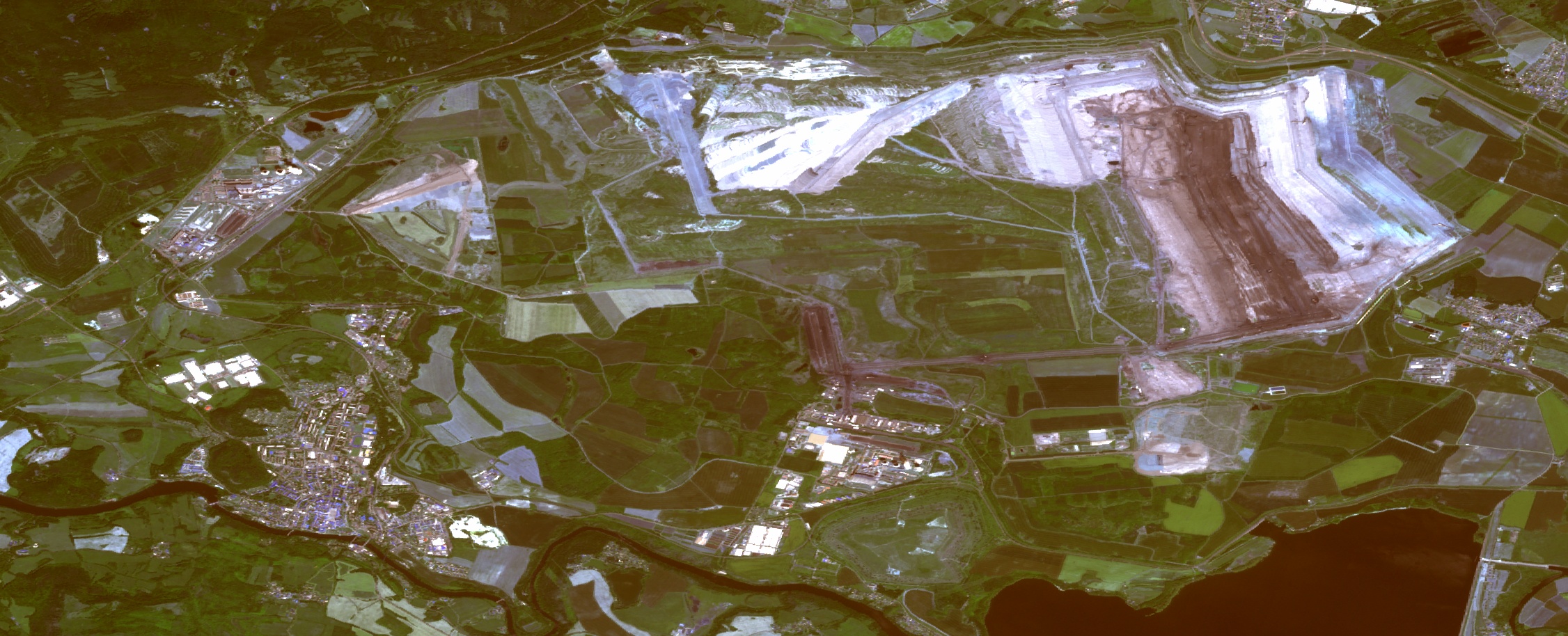}
    \end{subfigure}
    \begin{subfigure}[t]{0.18\textwidth}
        \includegraphics[width=\linewidth]{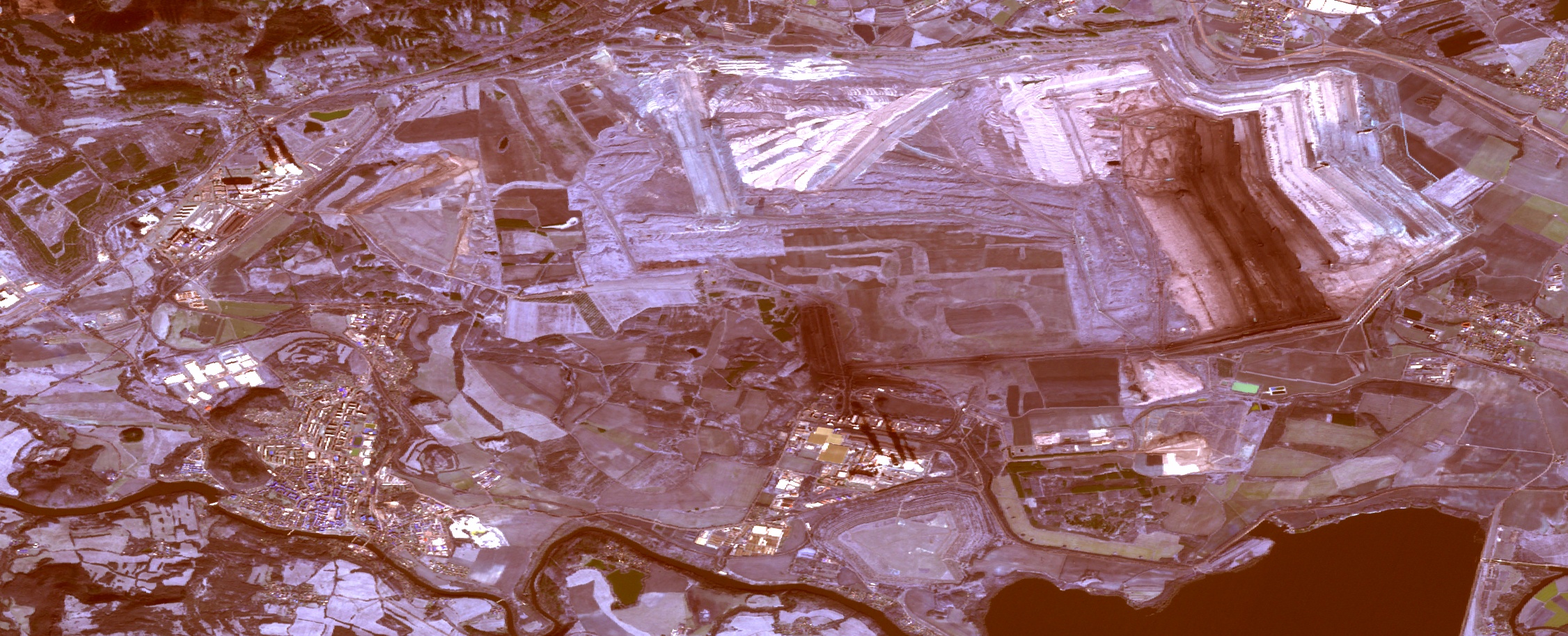}
    \end{subfigure}
    \begin{subfigure}[t]{0.18\textwidth}
        \includegraphics[width=\linewidth]{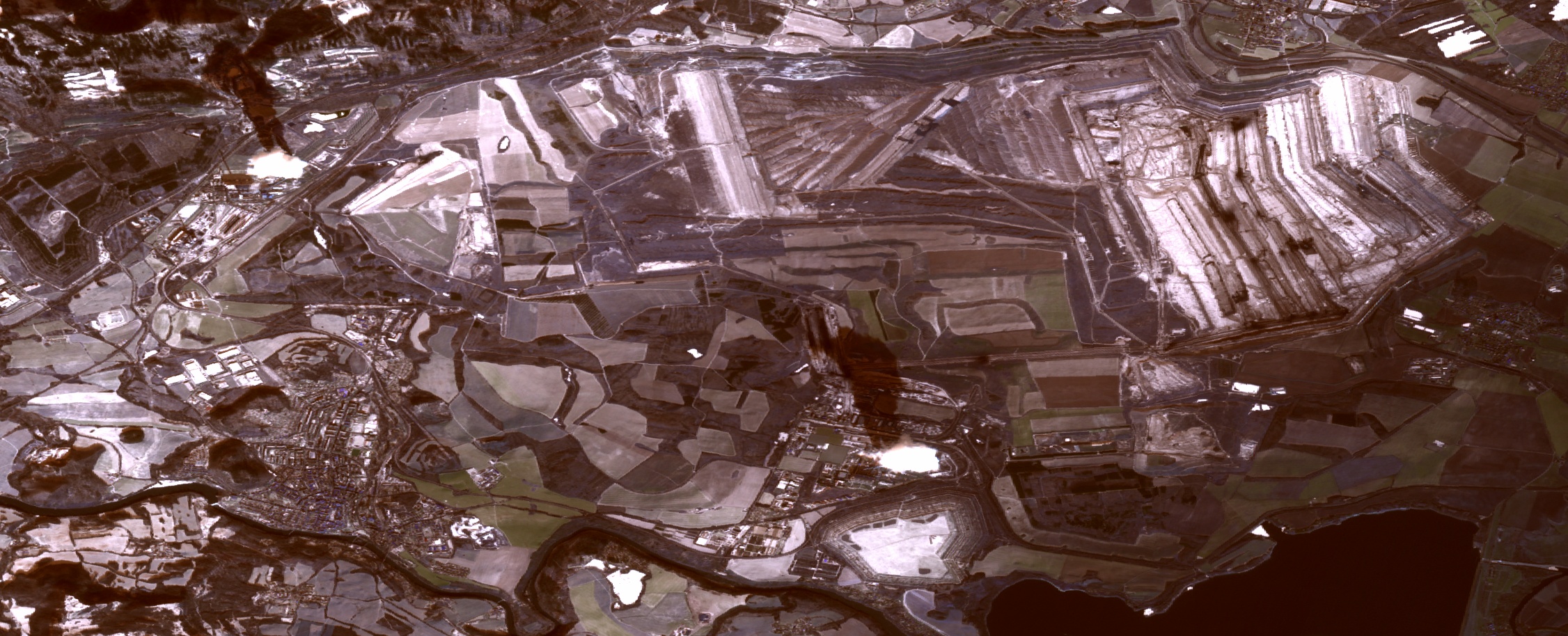}
    \end{subfigure}
    \begin{subfigure}[t]{0.18\textwidth}
        \includegraphics[width=\linewidth]{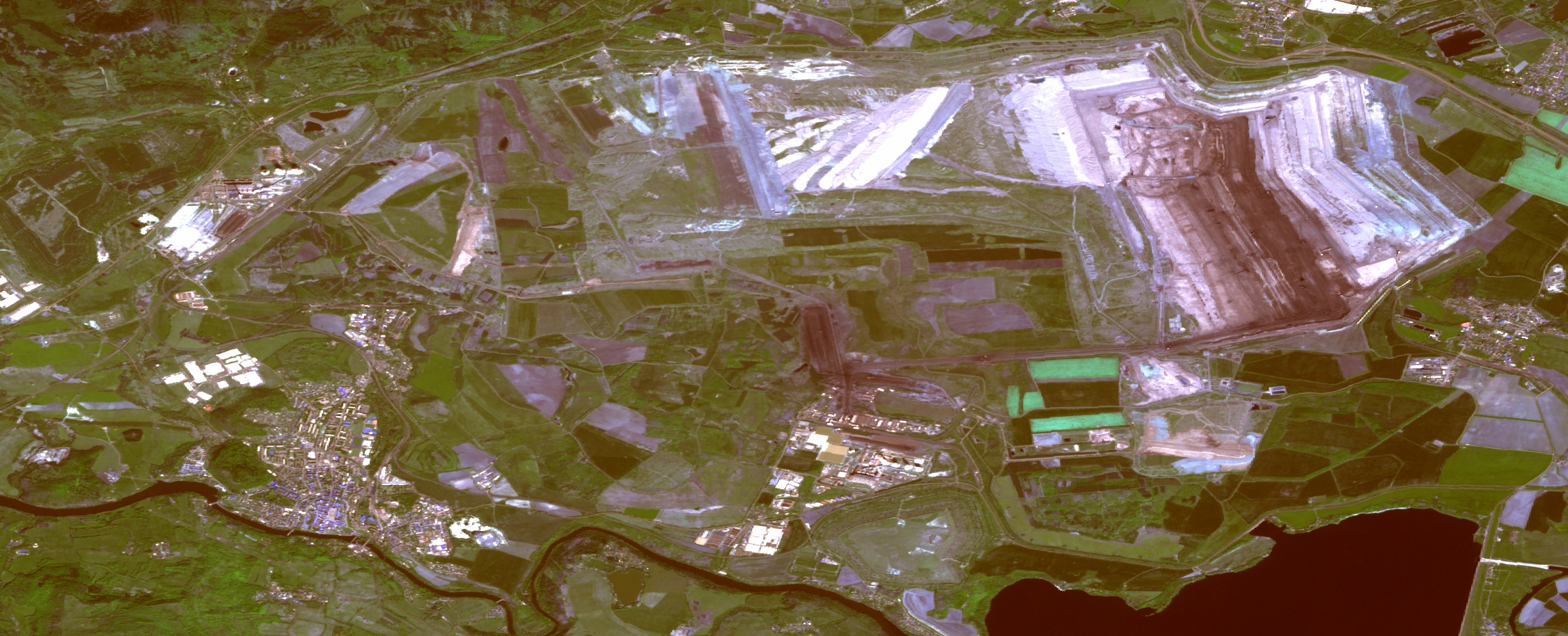}
    \end{subfigure}

    \caption{Multitemporal Earth observation data from Nástup-Tušimice Coal Mine in Czechia from 2015 (top-left) to 2024 (bottom-right). Only optical bands are demonstrated for visualization.}
    \label{fig:spectral_difference}
\end{figure*}
    
\begin{figure*}[!htbp]
    \centering
    \includegraphics[width=\linewidth]{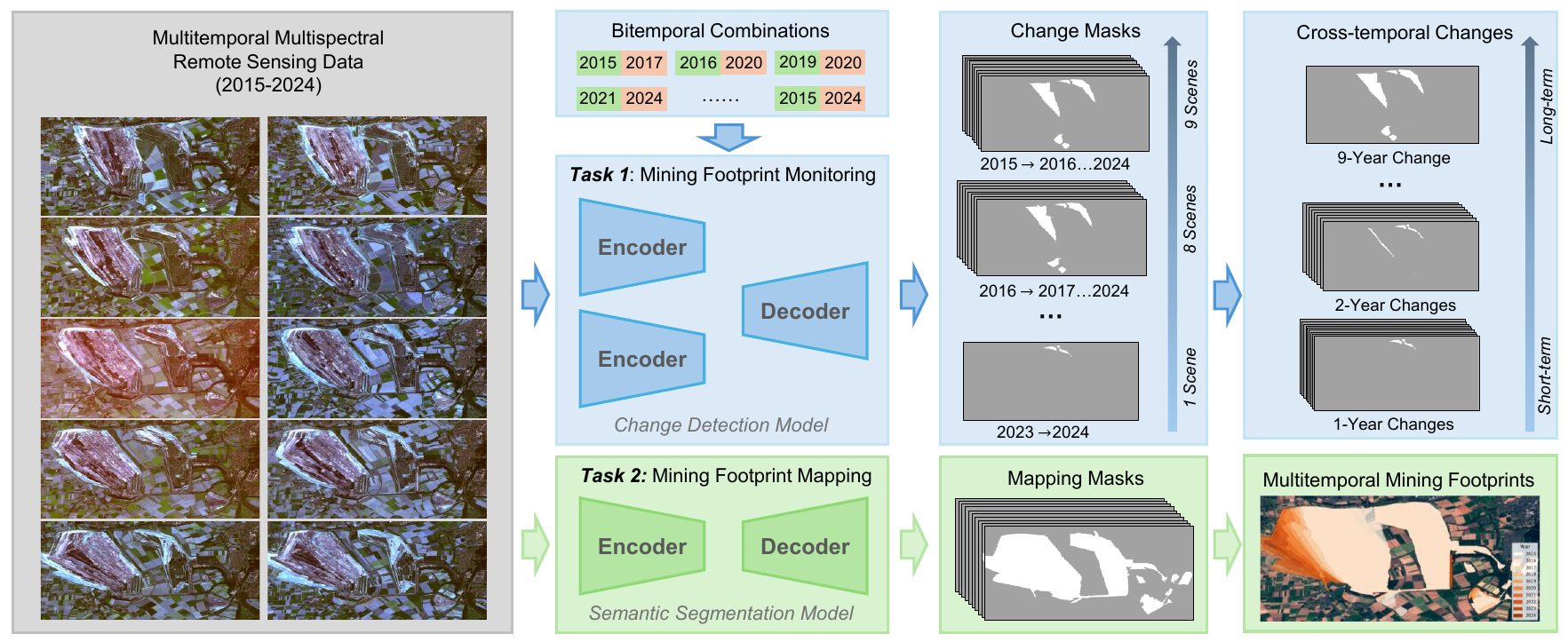}

    \caption{Overall of the EuroMineNet benchmark, consisting of two tasks of mining footprint monitoring and mapping. The mining footprint mapping focuses on the accurate mapping from a single observation by semantic segmentation models, while the mining footprint monitoring focuses on the identification of spatial-temporal variations triggered by the mining activities by change detection models. Only RGB bands are demonstrated for visualization.}
    \label{fig:main_fig}
\end{figure*}
\subsection{Task Overview}
As shown in Fig. \ref{fig:main_fig}, we introduce two complementary tasks based on the proposed EuroMineNet dataset: multitemporal mining footprint mapping and cross-temporal mining footprint monitoring. These tasks differ fundamentally in their objectives and methodological focus. The multitemporal mining footprint mapping task aims to produce accurate, pixel-level delineations of mining footprints from single-date Earth observations across multiple years. By generating consistent annual maps, this task supports detailed assessments of mining extent for each specific year, enabling reliable temporal comparisons without directly modeling inter-annual changes. In contrast, the cross-temporal mining footprint monitoring task focuses on detecting and characterizing mining-induced changes between any two temporal points, regardless of the interval length. This task captures both rapid short-term developments and gradual long-term transformations by explicitly modeling spatiotemporal differences in multitemporal imagery. Together, these tasks provide a unified yet flexible framework for advancing both high-accuracy annual mapping and robust temporal change analysis in mining monitoring.
\subsection{Multitemporal Mining Footprint Mapping}
To monitor the evolution of mining activities over time, we apply semantic segmentation to annual Sentinel-2 imagery spanning the past decade. Each yearly image is independently processed using a deep learning-based segmentation model to generate pixel-level binary maps that classify each pixel as either mine or non-mine. This bi-class mapping approach enables consistent delineation of mining footprints across time, capturing gradual expansions, infrastructure development, and land reclamation. As shown in Table \ref{tab:subsequent_years}, the mining footprint covers approximately 20 percent of all the pixels, with a slight increase by each subsequent year. 

\subsubsection{Temporal Consistency}
The multitemporal mining mapping dataset requires a semantic segmentation model that performs well not only on a single temporal, but also maintains a high accuracy among the mining scenes from other years. However, the mining sites observed at different times can possess different spectral features, due to the imaging conditions such as atmospheric conditions and seasonal variations, as shown in Fig. \ref{fig:spectral_difference}. These variations lead to the heterogeneous styles of images and can significantly interfere with the performance of the semantic segmentation models, which is often referred to as domain shift, a common issue in processing remote sensing imagery. As a result, the segmentation models usually encounter flickering predictions in mapping the mining footprint over different times, though the flickering area does not exactly change.

Therefore, we introduce the concept of temporal consistency as a new perspective to evaluate the model's performance in the segmentation of a multitemporal geospatial scene captured over a continuous period. 
We focus on the prediction of the mining site that should not be flikered into the wrong categories from different temporal observations, while the model should have temporal consistency by predicting the unchanged mining footprint or non-mining footprint consistently without any interference. 
As a result, we construct a change-aware temporal intersection over union (CA-TIoU) metric to evaluate the model's temporal consistency capacity. Let $P_{t_1}, P_{t_2}$ be the predicted masks from two temporal periods of $t_{1}$ and $t_{2}$, and $G_{t_1}, G_{t_2}$ be the corresponding annotated masks. We can first get the non-changed area that should be temporally consistent as:
\begin{equation}
    \mathcal{U}_{i, j} = \left\{ (x, y) \;\middle|\; G_{t_1}(x, y) = G_{t_{2}}(x, y) \right\},
\end{equation}
where $(x,y)$ is the coordinate of the pixels. Then the CA-TIoU can be obtained as:
\begin{equation}
    \mathrm{CA}  \text{-} \mathrm{TIoU}_{t_{1}, t_{2}} = \frac{P_{1}\cap P_{2} \cap \mathcal{U}}{(P_{1}\cup P_{2}) \cap \mathcal{U}}.
\end{equation}

We further apply the CA-TIoU to the multitemporal predictions for a comprehensive evaluation. 
Let $\{P_{t}\}_{t=1}^{T}$ be the predicted masks from $T$ temporal periods and $\{G_{t}\}_{t=1}^{T}$ be the corresponding masks of mining footprint.
We defined a local CA-TIoU (LCA-TIoU) to measure the local temporal consistency from subsequent years, and a global CA-TIoU (GCA-TIoU) to measure the global temporal consistency from all the combinations with full temporal coverage, which can be expressed as follows:
\begin{align}
\text{LCA-TIoU} 
&= \frac{1}{T - 1} \sum_{t=1}^{T-1} \text{CA-TIoU}_{t,t+1} \notag \\
&= \frac{1}{T - 1} \sum_{t=1}^{T-1} 
\frac{|(P_t \cap P_{t+1}) \cap \mathcal{U}_{t,t+1}|}
{|(P_t \cup P_{t+1}) \cap \mathcal{U}_{t,t+1}|}, \\
\text{GCA-TIoU} 
&= \frac{2}{T(T - 1)} \sum_{1 \le i < j \le T} \text{CA-TIoU}_{i,j} \notag \\
&= \frac{2}{T(T - 1)} \sum_{1 \le i < j \le T} 
\frac{|(P_i \cap P_j) \cap \mathcal{U}_{i,j}|}
{|(P_i \cup P_j) \cap \mathcal{U}_{i,j}|}, 
\end{align}
where $\mathcal{U}_{i, j} = \left\{ (x, y) \;\middle|\; G_i(x, y) = G_j(x, y) \right\}$ and $\mathcal{U}_{t, t+1} = \left\{ (x, y) \;\middle|\; G_t(x, y) = G_{t+1}(x, y) \right\}$.

\subsubsection{Semantic Segmentation Models}
The field of semantic segmentation has seen rapid advancements through a variety of architectural paradigms, which can be broadly grouped into two design paradigms: specialized segmentation architectures and unified backbone-based models. The first group comprises models such as UNet \citep{ronneberger2015u}, PSPNet \citep{zhao2017pyramid}, SQNet \citep{treml2016speeding}, and LinkNet \citep{chaurasia2017linknet}, which were explicitly designed for pixel-wise prediction with efficient parameters. For example, UNet is a symmetric encoder-decoder model with skip connections that was originally developed for biomedical image segmentation and remains widely used for its simplicity and effectiveness\citep{ronneberger2015u}. PSPNet introduces a pyramid pooling module to aggregate global and local context, which greatly improves segmentation performance in complex scenes \citep{zhao2017pyramid}. LinkNet extends this idea with residual shortcuts and a lightweight design suited for real-time inference \citep{chaurasia2017linknet}. SQNet achieved efficient segmentation with ELU activation functions, a SqueezeNet-like encoder, followed by parallel dilated convolutions, and a decoder with SharpMask-like refinement modules \citep{treml2016speeding}.
These models often feature custom design components built specifically to improve the efficiency of feature extraction and decoding.

The second paradigm reflects a shift toward more unified and modular segmentation frameworks, which aim to accommodate a wide range of segmentation tasks within a single architectural template. Rather than designing from the ground up, these models incorporate flexible, scalable components that can be easily adapted to different contexts. For example, UperNet-based models combine multi-level feature aggregation with spatial pyramid techniques to create a robust segmentation head that generalizes well across datasets \citep{xiao2018unified}. SegFormer departs from heavy decoder structures and opts for a lightweight, multi-resolution fusion strategy, achieving impressive accuracy with efficient computation \citep{xie2021segformer}. Mask2Former exemplifies the ambition of this paradigm by unifying semantic, instance, and panoptic segmentation into a single architecture based on masked attention and iterative refinement \citep{cheng2021mask2former}. These models emphasize versatility, reusability, and compatibility with modern vision frameworks, enabling consistent performance across tasks without the need for task-specific redesigns.

Together, these two paradigms reflect a shift in semantic segmentation research, from hand-crafted designs tuned for segmentation tasks to modular frameworks that can exploit the generalization power of large pretrained models while maintaining task-specific flexibility. In this study, we adopt both types of models to comprehensively evaluate their performances on the multitemporal multispectral mining footprint mapping application.

\subsection{Cross-temporal Change Detection}
\begin{figure*}[!htbp]
    \centering
    \includegraphics[width=.5\linewidth]{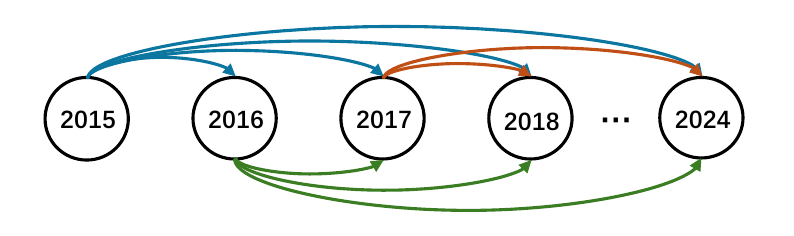}

    \caption{Illustration of cross-temporal change detection. We investigate the mining changes from different starting years with dynamic interval years across the whole period of the last decade. As a result, we obtained a total of $C_{10}^{2}=45$ bitemporal pairs for mining footprint monitoring from each scene.}
    \label{fig:cross_temporal}
\end{figure*}
\subsubsection{Bitemporal Combinations Crafting}
Monitoring changes from both short-period and long-period is critical to assessing the environmental impact driven by mining production. Therefore, we construct a cross-temporal change detection framework to monitor the spatiotemporal variations from different year intervals over a decade-long period. The year intervals are selected from one year to two subsequent years, to the longest interval of 9 years between a decade. With different year intervals preset, we crafted combinations of bitemporal image pairs and produced pixel-level change maps by comparing the LULC maps from two years selected. 

As shown in Table \ref{tab:year-interval} and \ref{fig:cross_temporal}, 9 scenes are obtained while the year interval is set to 1, and 1 scene can be obtained with the longest year interval of 9 years, resulting in a total of $C_{10}^{2} = 45$ bitemporal combinations for each scene. Meanwhile, the percentages of the accumulated changes increase by the interval years, from less than 1 percent for 1 interval year to 5.3\% for 9 interval years. Due to this dynamic progress for the mining development, it remains challenging for change detection models to accurately detect the short-term minor changes and long-term explicit changes simultaneously. As a result, the EuroMineNet dataset requires change detection models to avoid pseudo changes while avoiding missing changes for a more accurate mining monitoring process.

\subsubsection{Change Detection Models}
Change detection models are designed to capture spatiotemporal variations between bitemporal Earth observation (EO) images, with the primary objective of identifying changes in land surface conditions over time. This differs fundamentally from semantic segmentation models, which aim to map semantic categories from a single image. While both tasks rely on pixel-wise classification, change detection models incorporate temporal dynamics, making them more suitable for monitoring transitions in land use and land cover.

Most change detection models follow an encoder–decoder architecture, where the encoder is typically implemented as a Siamese network. In this design, two parameter-shared branches of the encoder simultaneously extract deep features from the two input images. These features are then fused, often through concatenation, subtraction, or attention-based fusion mechanisms, to generate a representation of the temporal change, which is subsequently decoded into a pixel-level change map.

In this work, we adopt the unified change detection (UCD) framework developed in MineNetCD \citep{yu2024minenetcd}, which integrates 20 state-of-the-art change detection models, covering both convolutional and transformer-based architectures. A prominent example is the adaptation of the UNet architecture for CD tasks \citep{daudt2018fully}. The model duplicates the encoder in a Siamese configuration and fuses multiscale features from both temporal branches using skip connections to feed into the decoder. Building upon this architecture, many subsequent methods introduce spatiotemporal fusion modules to improve the representation of changes. These include multilayer perceptrons (MLPs) for deep feature alignment \citep{bandara2022transformer}, token-based transformers for semantic relation modeling \citep{chen2021remote}, and 3D convolutional neural networks (3D CNNs) for capturing local and temporal context \citep{ye2023adjacent}.

However, most of these approaches are optimized for detecting relatively stable, long-term changes, and they often struggle with short-term or incremental dynamics due to the lack of training data covering multiple time scales. Furthermore, current benchmarks primarily focus on optical imagery, limiting the generalization of change detection models to broader environmental monitoring scenarios that require multispectral information.

To address these limitations, we leverage the EuroMineNet dataset, which provides large-scale, densely sampled bitemporal image pairs with rich multispectral content. This allows for the training and evaluation of change detection models under diverse temporal intervals and environmental contexts. To accommodate multispectral data, we adapt the first convolutional layer of CNN-based models and the patch embedding layer of transformer-based models to accept multi-band inputs, enabling effective feature extraction from Sentinel-2 imagery. This setup supports both short-term and long-term monitoring and advances change detection research beyond traditional optical-only benchmarks.

\subsection{Training and Evaluation}
\subsubsection{Training Objectives}
We utilized cross-entropy loss to train the semantic segmentation and change detection models by measuring the pixel-wise discrepancy between predictions and ground truths, as follows:
\begin{equation}
    \mathcal{L}(P(i), G(i))=-\frac{1}{N} \sum_{i=1}^{N} (g_{i}\log(p_{i})+(1-g_{i})\log(1-p_{i})),
\end{equation}
where $i$ indexes a pixel, $p_{i}$ indicates the predicted possibility of the target of the $i$-th pixel of the prediction $P$, and $g_{i}$ denotes the label of the $i$-th pixel of the ground truth $G$. With the multitemporal inputs from a single scene, we accumulate the multitemporal loss for segmentation as follows:
\begin{equation}
    \mathcal{L}_{seg}=\frac{1}{T}\sum_{t=1}^{T}(P_{t}, G_{t}).
\end{equation}
Furthermore, we construct the multitemporal change detection loss as follows:
\begin{equation}
   \mathcal{L}_{CD}=\frac{2}{T(T-1)}\sum_{1\le t_{1}<t_{2}\leq T}(P_{t_{1}, t_{2}}, G_{t_{1}, t_2}), 
\end{equation}
where $G_{t_{1}, t_2}=|G_{t_{1}}-G_{t_{2}}|$, and $P_{t_{1}, t_{2}}$ is the predicted probability of the change detection model from Earth observation data by year $t_{1}$ and $t_{2}$.

Additionally, we apply the same loss accumulation strategy for other loss functions that are designed by some specific models, such as auxiliary loss in UperNet \citep{xiao2018unified} and mask classification loss in Mask2former \citep{cheng2021mask2former}.
\subsubsection{Evaluation Metrics}
We utilize the F1 score as the foundational metric to evaluate model performance, as it effectively balances precision and recall, particularly in imbalanced classification scenarios in mining footprint mapping and monitoring, as shown in Table \ref{tab:subsequent_years} and \ref{tab:year-interval}. Let the mining footprint or the changed area be the positive class; we can calculate the F1 score as follows:
\begin{align}
    \textrm{Pre}&=TP/(TP+FP),\\
    \textrm{Rec}&=TP/(TP+FN),\\
    \textrm{F1}&=(\textrm{Pre}\times \textrm{Rec})/(Pre+Rec), \nonumber\\
    &=2TP/(2TP+FP+FN),
\end{align}
where $TP$ and $TN$ are the numbers of pixels that are correctly detected in the positive and negative classes, respectively. $FP$ and $FN$ are the numbers of pixels that are wrongly detected in the positive and negative classes, respectively. 

Building on this, we further compute derived metrics for multitemporal mining mapping and monitoring tasks with different strategies. On the one hand, for multitemporal mining footprint mapping, we evaluate the models for different years that have a different ratio of mining footprint. After that, we obtain an overall F1 score by calculating the F1 from all the years. We also utilized the LCA-TIoU and GCA-TIoU to evaluate the temporal consistency capability of the models. On the other hand, we calculate the F1 scores for change detection results by the year intervals to provide deeper insights into the models’ ability to detect both short-term and long-term changes. For each year interval setting, we calculated the average of the F1 scores derived from multiple scenes from bitemporal combinations. All the metrics are averaged by different mining sites, with the weight given by the number of pixels.

\section{Experimental Results}
\subsection{Dataset Preparation}
We cropped the Earth observation data and the annotated masks into patches for convenient computation. As the smallest scene is only of size $109\times 586$, we determine the patch size as $160 \times 160$ and we cropped all the data into patches without overlapping, except for the last patch of the height and width dimensions. For the data for which the width or height is smaller than $160$, we padded the data by duplicating part of it. As a result, we obtain $5133$ patches from 133 mining sites for one temporal scene, which accumulated to $51330$ samples for all the temporal observations.

We then split the sites into training, validation, and testing with a ratio of 70\%, 10\%, and 20\%. For multitemporal mining monitoring,  we obtain 35490, 5730, and 10110 samples for training, validation, and testing, respectively. For change detection, we have 159705, 25785, and 45495 samples for training, validation, and testing, respectively.

\subsection{Benchmark Methods}
For multitemporal mining footprint mapping, we selected 11 methods, including 6 models with flexible backbones: Deeplab V3 \citep{chen2017rethinking} with MobileViTV2 \citep{mehta2022separable}, DeeplabV3P \citep{chen2018encoder} with ResNet-101 \citep{he2016deep} and MobileNetV2 \citep{sandler2018mobilenetv2}, Mask2Former \citep{cheng2021mask2former} with Swin Transformer Base (SwinT-B) \citep{liu2021swin}, UperNet \citep{xiao2018unified} with ConvNext-B5 and SwinT-B \citep{liu2021swin}; and 5 pre-built fixed models: LinkNet \citep{chaurasia2017linknet}, UNet \citep{ronneberger2015u}, PSPNet \citep{zhao2017pyramid}, SQNet \citep{treml2016speeding}, and Segformer \citep{xie2021segformer}. For the flexible models with backbones, we utilized pretrained weights from the Huggingface hub for training initialization. 

For cross-temporal mining footprint monitoring, we selected 19 deep learning-based change detection models implemented in the UCD framework: a lightweight network with progressive aggregation and supervised attention (A2Net) \citep{Li_2023_A2Net}, an adjacent-level feature cross-fusion with 3-D CNN (AFCF3D) \citep{ye2023adjacent}, a bitemporal image transformer (BIT) network \citep{chen2021remote}, a change guiding network (CGNet) \citep{han2023change}, a transformer-based Siamese network for change detection (ChangeFormer) \citep{bandara2022transformer}, a dual-branch multi-level inter-temporal network (DMINet) \citep{feng2023change}, a dual task constrained deep Siamese convolutional network (DTCDSCN) \citep{liu2020building}, fully convolutional siamese networks for change detection (FC-EF) \citep{daudt2018fully}, a fully convolutional network within pyramid pooling (FCNPP) \citep{lei2019landslide}, HANet \citep{han2023hanet}, a hierarchical change guiding map network (HCGMNet) \citep{han2023hcgmnet}, an intra-scale cross-interaction and inter-scale feature fusion network (ICIFNet) \citep{feng2022icif}, a deep multi-scale Siamese network with parallel convolutional structure and self-attention (MSPSNet) \citep{guo2021deep}, a region detail preserving network (RDPNet) \citep{chen2022rdp}, a residual UNet (ResUnet) \citep{9553995}, a fully convolutional Siamese concatenated UNet (SiamUnet-Conc)  \citep{daudt2018fully}, a fully convolutional Siamese difference UNet (SiamUnet-Diff) \citep{daudt2018fully}, an integrated Siamese network and nested U-Net (SNUNet) \citep{fang2021snunet}, a network via temporal feature interaction and guided refinement (TFI-GR) \citep{li2022remote}, and a lightweight and effective change detection model called TinyCD \citep{codegoni2023tinycd}. 
\subsection{Experimental Settings}
We utilized the UCD \citep{yu2024minenetcd} framework to run the experiments for change detection, while we adopted the transformers deep learning framework \citep{wolf-etal-2020-transformers} to train semantic segmentation models for mining footprint mapping. Overall, we obtain 11 semantic segmentation models and 20 change detection models for benchmarking the EuroMineNet dataset.

All the experiments are run with the same hyperparameters for a fair comparison. We adopted the Adam optimizer with the learning rate set to $1e-4$. The batch size was set to 32 for each GPU. In addition, we employ a cosine annealing scheduler that gradually reduces the learning rate to $1e-7$ for better model convergence. All experiments were conducted under the
Slurm High-performance computing (HPC) system with a 128-core CPU and 8 NVIDIA Tesla A100 GPUs (40GB of RAM). In addition, the Accelerate \citep{accelerate} package is adopted for fully sharded data-parallel computing to speed up the computation of the models in our multi-GPU environment.

\subsection{Results for Multitemporal mining footprint Mapping}
\begin{figure*}[htbp]
\centering

\newcommand{\imgwidth}{0.08\textwidth}

\newcommand{\rowlabel}[1]{%
  \begin{rotate}{90}%
     \hspace{1em} \footnotesize #1%
  \end{rotate}%
}
\raisebox{0.5\height}{\rowlabel{2015}}\hspace{2pt}%
\begin{subfigure}[t]{\imgwidth}
    \includegraphics[width=\linewidth]{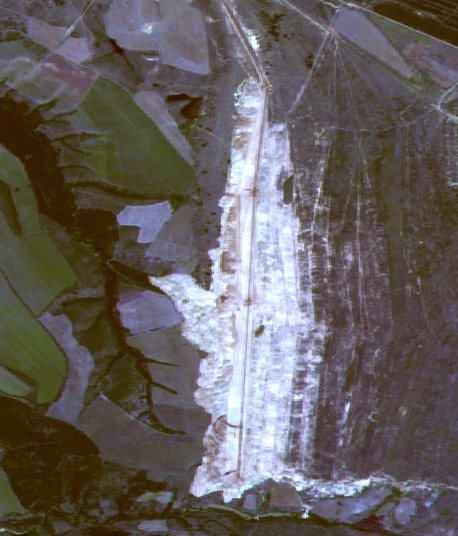}
\end{subfigure}\hfill
\begin{subfigure}[t]{\imgwidth}
    \includegraphics[width=\linewidth]{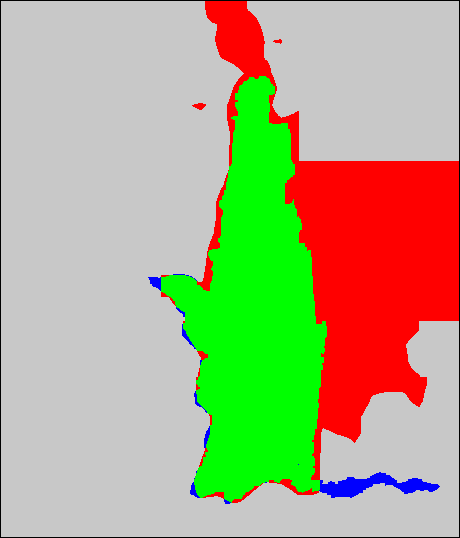}
\end{subfigure}\hfill
\begin{subfigure}[t]{\imgwidth}
    \includegraphics[width=\linewidth]{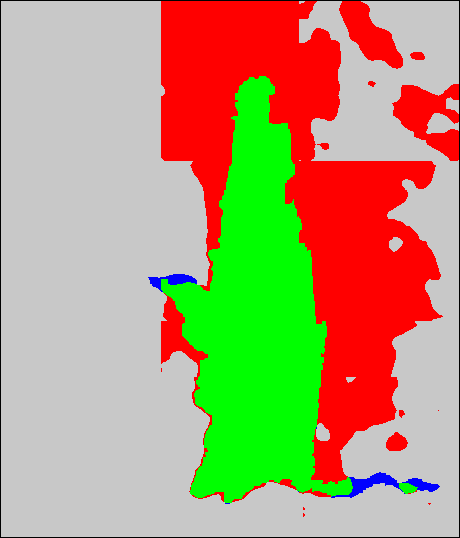}
\end{subfigure}\hfill
\begin{subfigure}[t]{\imgwidth}
    \includegraphics[width=\linewidth]{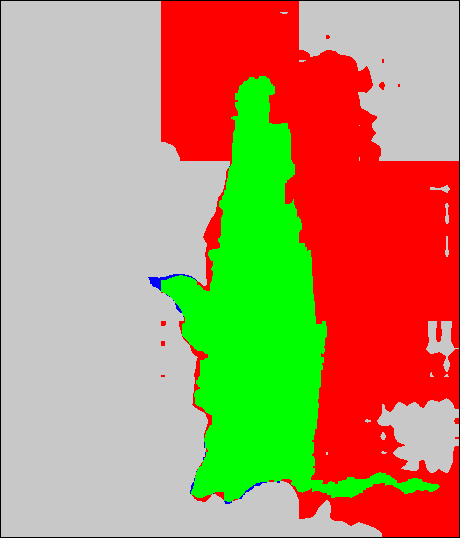}
\end{subfigure}\hfill
\begin{subfigure}[t]{\imgwidth}
    \includegraphics[width=\linewidth]{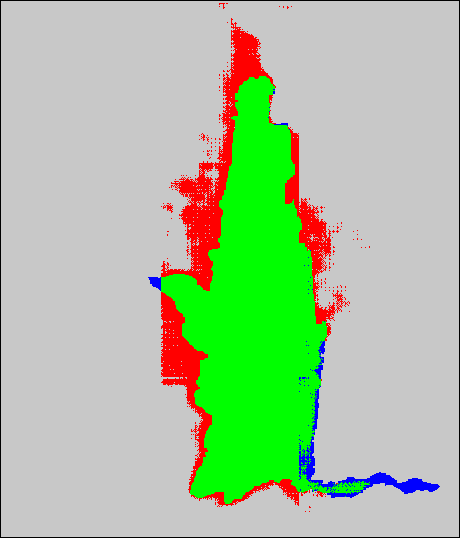}
\end{subfigure}\hfill
\begin{subfigure}[t]{\imgwidth}
    \includegraphics[width=\linewidth]{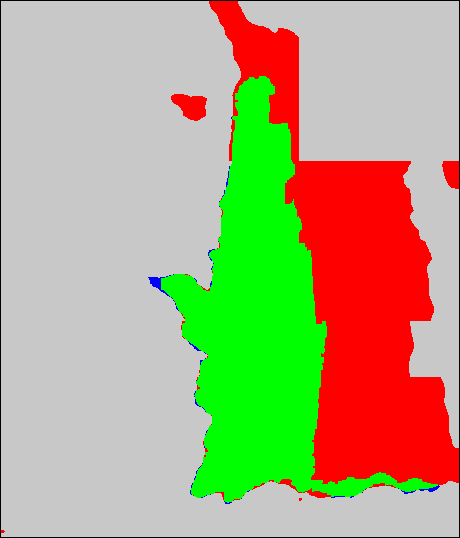}
\end{subfigure}\hfill
\begin{subfigure}[t]{\imgwidth}
    \includegraphics[width=\linewidth]{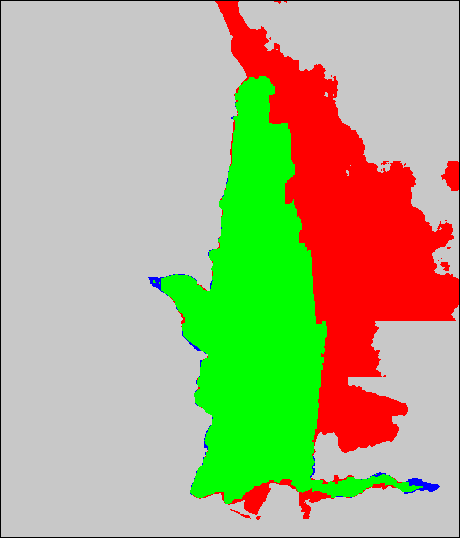}
\end{subfigure}\hfill
\begin{subfigure}[t]{\imgwidth}
    \includegraphics[width=\linewidth]{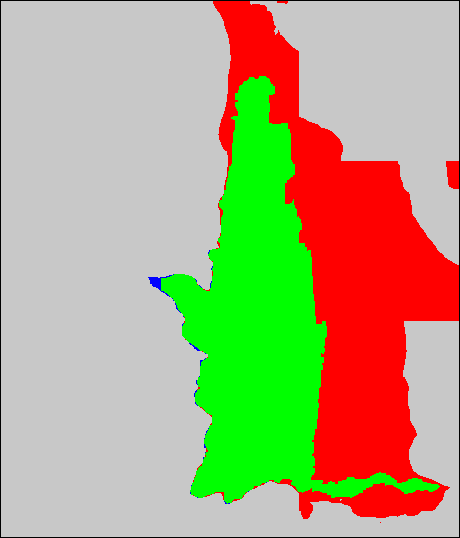}
\end{subfigure}\hfill
\begin{subfigure}[t]{\imgwidth}
    \includegraphics[width=\linewidth]{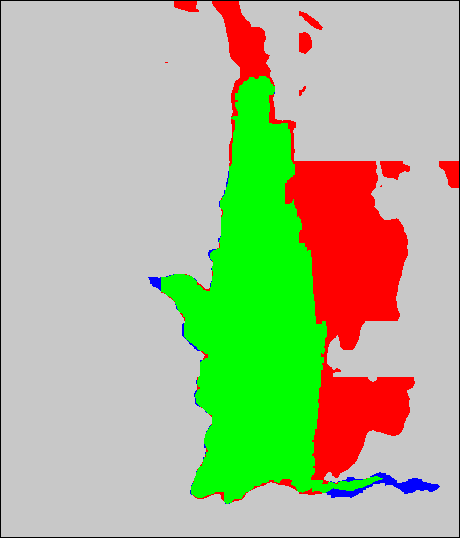}
\end{subfigure}\hfill
\begin{subfigure}[t]{\imgwidth}
    \includegraphics[width=\linewidth]{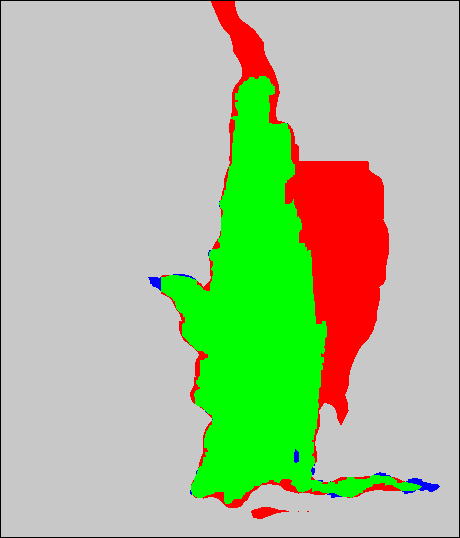}
\end{subfigure}\hfill
\begin{subfigure}[t]{\imgwidth}
    \includegraphics[width=\linewidth]{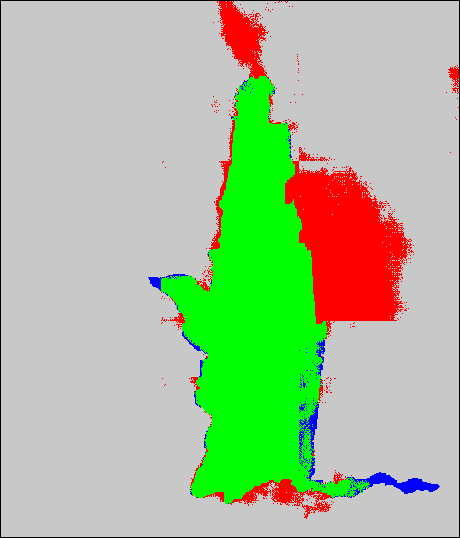}
\end{subfigure}\hfill
\begin{subfigure}[t]{\imgwidth}
    \includegraphics[width=\linewidth]{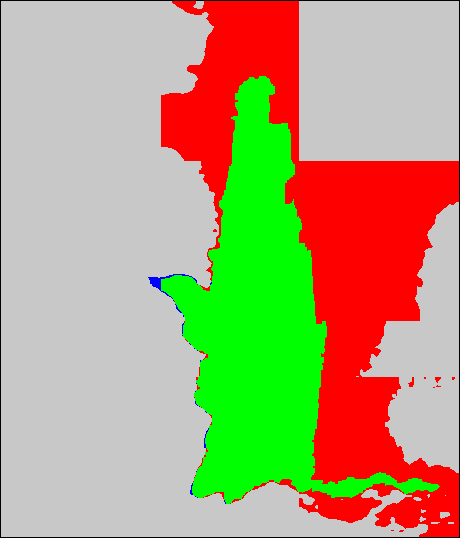}
\end{subfigure}\\[2pt]

\raisebox{0.5\height}{\rowlabel{2016}}\hspace{2pt}%
\begin{subfigure}[t]{\imgwidth}
    \includegraphics[width=\linewidth]{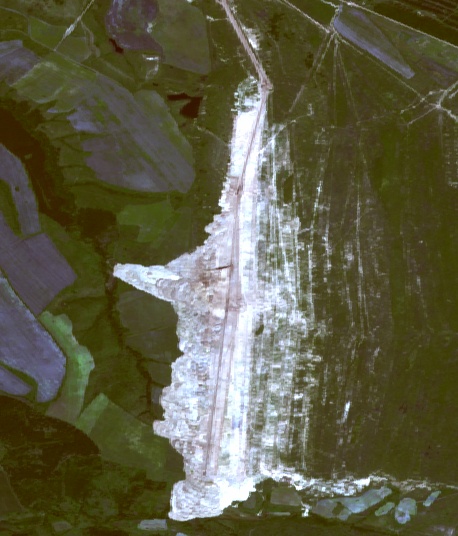}
\end{subfigure}\hfill
\begin{subfigure}[t]{\imgwidth}
    \includegraphics[width=\linewidth]{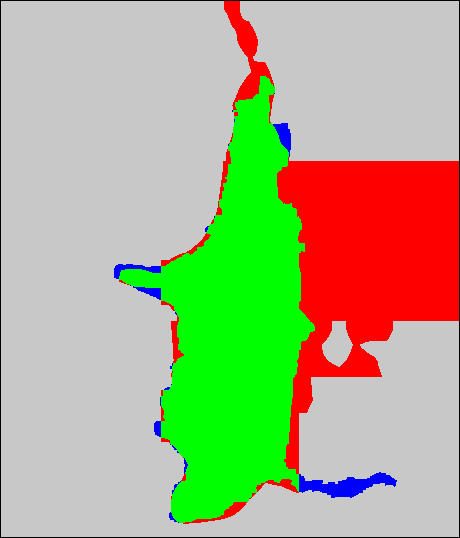}
\end{subfigure}\hfill
\begin{subfigure}[t]{\imgwidth}
    \includegraphics[width=\linewidth]{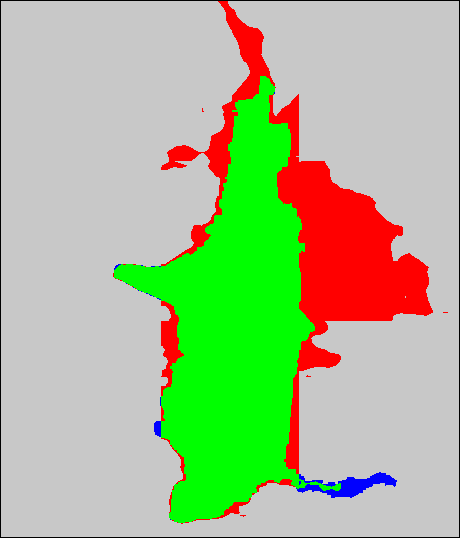}
\end{subfigure}\hfill
\begin{subfigure}[t]{\imgwidth}
    \includegraphics[width=\linewidth]{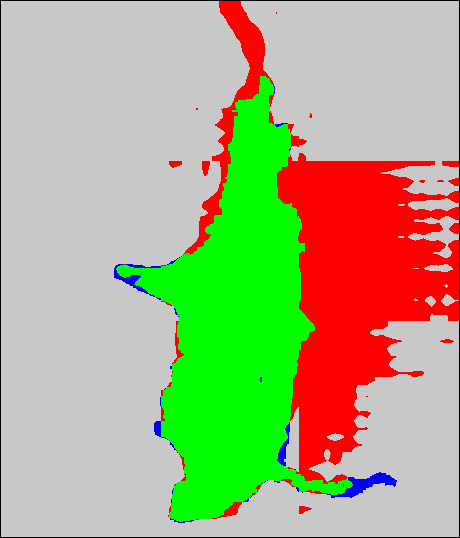}
\end{subfigure}\hfill
\begin{subfigure}[t]{\imgwidth}
    \includegraphics[width=\linewidth]{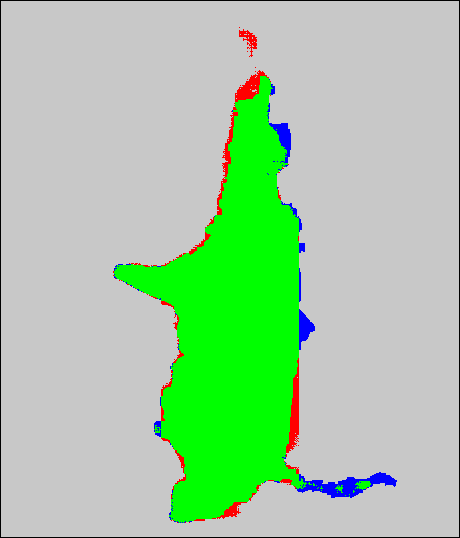}
\end{subfigure}\hfill
\begin{subfigure}[t]{\imgwidth}
    \includegraphics[width=\linewidth]{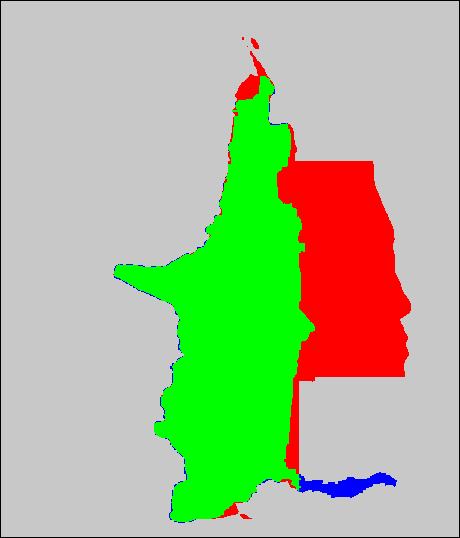}
\end{subfigure}\hfill
\begin{subfigure}[t]{\imgwidth}
    \includegraphics[width=\linewidth]{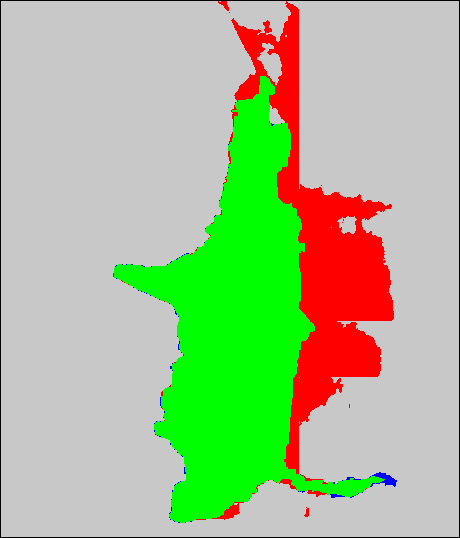}
\end{subfigure}\hfill
\begin{subfigure}[t]{\imgwidth}
    \includegraphics[width=\linewidth]{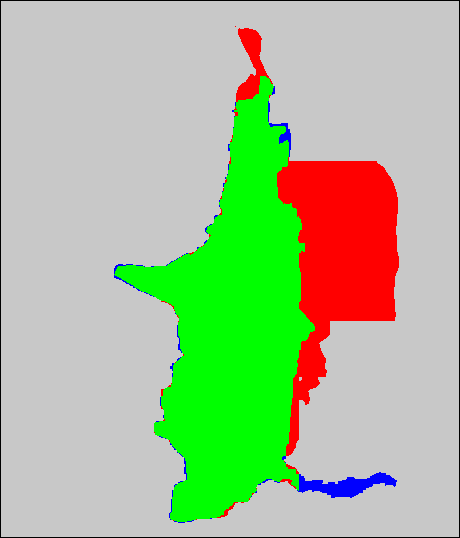}
\end{subfigure}\hfill
\begin{subfigure}[t]{\imgwidth}
    \includegraphics[width=\linewidth]{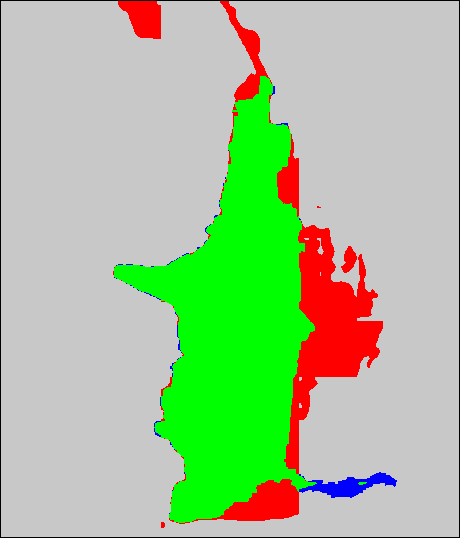}
\end{subfigure}\hfill
\begin{subfigure}[t]{\imgwidth}
    \includegraphics[width=\linewidth]{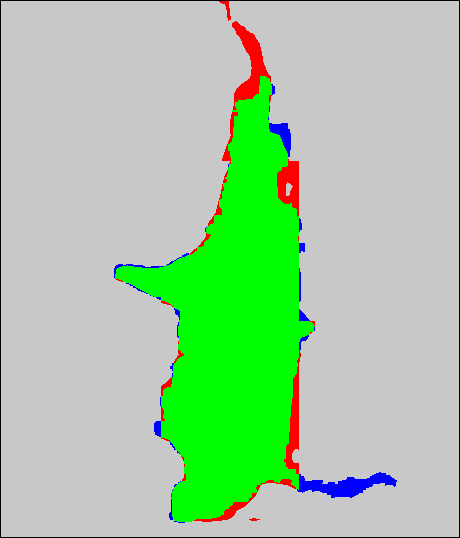}
\end{subfigure}\hfill
\begin{subfigure}[t]{\imgwidth}
    \includegraphics[width=\linewidth]{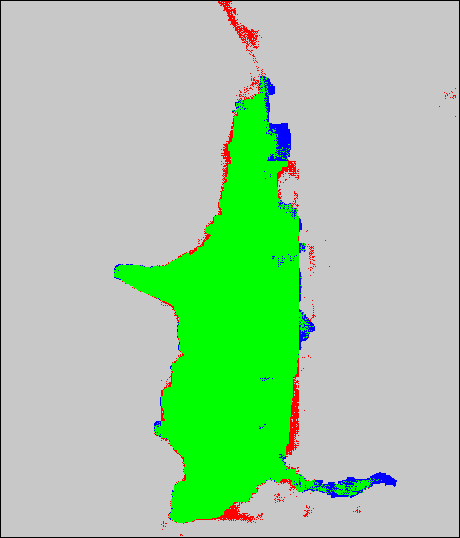}
\end{subfigure}\hfill
\begin{subfigure}[t]{\imgwidth}
    \includegraphics[width=\linewidth]{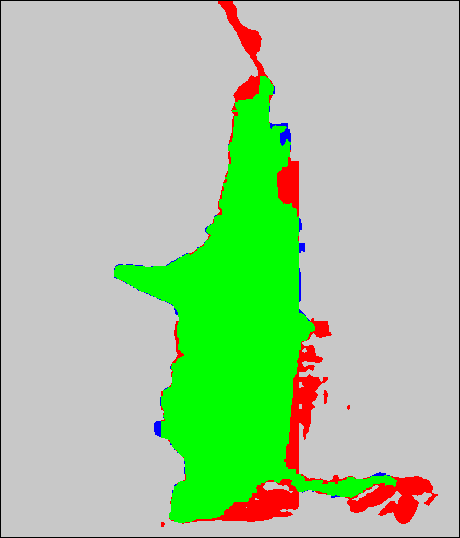}
\end{subfigure}\\[2pt]

\raisebox{0.5\height}{\rowlabel{2017}}\hspace{2pt}%
\begin{subfigure}[t]{\imgwidth}
    \includegraphics[width=\linewidth]{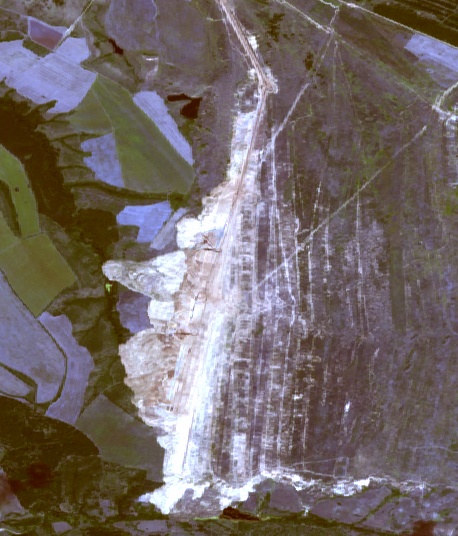}
\end{subfigure}\hfill
\begin{subfigure}[t]{\imgwidth}
    \includegraphics[width=\linewidth]{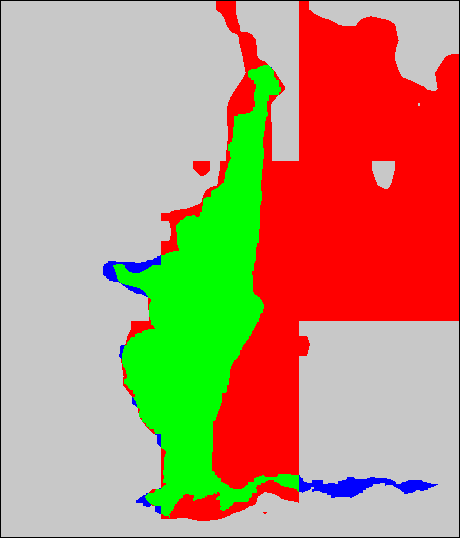}
\end{subfigure}\hfill
\begin{subfigure}[t]{\imgwidth}
    \includegraphics[width=\linewidth]{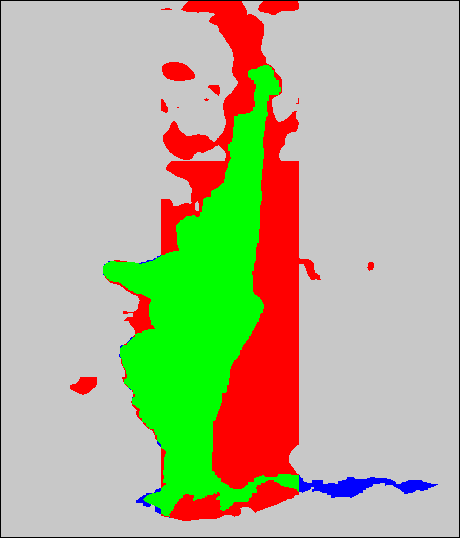}
\end{subfigure}\hfill
\begin{subfigure}[t]{\imgwidth}
    \includegraphics[width=\linewidth]{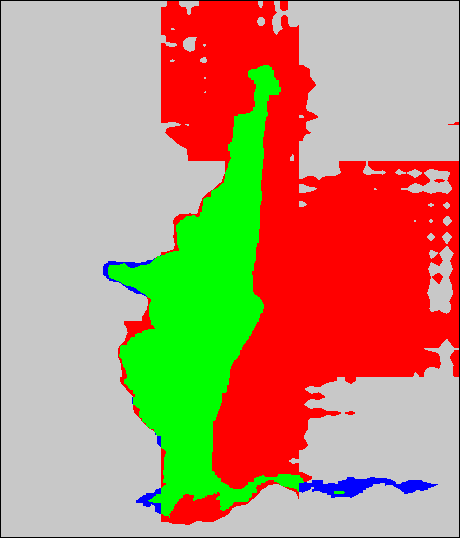}
\end{subfigure}\hfill
\begin{subfigure}[t]{\imgwidth}
    \includegraphics[width=\linewidth]{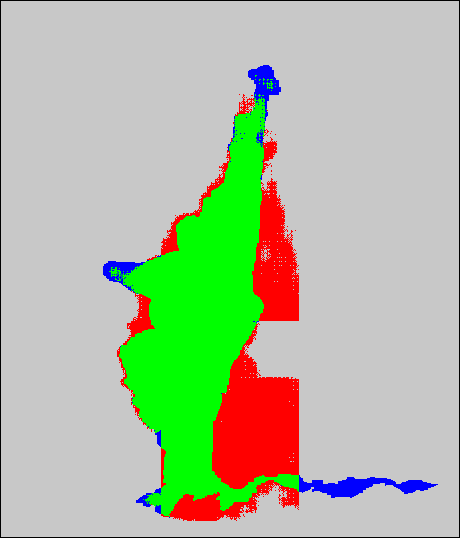}
\end{subfigure}\hfill
\begin{subfigure}[t]{\imgwidth}
    \includegraphics[width=\linewidth]{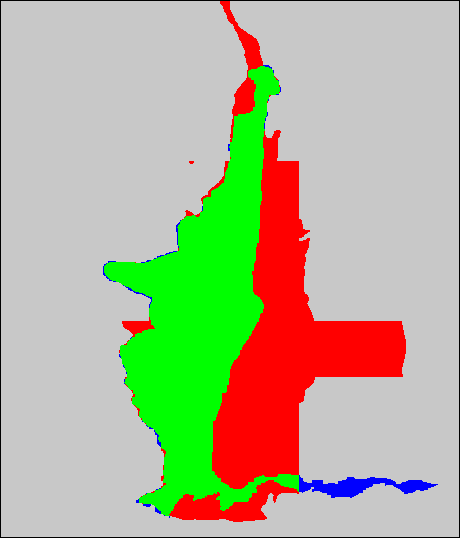}
\end{subfigure}\hfill
\begin{subfigure}[t]{\imgwidth}
    \includegraphics[width=\linewidth]{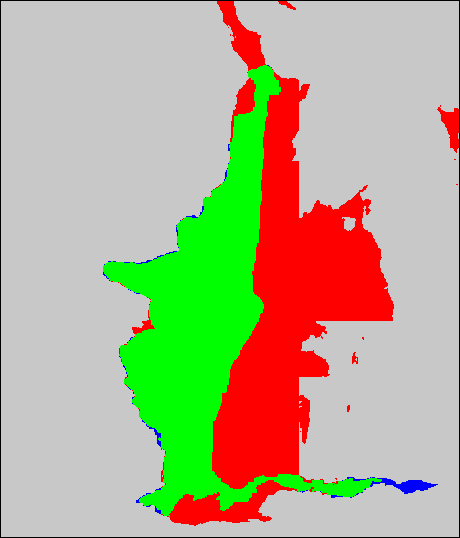}
\end{subfigure}\hfill
\begin{subfigure}[t]{\imgwidth}
    \includegraphics[width=\linewidth]{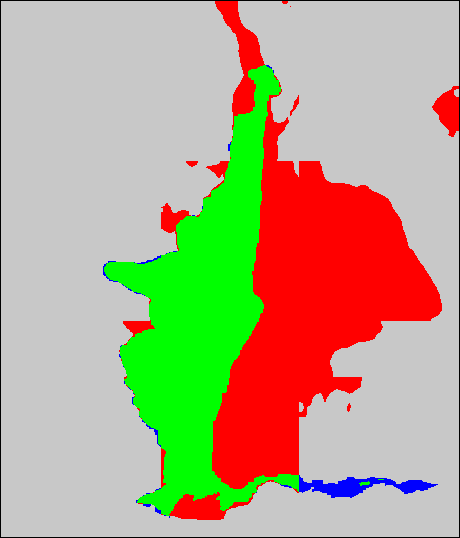}
\end{subfigure}\hfill
\begin{subfigure}[t]{\imgwidth}
    \includegraphics[width=\linewidth]{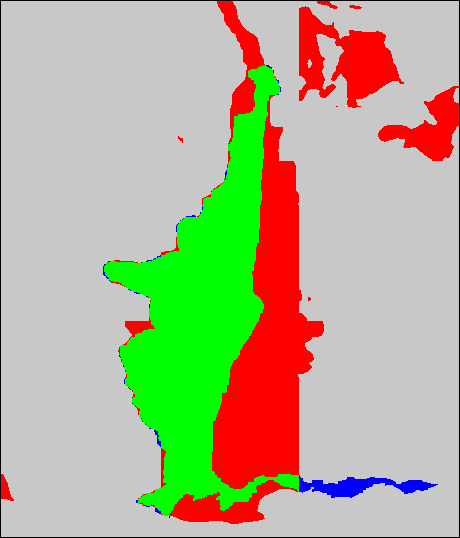}
\end{subfigure}\hfill
\begin{subfigure}[t]{\imgwidth}
    \includegraphics[width=\linewidth]{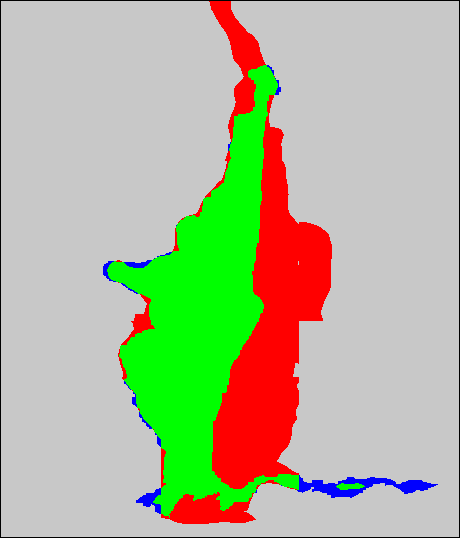}
\end{subfigure}\hfill
\begin{subfigure}[t]{\imgwidth}
    \includegraphics[width=\linewidth]{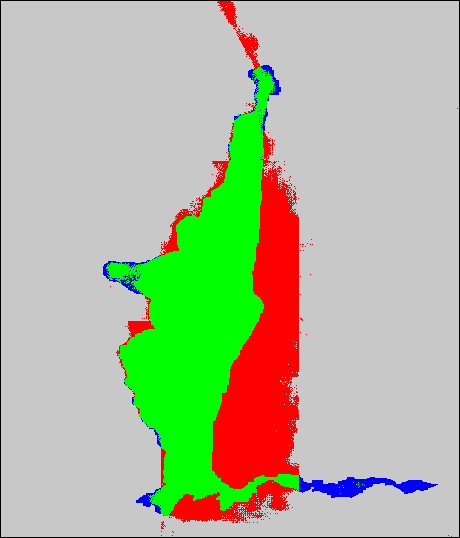}
\end{subfigure}\hfill
\begin{subfigure}[t]{\imgwidth}
    \includegraphics[width=\linewidth]{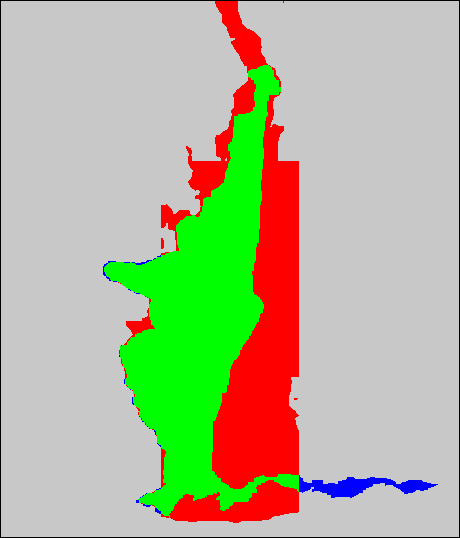}
\end{subfigure}\\[2pt]

\raisebox{0.5\height}{\rowlabel{2018}}\hspace{2pt}%
\begin{subfigure}[t]{\imgwidth}
    \includegraphics[width=\linewidth]{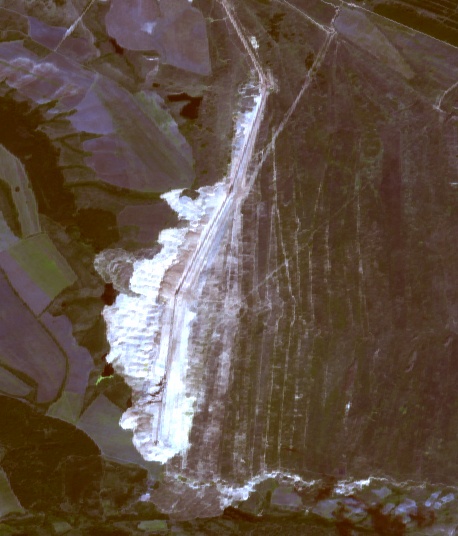}
\end{subfigure}\hfill
\begin{subfigure}[t]{\imgwidth}
    \includegraphics[width=\linewidth]{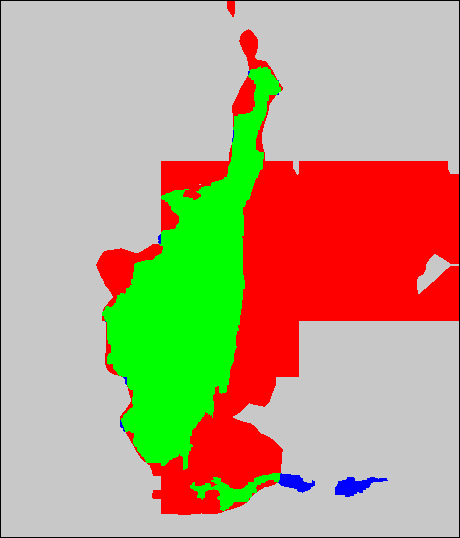}
\end{subfigure}\hfill
\begin{subfigure}[t]{\imgwidth}
    \includegraphics[width=\linewidth]{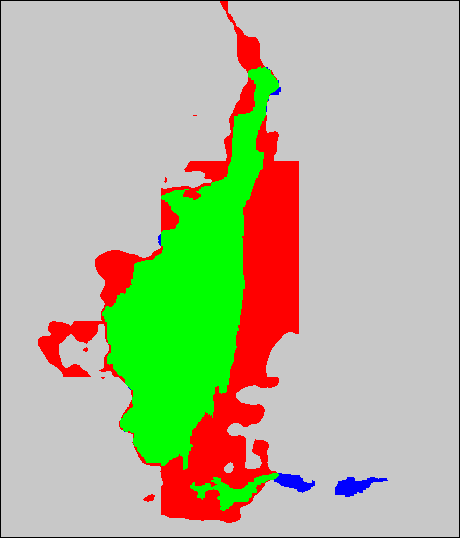}
\end{subfigure}\hfill
\begin{subfigure}[t]{\imgwidth}
    \includegraphics[width=\linewidth]{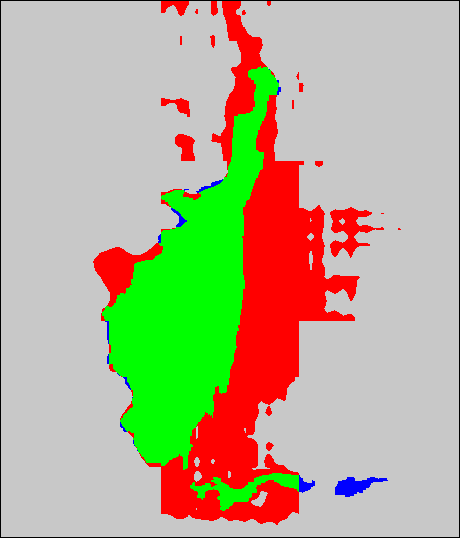}
\end{subfigure}\hfill
\begin{subfigure}[t]{\imgwidth}
    \includegraphics[width=\linewidth]{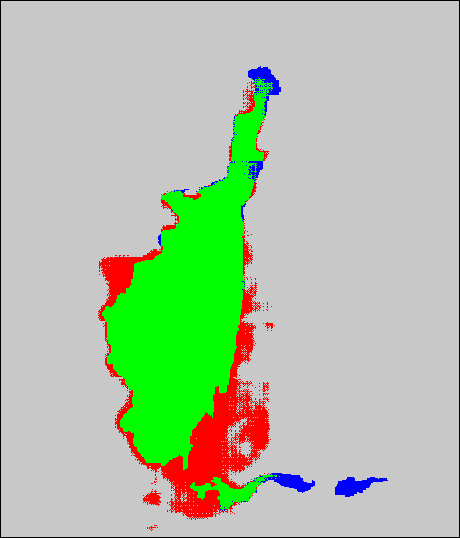}
\end{subfigure}\hfill
\begin{subfigure}[t]{\imgwidth}
    \includegraphics[width=\linewidth]{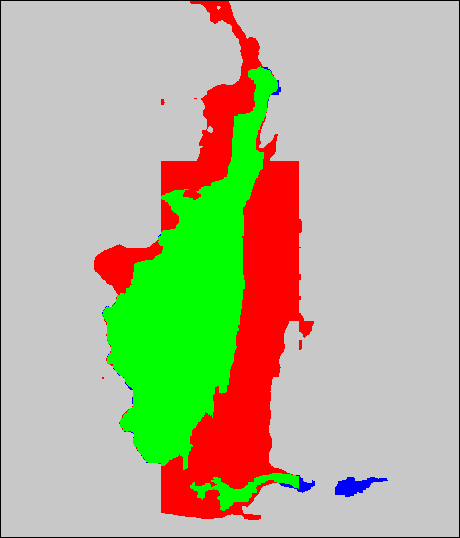}
\end{subfigure}\hfill
\begin{subfigure}[t]{\imgwidth}
    \includegraphics[width=\linewidth]{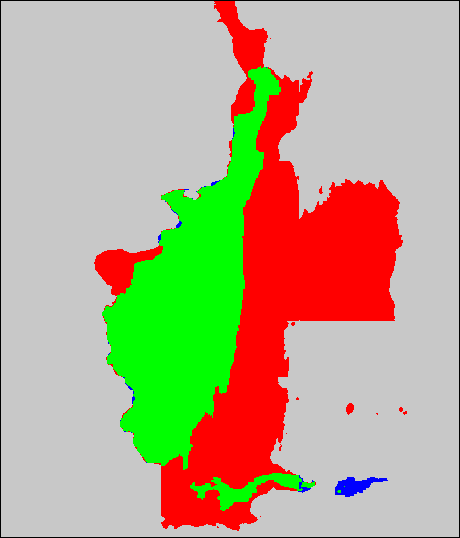}
\end{subfigure}\hfill
\begin{subfigure}[t]{\imgwidth}
    \includegraphics[width=\linewidth]{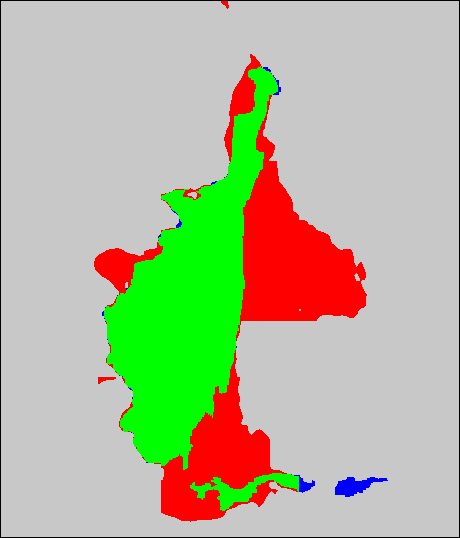}
\end{subfigure}\hfill
\begin{subfigure}[t]{\imgwidth}
    \includegraphics[width=\linewidth]{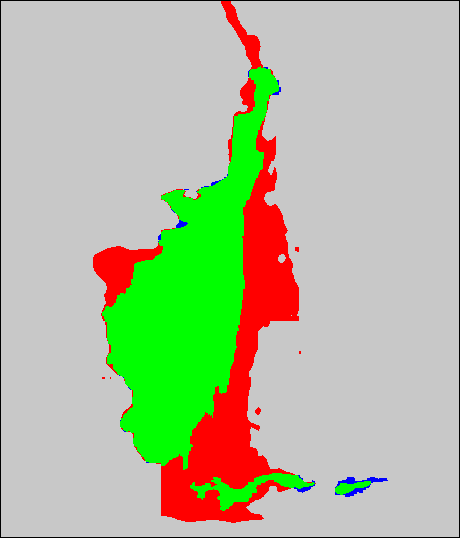}
\end{subfigure}\hfill
\begin{subfigure}[t]{\imgwidth}
    \includegraphics[width=\linewidth]{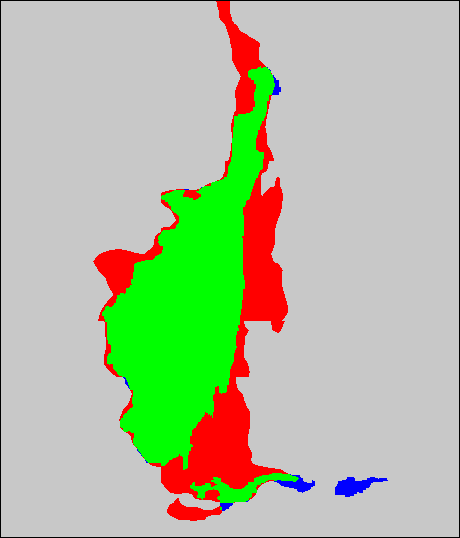}
\end{subfigure}\hfill
\begin{subfigure}[t]{\imgwidth}
    \includegraphics[width=\linewidth]{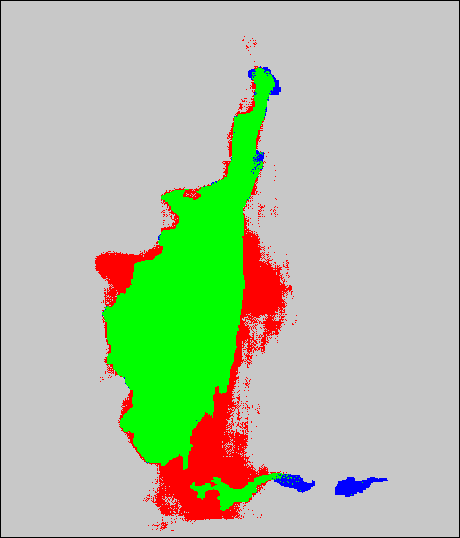}
\end{subfigure}\hfill
\begin{subfigure}[t]{\imgwidth}
    \includegraphics[width=\linewidth]{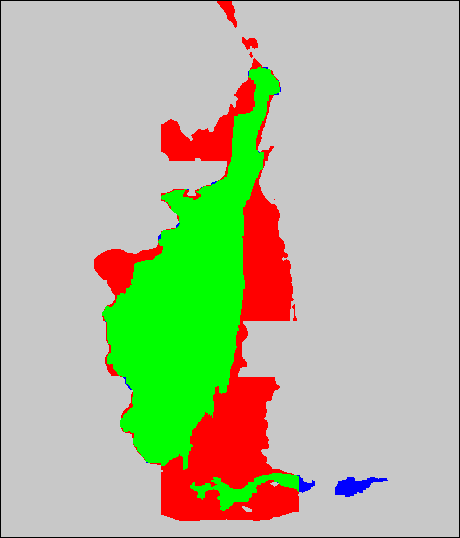}
\end{subfigure}\\[2pt]

\raisebox{0.5\height}{\rowlabel{2019}}\hspace{2pt}%
\begin{subfigure}[t]{\imgwidth}
    \includegraphics[width=\linewidth]{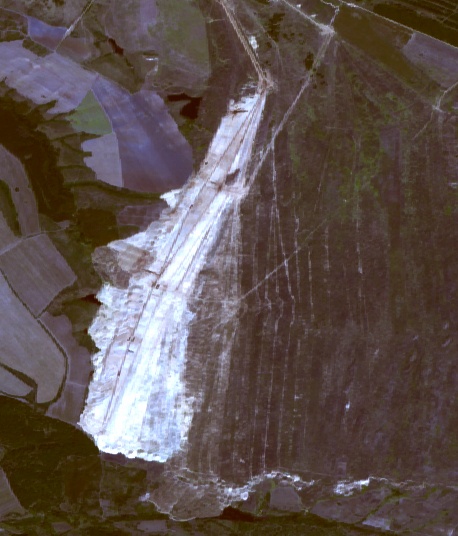}
\end{subfigure}\hfill
\begin{subfigure}[t]{\imgwidth}
    \includegraphics[width=\linewidth]{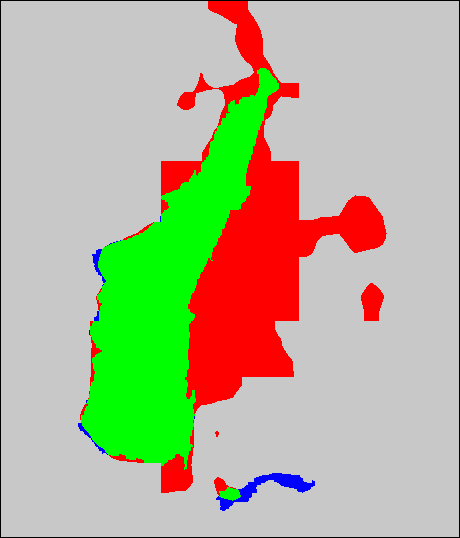}
\end{subfigure}\hfill
\begin{subfigure}[t]{\imgwidth}
    \includegraphics[width=\linewidth]{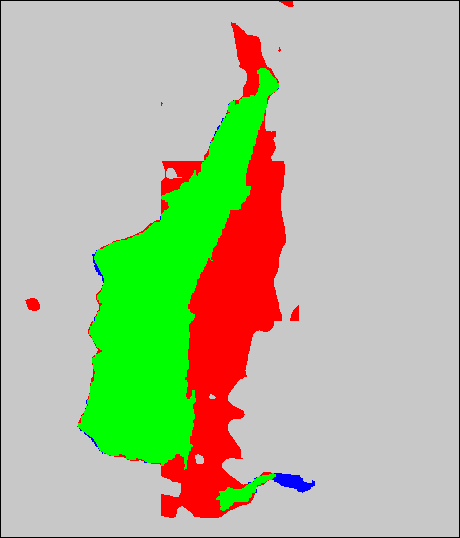}
\end{subfigure}\hfill
\begin{subfigure}[t]{\imgwidth}
    \includegraphics[width=\linewidth]{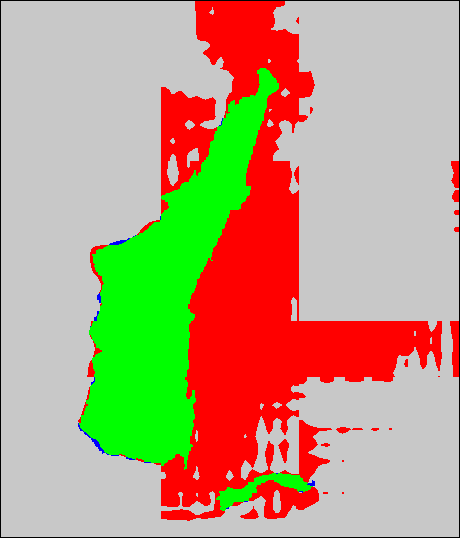}
\end{subfigure}\hfill
\begin{subfigure}[t]{\imgwidth}
    \includegraphics[width=\linewidth]{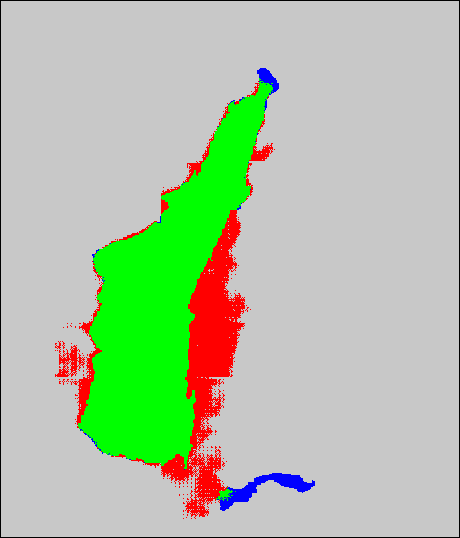}
\end{subfigure}\hfill
\begin{subfigure}[t]{\imgwidth}
    \includegraphics[width=\linewidth]{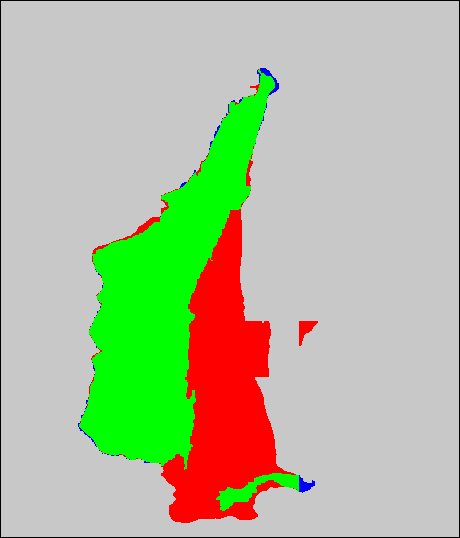}
\end{subfigure}\hfill
\begin{subfigure}[t]{\imgwidth}
    \includegraphics[width=\linewidth]{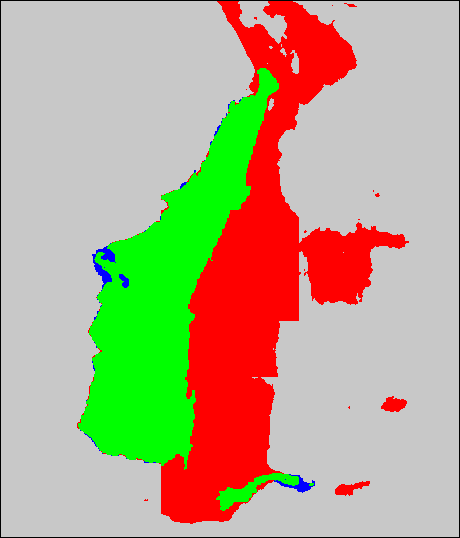}
\end{subfigure}\hfill
\begin{subfigure}[t]{\imgwidth}
    \includegraphics[width=\linewidth]{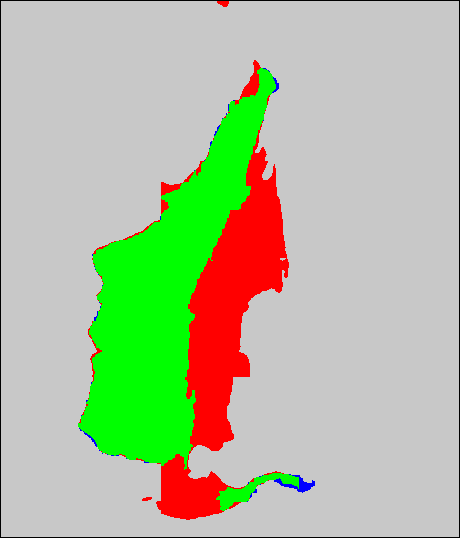}
\end{subfigure}\hfill
\begin{subfigure}[t]{\imgwidth}
    \includegraphics[width=\linewidth]{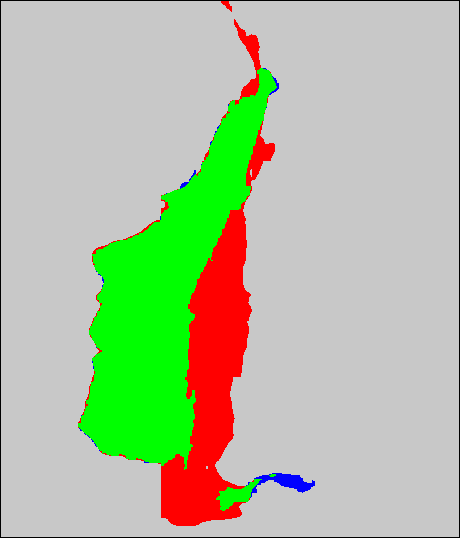}
\end{subfigure}\hfill
\begin{subfigure}[t]{\imgwidth}
    \includegraphics[width=\linewidth]{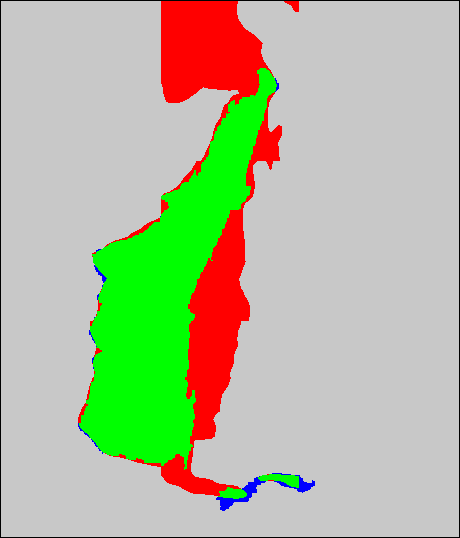}
\end{subfigure}\hfill
\begin{subfigure}[t]{\imgwidth}
    \includegraphics[width=\linewidth]{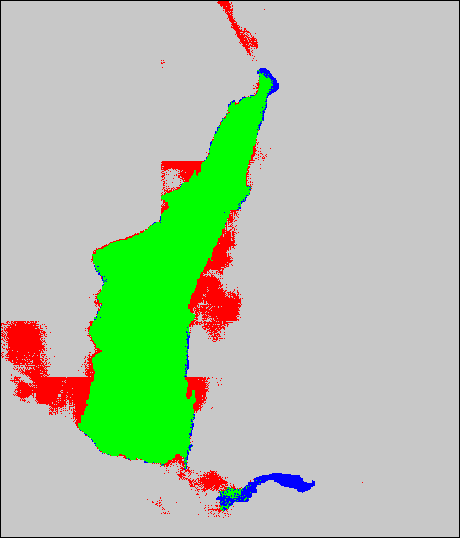}
\end{subfigure}\hfill
\begin{subfigure}[t]{\imgwidth}
    \includegraphics[width=\linewidth]{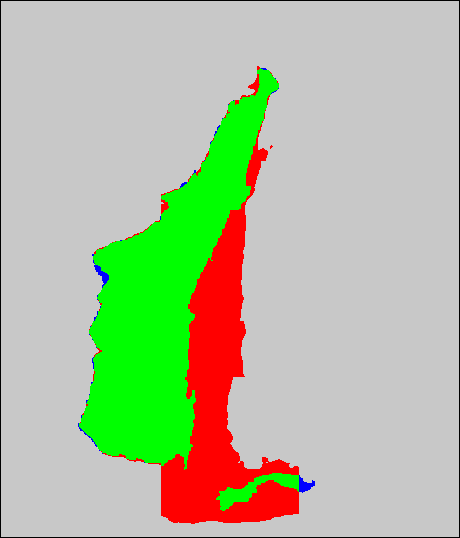}
\end{subfigure}\\[2pt]

\raisebox{0.5\height}{\rowlabel{2020}}\hspace{2pt}%
\begin{subfigure}[t]{\imgwidth}
    \includegraphics[width=\linewidth]{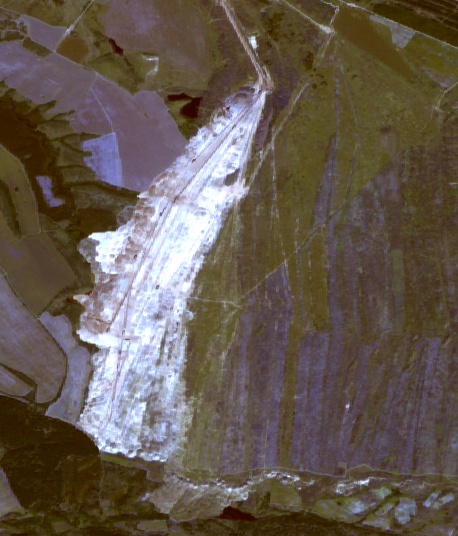}
\end{subfigure}\hfill
\begin{subfigure}[t]{\imgwidth}
    \includegraphics[width=\linewidth]{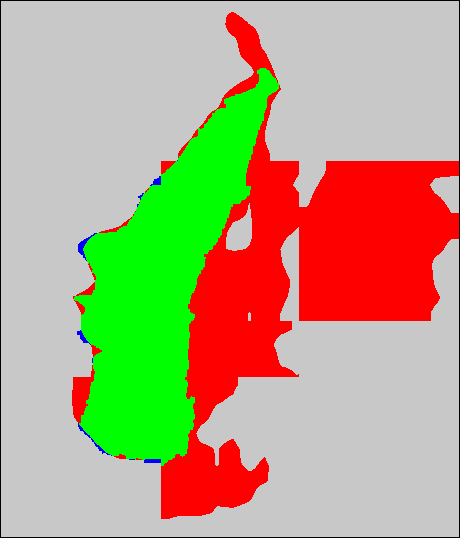}
\end{subfigure}\hfill
\begin{subfigure}[t]{\imgwidth}
    \includegraphics[width=\linewidth]{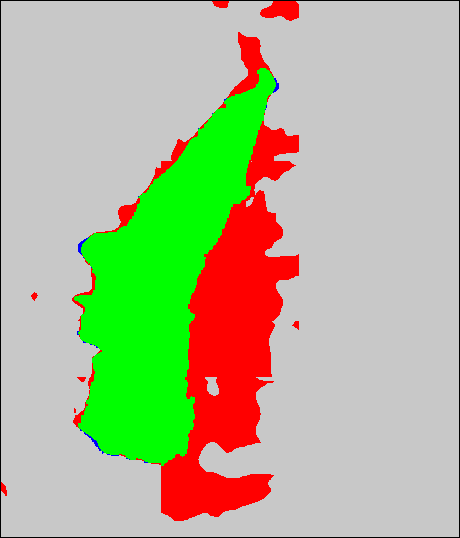}
\end{subfigure}\hfill
\begin{subfigure}[t]{\imgwidth}
    \includegraphics[width=\linewidth]{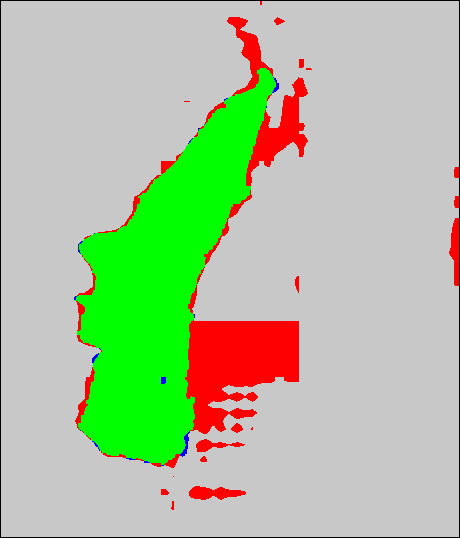}
\end{subfigure}\hfill
\begin{subfigure}[t]{\imgwidth}
    \includegraphics[width=\linewidth]{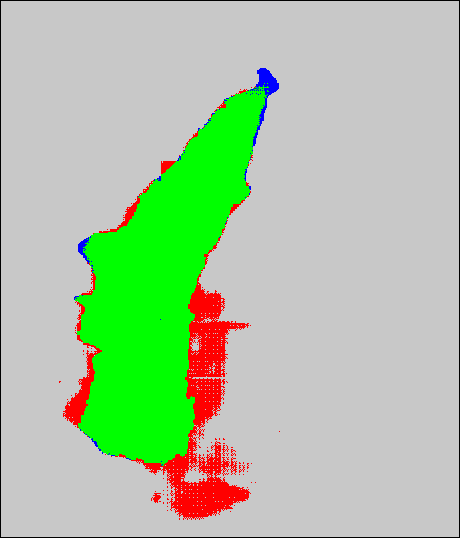}
\end{subfigure}\hfill
\begin{subfigure}[t]{\imgwidth}
    \includegraphics[width=\linewidth]{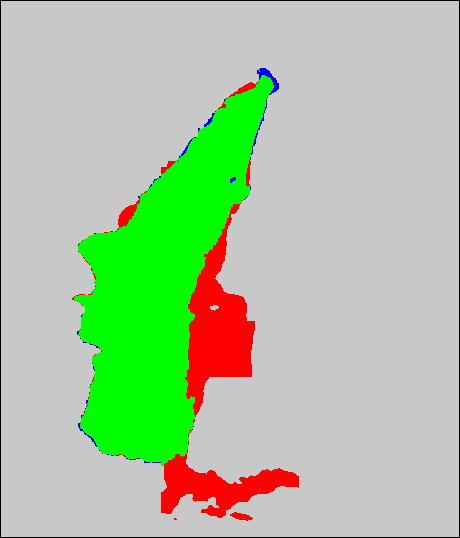}
\end{subfigure}\hfill
\begin{subfigure}[t]{\imgwidth}
    \includegraphics[width=\linewidth]{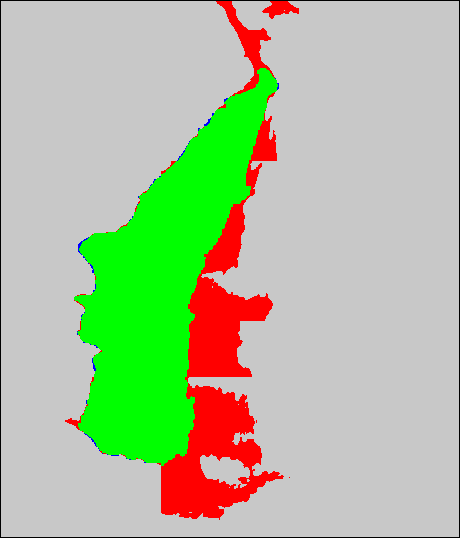}
\end{subfigure}\hfill
\begin{subfigure}[t]{\imgwidth}
    \includegraphics[width=\linewidth]{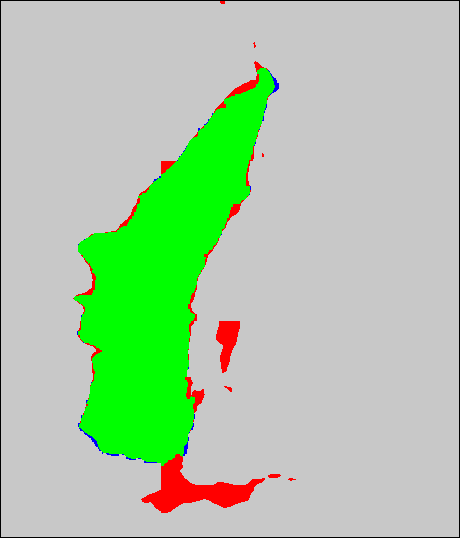}
\end{subfigure}\hfill
\begin{subfigure}[t]{\imgwidth}
    \includegraphics[width=\linewidth]{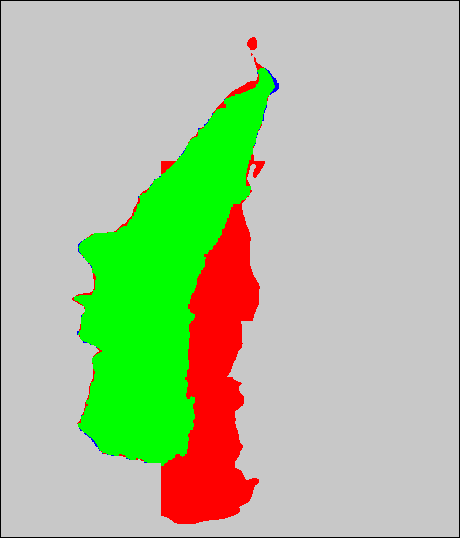}
\end{subfigure}\hfill
\begin{subfigure}[t]{\imgwidth}
    \includegraphics[width=\linewidth]{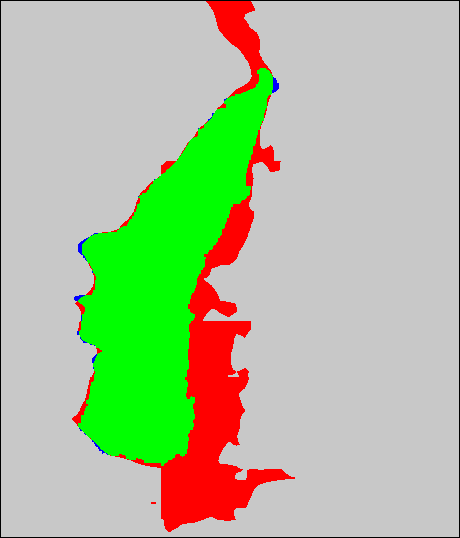}
\end{subfigure}\hfill
\begin{subfigure}[t]{\imgwidth}
    \includegraphics[width=\linewidth]{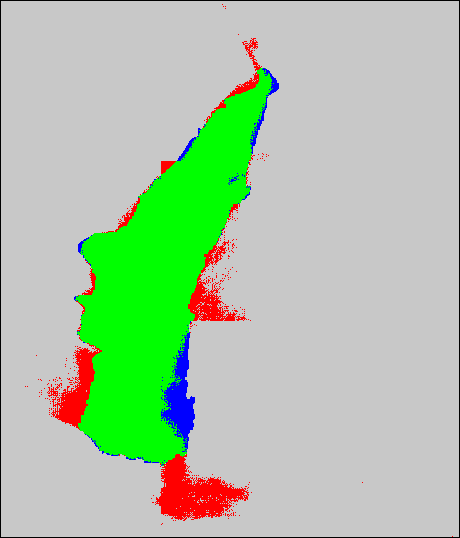}
\end{subfigure}\hfill
\begin{subfigure}[t]{\imgwidth}
    \includegraphics[width=\linewidth]{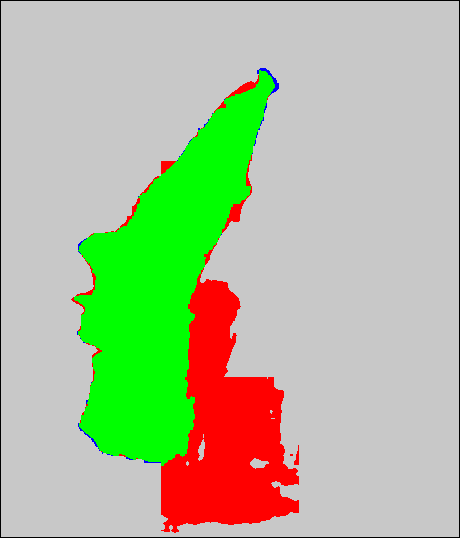}
\end{subfigure}\\[2pt]

\raisebox{0.5\height}{\rowlabel{2021}}\hspace{2pt}%
\begin{subfigure}[t]{\imgwidth}
    \includegraphics[width=\linewidth]{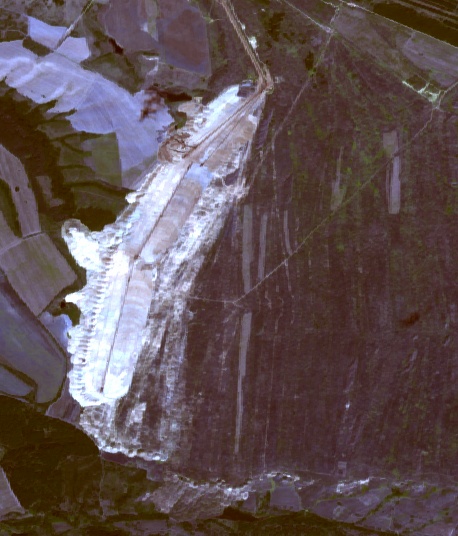}
\end{subfigure}\hfill
\begin{subfigure}[t]{\imgwidth}
    \includegraphics[width=\linewidth]{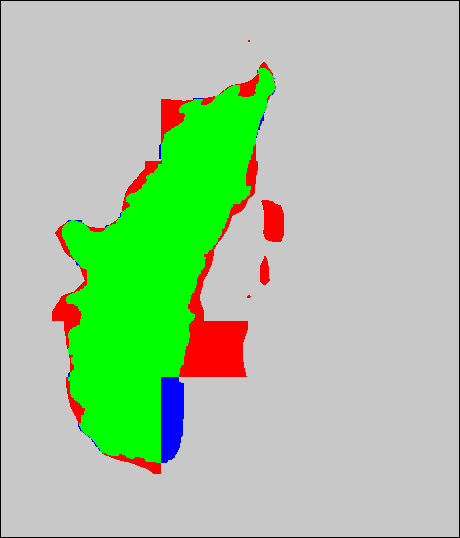}
\end{subfigure}\hfill
\begin{subfigure}[t]{\imgwidth}
    \includegraphics[width=\linewidth]{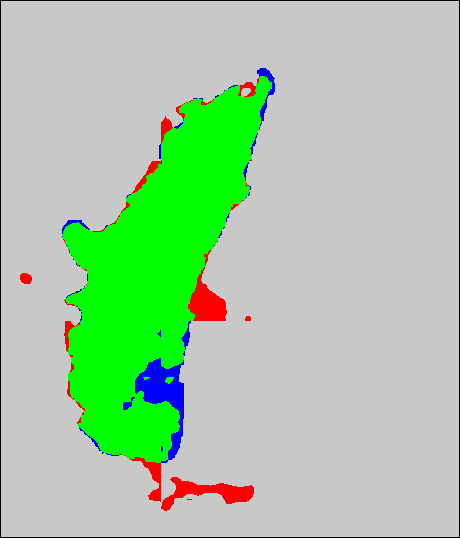}
\end{subfigure}\hfill
\begin{subfigure}[t]{\imgwidth}
    \includegraphics[width=\linewidth]{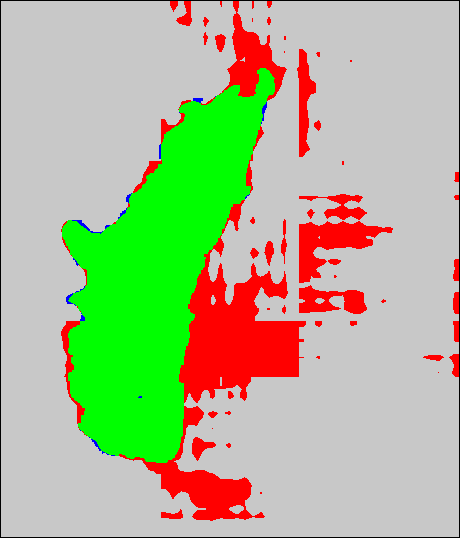}
\end{subfigure}\hfill
\begin{subfigure}[t]{\imgwidth}
    \includegraphics[width=\linewidth]{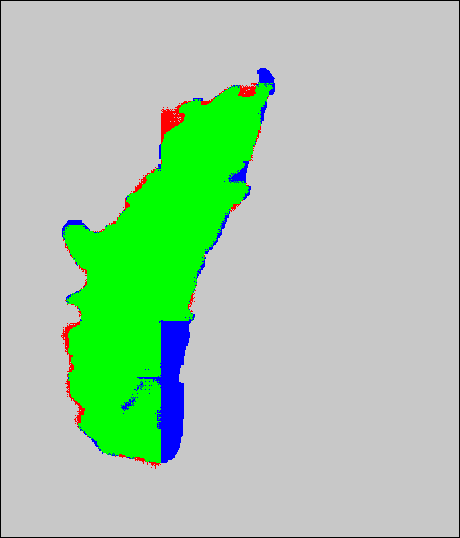}
\end{subfigure}\hfill
\begin{subfigure}[t]{\imgwidth}
    \includegraphics[width=\linewidth]{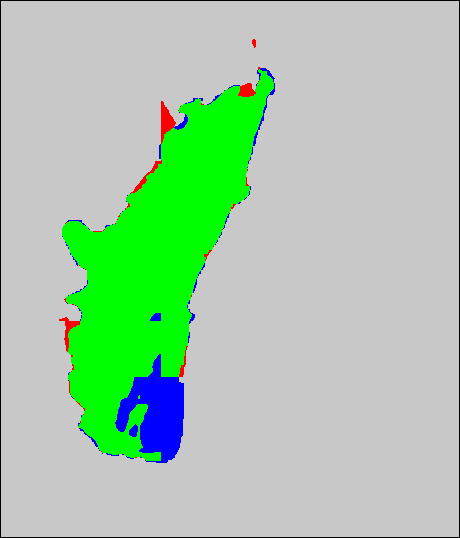}
\end{subfigure}\hfill
\begin{subfigure}[t]{\imgwidth}
    \includegraphics[width=\linewidth]{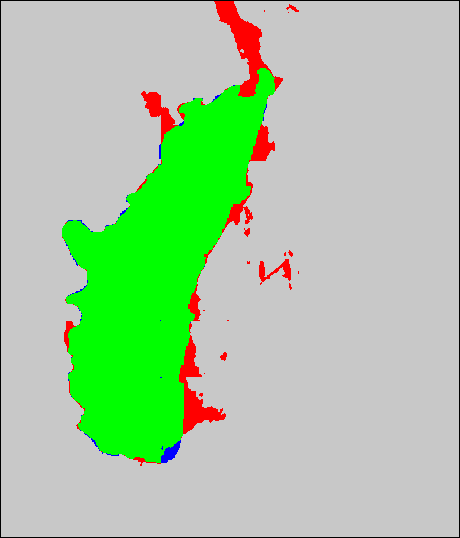}
\end{subfigure}\hfill
\begin{subfigure}[t]{\imgwidth}
    \includegraphics[width=\linewidth]{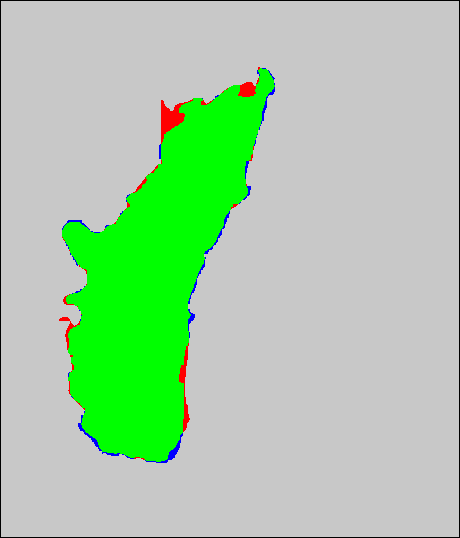}
\end{subfigure}\hfill
\begin{subfigure}[t]{\imgwidth}
    \includegraphics[width=\linewidth]{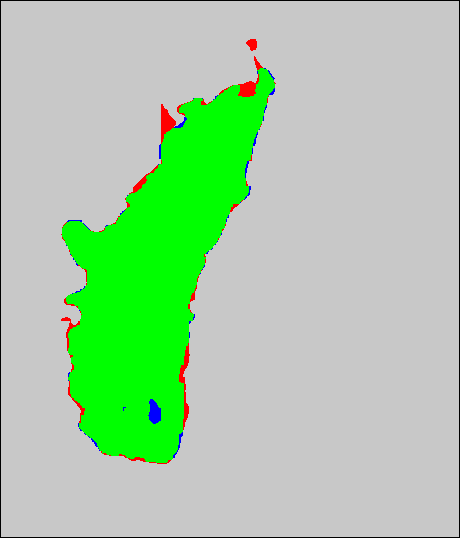}
\end{subfigure}\hfill
\begin{subfigure}[t]{\imgwidth}
    \includegraphics[width=\linewidth]{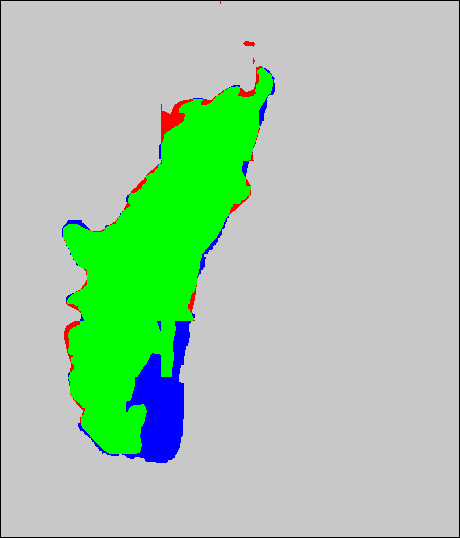}
\end{subfigure}\hfill
\begin{subfigure}[t]{\imgwidth}
    \includegraphics[width=\linewidth]{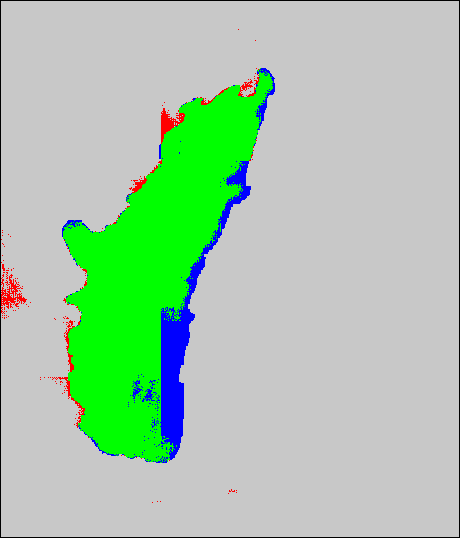}
\end{subfigure}\hfill
\begin{subfigure}[t]{\imgwidth}
    \includegraphics[width=\linewidth]{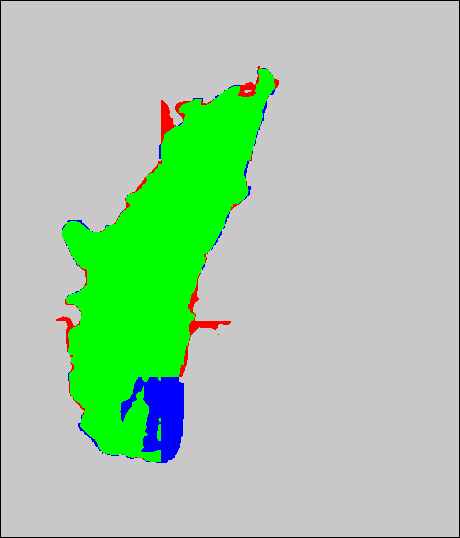}
\end{subfigure}\\[2pt]

\raisebox{0.5\height}{\rowlabel{2022}}\hspace{2pt}%
\begin{subfigure}[t]{\imgwidth}
    \includegraphics[width=\linewidth]{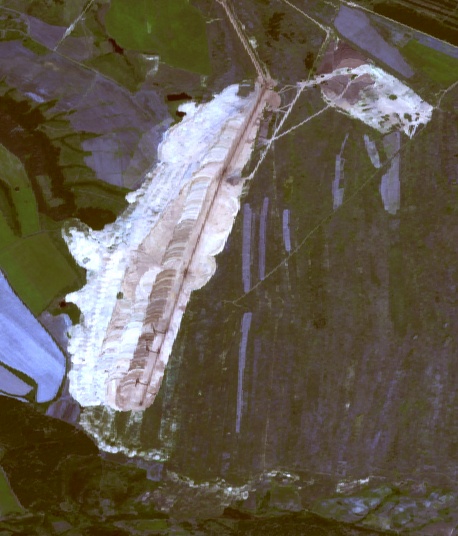}
\end{subfigure}\hfill
\begin{subfigure}[t]{\imgwidth}
    \includegraphics[width=\linewidth]{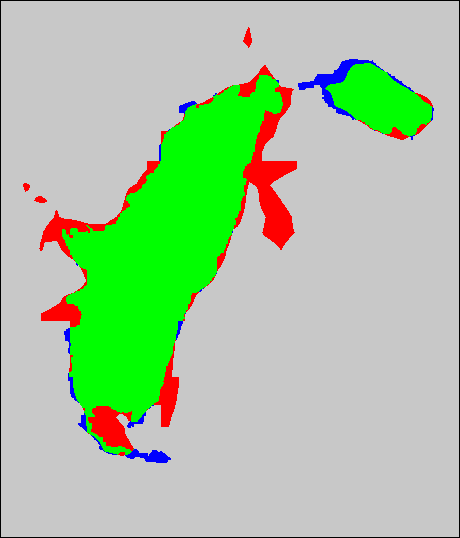}
\end{subfigure}\hfill
\begin{subfigure}[t]{\imgwidth}
    \includegraphics[width=\linewidth]{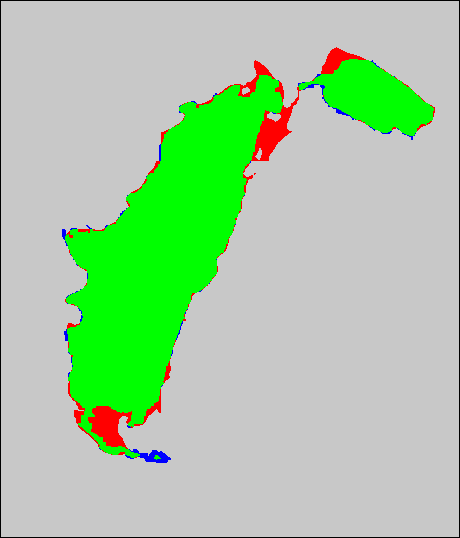}
\end{subfigure}\hfill
\begin{subfigure}[t]{\imgwidth}
    \includegraphics[width=\linewidth]{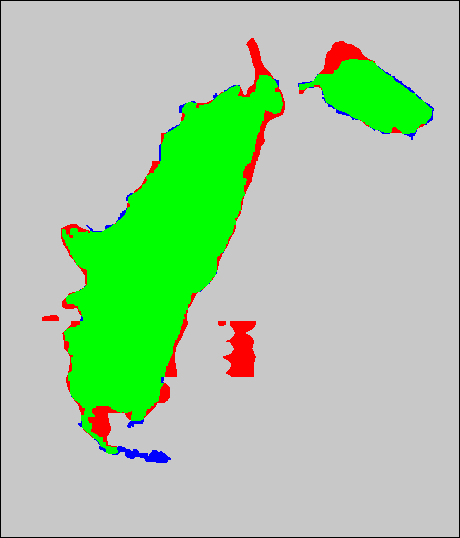}
\end{subfigure}\hfill
\begin{subfigure}[t]{\imgwidth}
    \includegraphics[width=\linewidth]{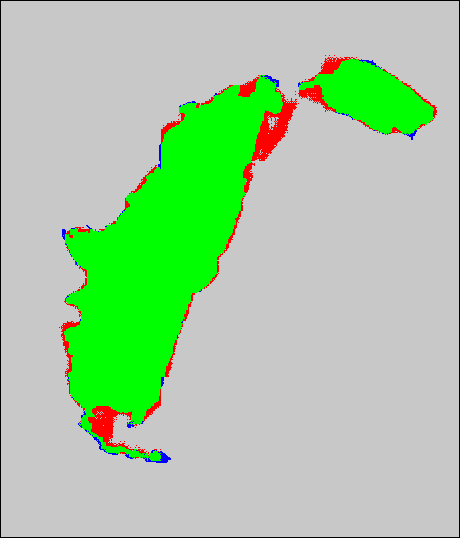}
\end{subfigure}\hfill
\begin{subfigure}[t]{\imgwidth}
    \includegraphics[width=\linewidth]{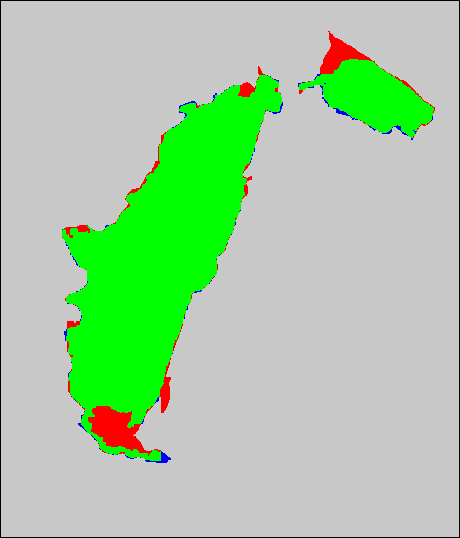}
\end{subfigure}\hfill
\begin{subfigure}[t]{\imgwidth}
    \includegraphics[width=\linewidth]{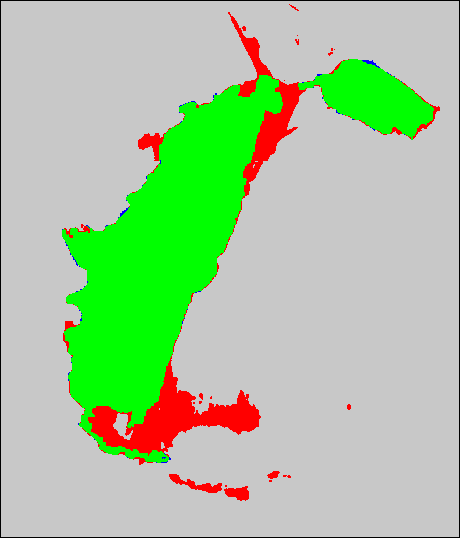}
\end{subfigure}\hfill
\begin{subfigure}[t]{\imgwidth}
    \includegraphics[width=\linewidth]{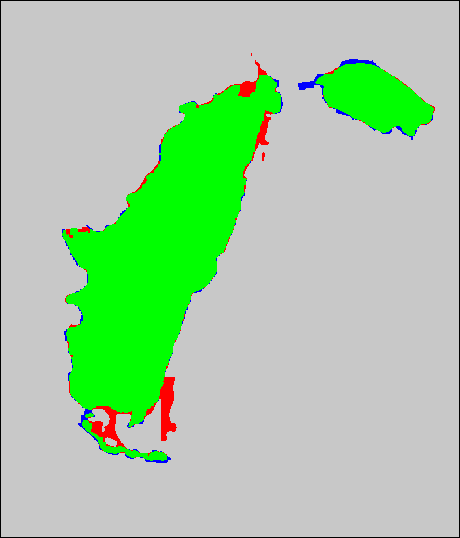}
\end{subfigure}\hfill
\begin{subfigure}[t]{\imgwidth}
    \includegraphics[width=\linewidth]{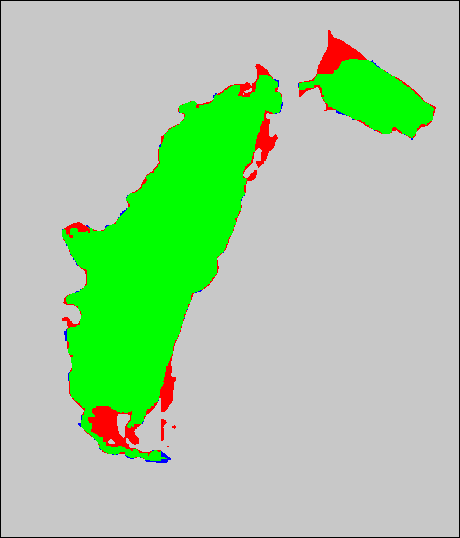}
\end{subfigure}\hfill
\begin{subfigure}[t]{\imgwidth}
    \includegraphics[width=\linewidth]{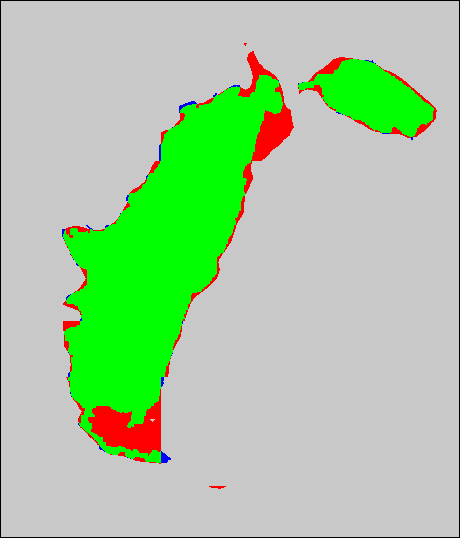}
\end{subfigure}\hfill
\begin{subfigure}[t]{\imgwidth}
    \includegraphics[width=\linewidth]{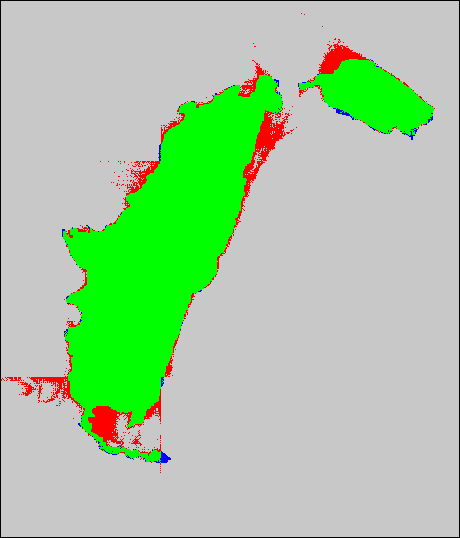}
\end{subfigure}\hfill
\begin{subfigure}[t]{\imgwidth}
    \includegraphics[width=\linewidth]{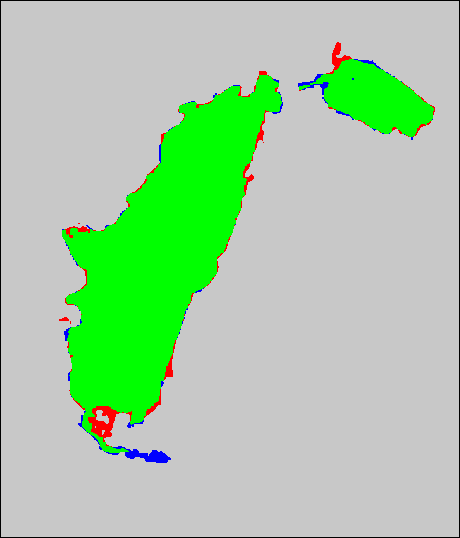}
\end{subfigure}\\[2pt]

\raisebox{0.5\height}{\rowlabel{2023}}\hspace{2pt}%
\begin{subfigure}[t]{\imgwidth}
    \includegraphics[width=\linewidth]{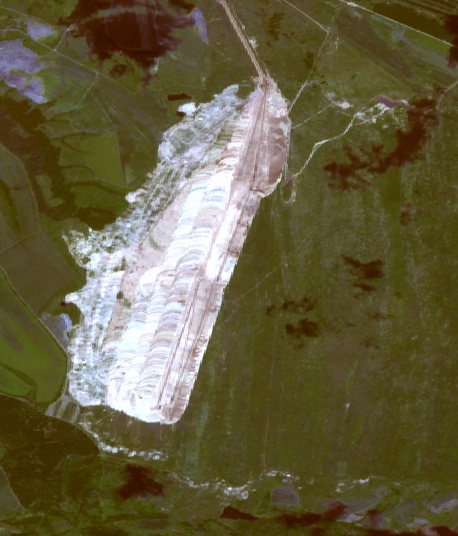}
\end{subfigure}\hfill
\begin{subfigure}[t]{\imgwidth}
    \includegraphics[width=\linewidth]{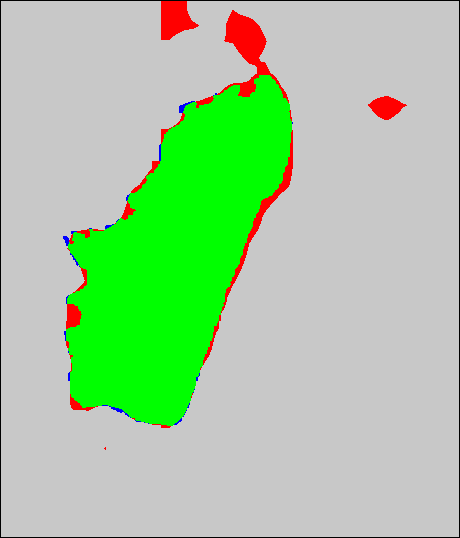}
\end{subfigure}\hfill
\begin{subfigure}[t]{\imgwidth}
    \includegraphics[width=\linewidth]{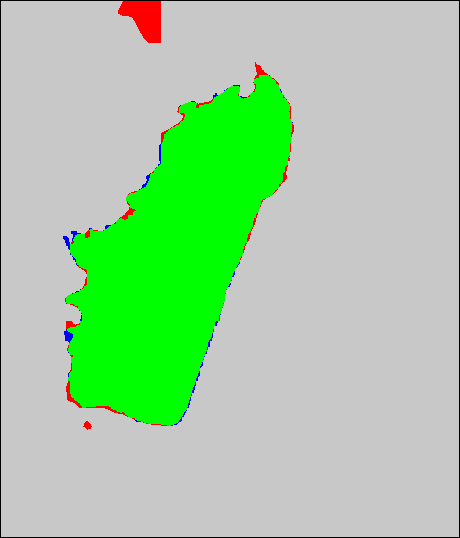}
\end{subfigure}\hfill
\begin{subfigure}[t]{\imgwidth}
    \includegraphics[width=\linewidth]{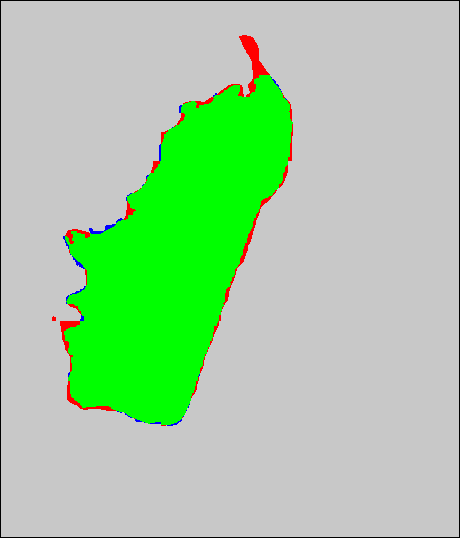}
\end{subfigure}\hfill
\begin{subfigure}[t]{\imgwidth}
    \includegraphics[width=\linewidth]{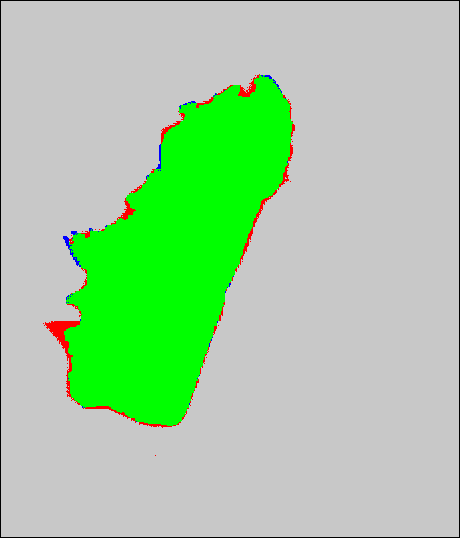}
\end{subfigure}\hfill
\begin{subfigure}[t]{\imgwidth}
    \includegraphics[width=\linewidth]{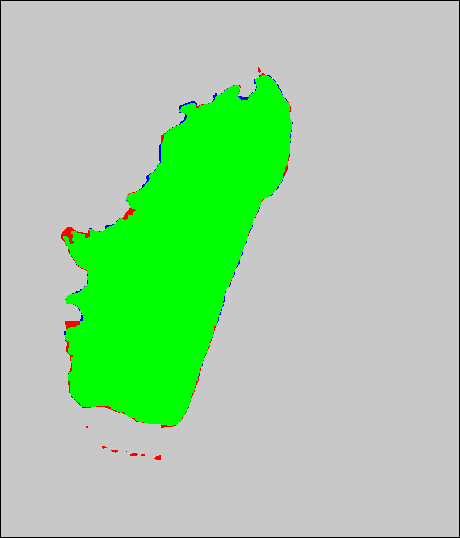}
\end{subfigure}\hfill
\begin{subfigure}[t]{\imgwidth}
    \includegraphics[width=\linewidth]{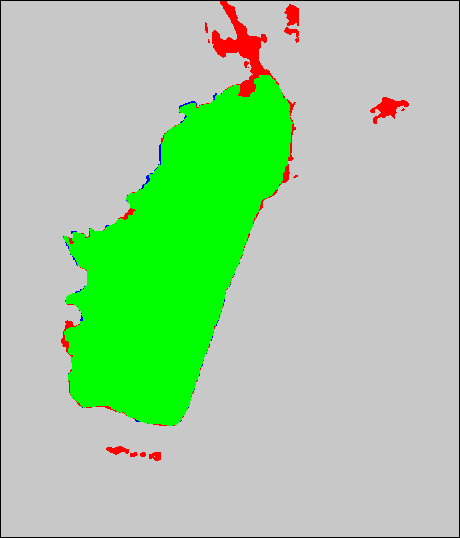}
\end{subfigure}\hfill
\begin{subfigure}[t]{\imgwidth}
    \includegraphics[width=\linewidth]{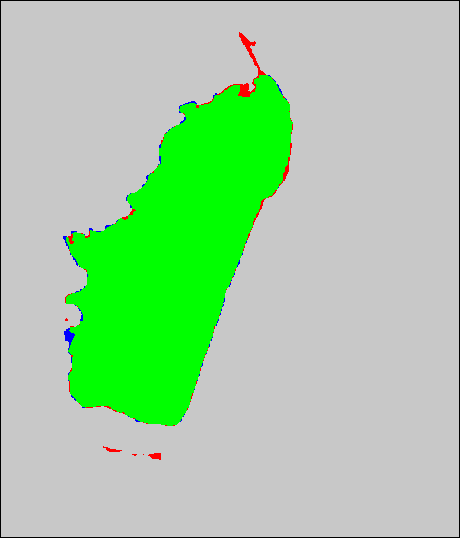}
\end{subfigure}\hfill
\begin{subfigure}[t]{\imgwidth}
    \includegraphics[width=\linewidth]{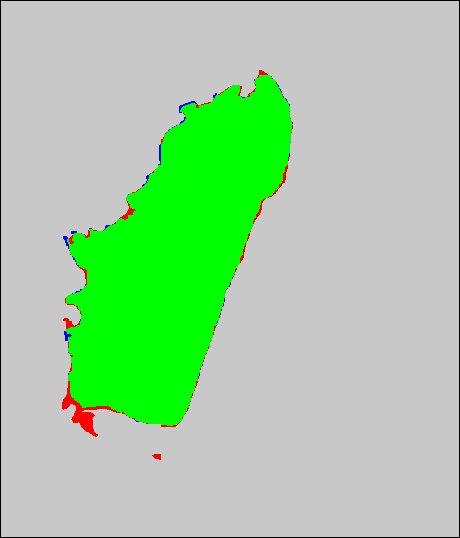}
\end{subfigure}\hfill
\begin{subfigure}[t]{\imgwidth}
    \includegraphics[width=\linewidth]{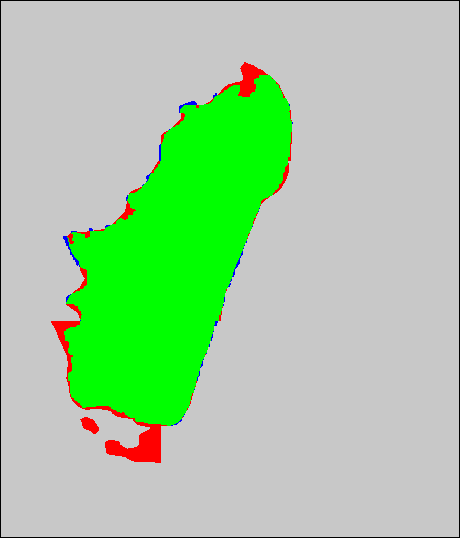}
\end{subfigure}\hfill
\begin{subfigure}[t]{\imgwidth}
    \includegraphics[width=\linewidth]{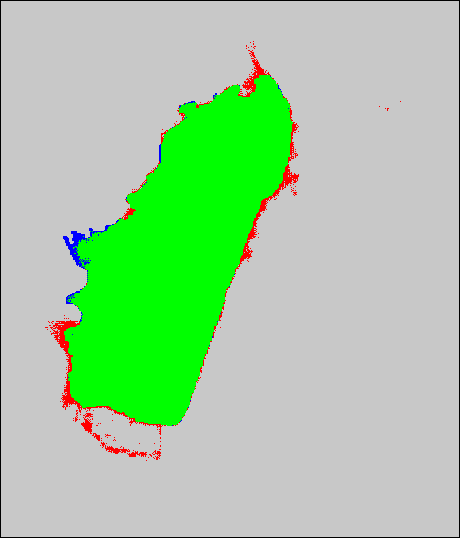}
\end{subfigure}\hfill
\begin{subfigure}[t]{\imgwidth}
    \includegraphics[width=\linewidth]{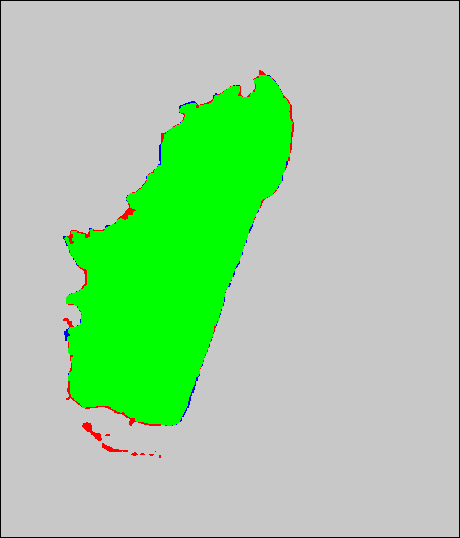}
\end{subfigure}\\[2pt]

\raisebox{0.5\height}{\rowlabel{2024}}\hspace{2pt}%
\begin{subfigure}[t]{\imgwidth}
    \includegraphics[width=\linewidth]{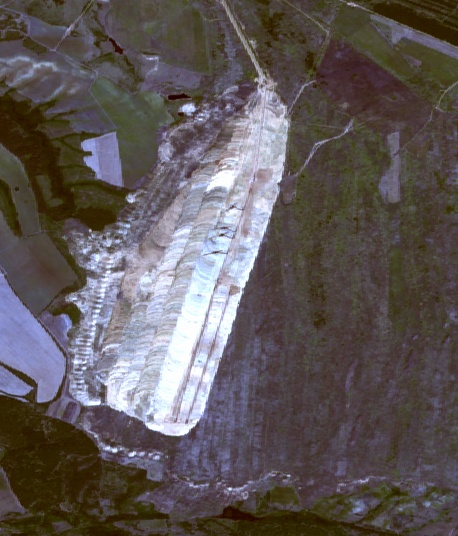}
    \caption{}
\end{subfigure}\hfill
\begin{subfigure}[t]{\imgwidth}
    \includegraphics[width=\linewidth]{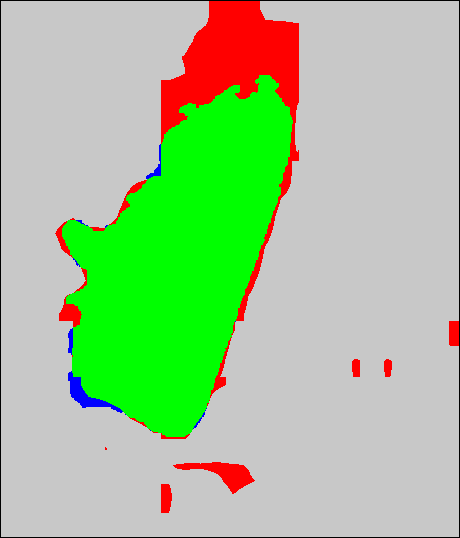}
    \caption{}
\end{subfigure}\hfill
\begin{subfigure}[t]{\imgwidth}
    \includegraphics[width=\linewidth]{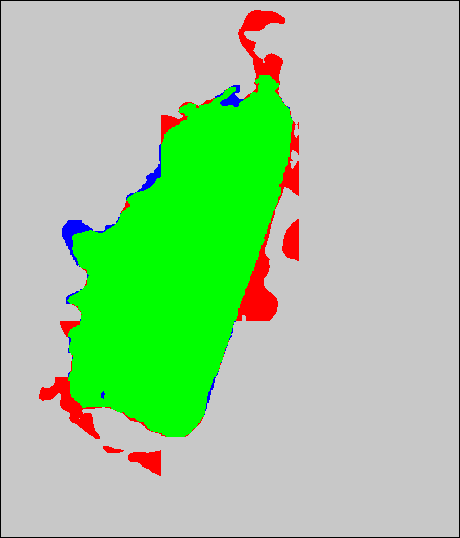}
    \caption{}
\end{subfigure}\hfill
\begin{subfigure}[t]{\imgwidth}
    \includegraphics[width=\linewidth]{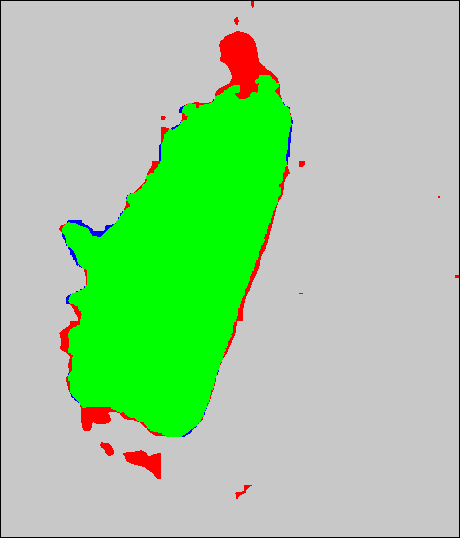}
    \caption{}
\end{subfigure}\hfill
\begin{subfigure}[t]{\imgwidth}
    \includegraphics[width=\linewidth]{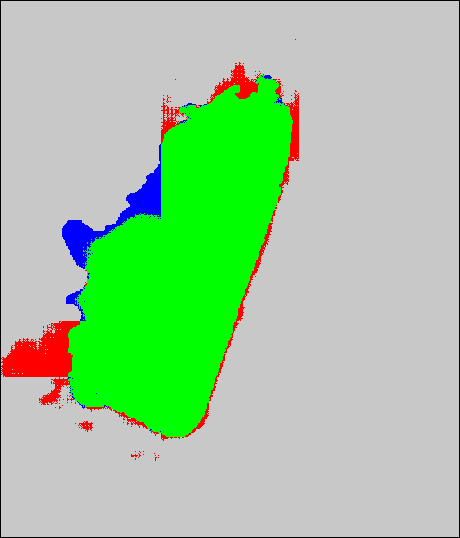}
    \caption{}
\end{subfigure}\hfill
\begin{subfigure}[t]{\imgwidth}
    \includegraphics[width=\linewidth]{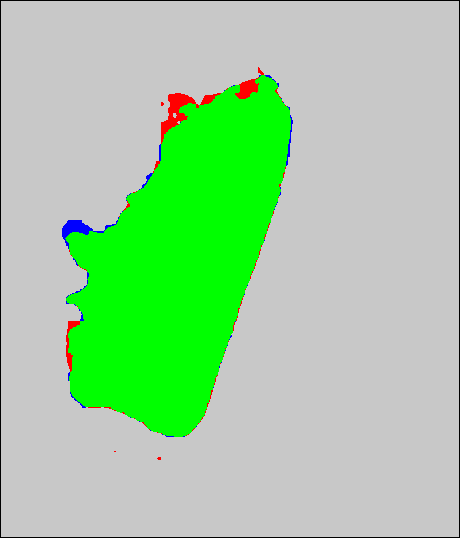}
    \caption{}
\end{subfigure}\hfill
\begin{subfigure}[t]{\imgwidth}
    \includegraphics[width=\linewidth]{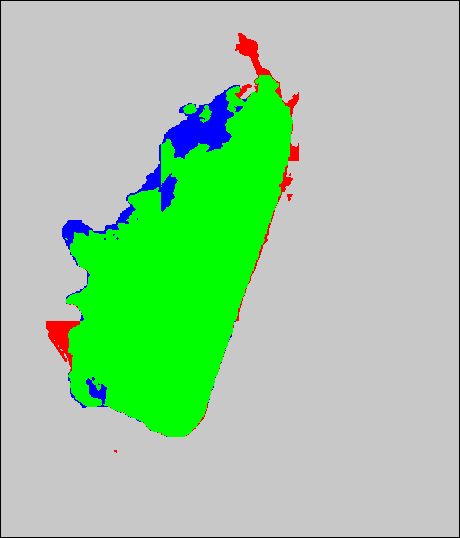}
    \caption{}
\end{subfigure}\hfill
\begin{subfigure}[t]{\imgwidth}
    \includegraphics[width=\linewidth]{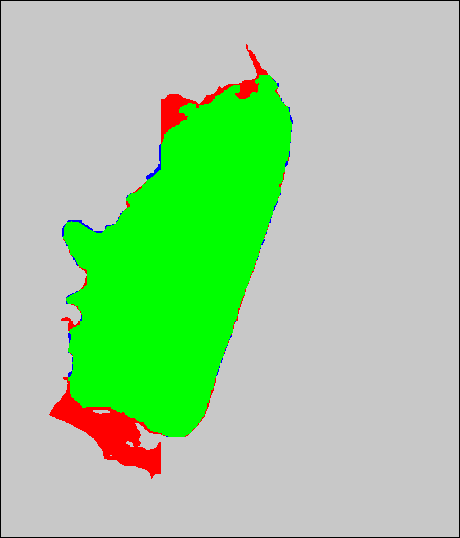}
    \caption{}
\end{subfigure}\hfill
\begin{subfigure}[t]{\imgwidth}
    \includegraphics[width=\linewidth]{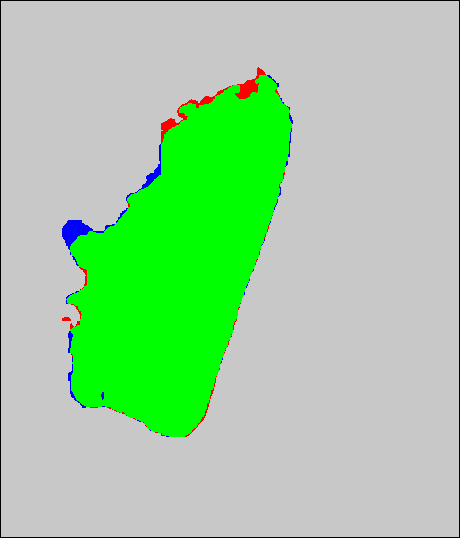}
    \caption{}
\end{subfigure}\hfill
\begin{subfigure}[t]{\imgwidth}
    \includegraphics[width=\linewidth]{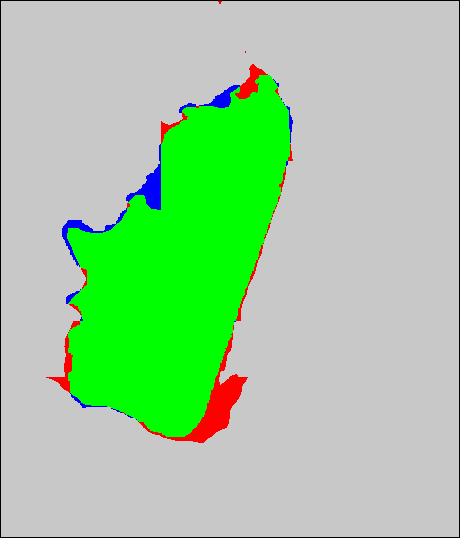}
    \caption{}
\end{subfigure}\hfill
\begin{subfigure}[t]{\imgwidth}
    \includegraphics[width=\linewidth]{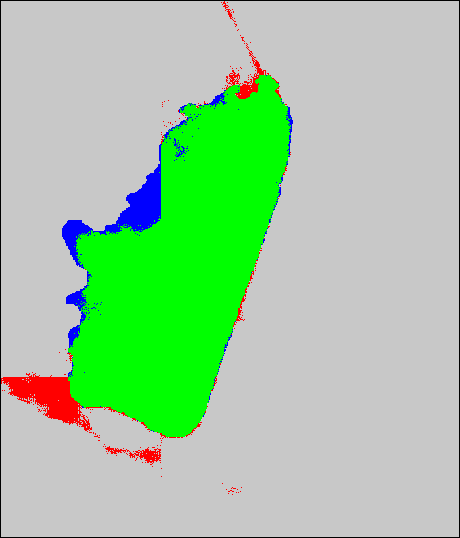}
    \caption{}
\end{subfigure}\hfill
\begin{subfigure}[t]{\imgwidth}
    \includegraphics[width=\linewidth]{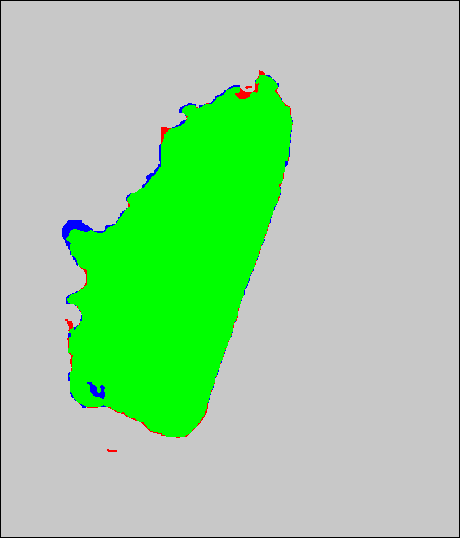}
    \caption{}
\end{subfigure}\\[2pt]

\caption{Qualitative comparison for multitemporal mining footprint mapping, demonstrated by the site Maristsa Iztok Complex, Bulgaria. The pixels of TP, TN, FP, and FN are indicated in green, grey, red, and blue, respectively. (a) Data input (only RGB bands are demonstrated for visualization) (b) DeeplabV3 (MobileViTV2) (c) DeeplabV3P (ResNet-101) (d) DeeplabV3P (MobileNetV2) (e) LinkNet (f) Mask2Former (SwinT-B) (g) UNet (h) UperNet (ConvNext-B5) (i) UperNet (SwinT-B) (j) PSPNet (k) SQNet (l) Segformer}
\label{fig:multi_mining_map}
\end{figure*}

The quantitative results in Table \ref{tab:multi_map_result} and qualitative results in Fig. \ref{fig:multi_mining_map} demonstrate notable variations in performance across both years and model architectures. In terms of overall F1-score (OF1), UperNet (SwinT-B) achieves the highest score (0.8558), followed closely by Segformer (0.8511) and Mask2Former (SwinT-B) (0.8480). These results indicate that transformer-based backbones generally provide stronger temporal consistency and spatial discrimination for mining footprint mapping compared to traditional CNN-based designs. Models such as SQNet and PSPNet lag behind, suggesting that shallower or less context-aware architectures struggle with the heterogeneous characteristics of multi-temporal mining scenes.

Yearly performance trends reveal a relatively stable mapping accuracy, with most models showing only minor fluctuations across different time intervals. This stability reflects the robustness of modern segmentation architectures in handling moderate temporal variations in remote sensing imagery. Nonetheless, performance dips are observed in certain intermediate years (e.g., 2017 for several models), potentially due to challenging seasonal or atmospheric conditions in those image sets.

The temporal consistency metrics, GCA-TIoU and LCA-TIoU, provide additional insights beyond per-year mapping accuracy. Here, UperNet (SwinT-B) again outperforms all other methods, achieving the highest GCA-TIoU (0.8033) and LCA-TIoU (0.8400), indicating that it maintains the most consistent predictions across both global and local unchanged areas over time. Segformer ranks second in both metrics, reinforcing its strength in temporal stability. Interestingly, some models with competitive OF1 scores, such as UNet and UperNet (ConvNext-B5), show slightly lower CA-TIoU values, implying that high per-year accuracy does not always translate to strong multi-year temporal coherence.

Overall, these results suggest that transformer-based architectures not only enhance spatial segmentation accuracy but also improve temporal consistency in long-term mining footprint monitoring. The CA-TIoU analysis proves valuable in identifying models that deliver stable change patterns over time, which is critical for applications where temporal reliability is as important as per-epoch accuracy.

\begin{figure*}[htbp]
\captionsetup[subfigure]{labelformat=empty}
\centering
\newcommand{\imgwidthI}{0.093\textwidth}
\newcommand{\imgwidth}{0.105\textwidth}

\newcommand{\rowlabel}[1]{%
  \begin{rotate}{90}%
     \hspace{.4em} \footnotesize #1%
  \end{rotate}%
}
\raisebox{0.5\height}{\rowlabel{\scriptsize (a)}}\hspace{2pt}%
\begin{subfigure}[t]{\imgwidthI}
    \includegraphics[width=\linewidth]{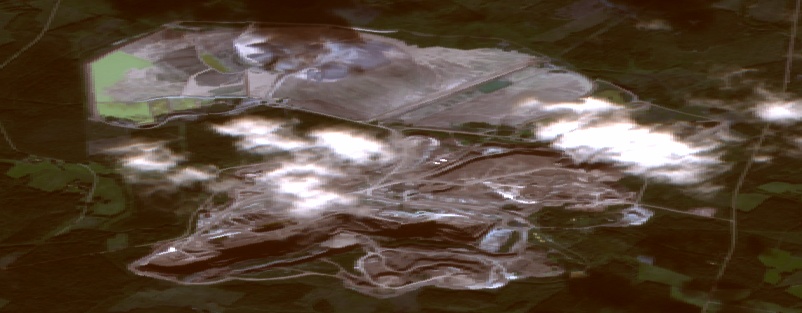}
\end{subfigure}\hfill
\begin{subfigure}[t]{\imgwidthI}
    \includegraphics[width=\linewidth]{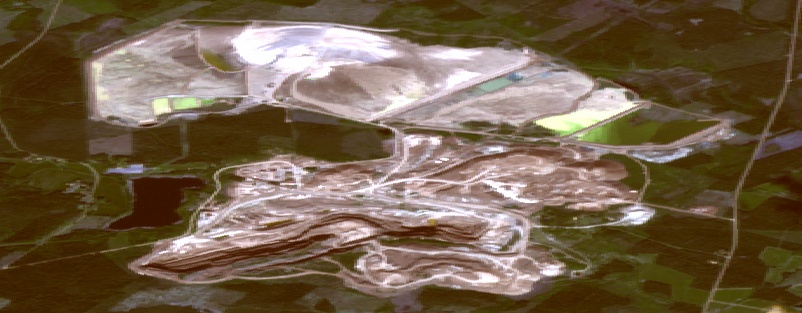}
\end{subfigure}\hfill
\begin{subfigure}[t]{\imgwidthI}
    \includegraphics[width=\linewidth]{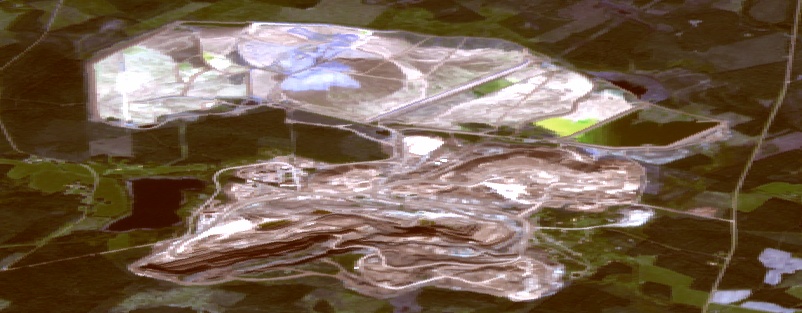}
\end{subfigure}\hfill
\begin{subfigure}[t]{\imgwidthI}
    \includegraphics[width=\linewidth]{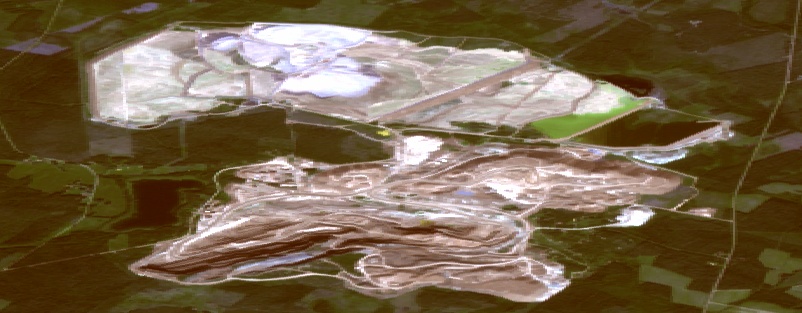}
\end{subfigure}\hfill
\begin{subfigure}[t]{\imgwidthI}
    \includegraphics[width=\linewidth]{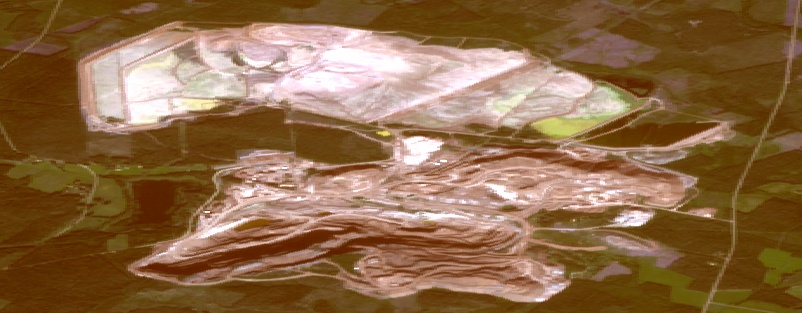}
\end{subfigure}\hfill
\begin{subfigure}[t]{\imgwidthI}
    \includegraphics[width=\linewidth]{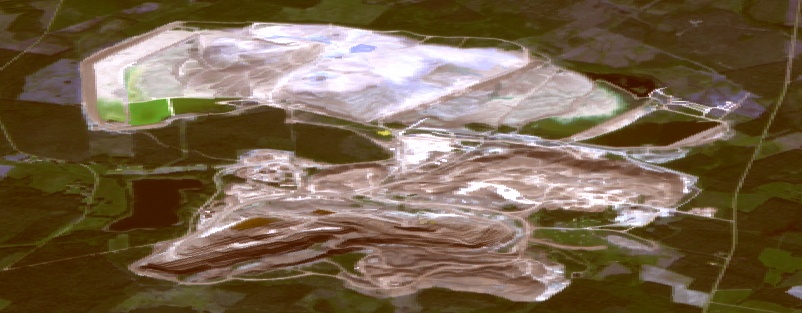}
\end{subfigure}\hfill
\begin{subfigure}[t]{\imgwidthI}
    \includegraphics[width=\linewidth]{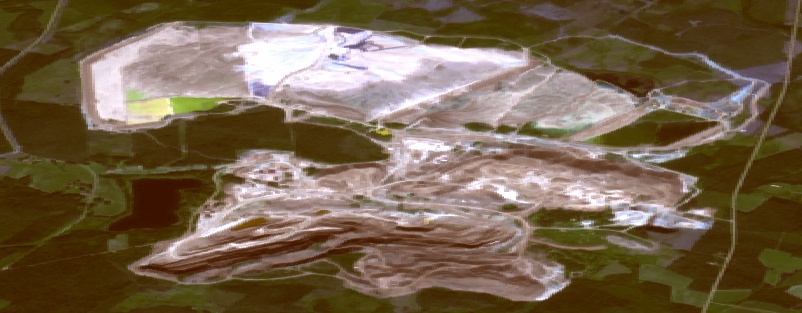}
\end{subfigure}\hfill
\begin{subfigure}[t]{\imgwidthI}
    \includegraphics[width=\linewidth]{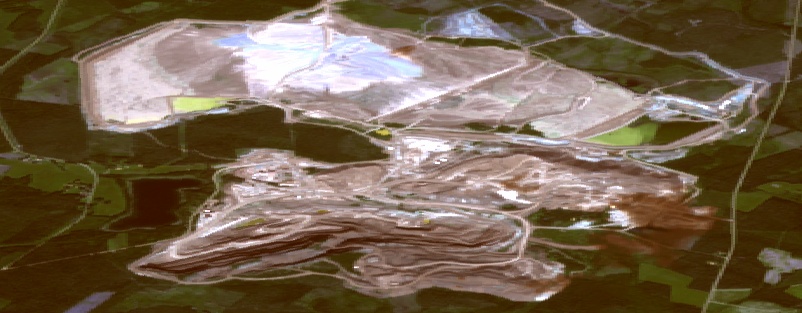}
\end{subfigure}\hfill
\begin{subfigure}[t]{\imgwidthI}
    \includegraphics[width=\linewidth]{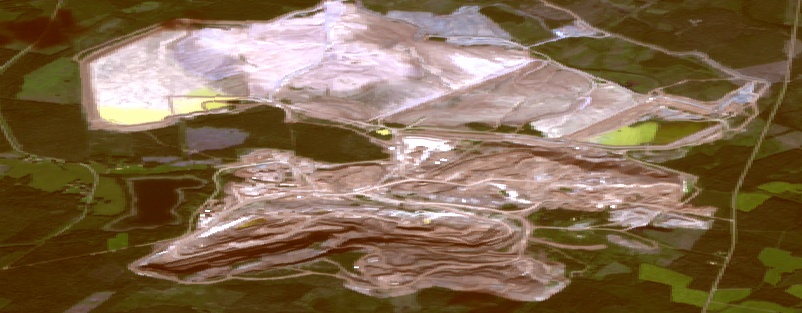}
\end{subfigure}\hfill
\begin{subfigure}[t]{\imgwidthI}
    \includegraphics[width=\linewidth]{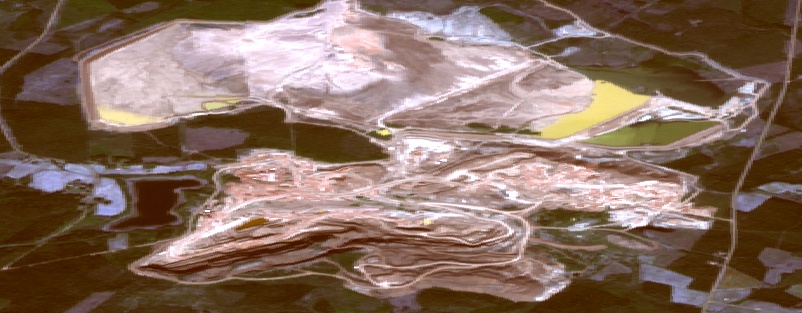}
\end{subfigure}\\[2pt]
\raisebox{0.5\height}{\rowlabel{\scriptsize (b)}}\hspace{2pt}%
\begin{subfigure}[t]{\imgwidth}
    \includegraphics[width=\linewidth]{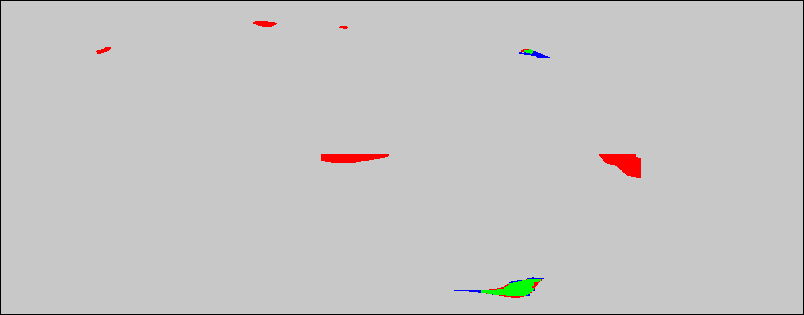}
\end{subfigure}\hfill
\begin{subfigure}[t]{\imgwidth}
    \includegraphics[width=\linewidth]{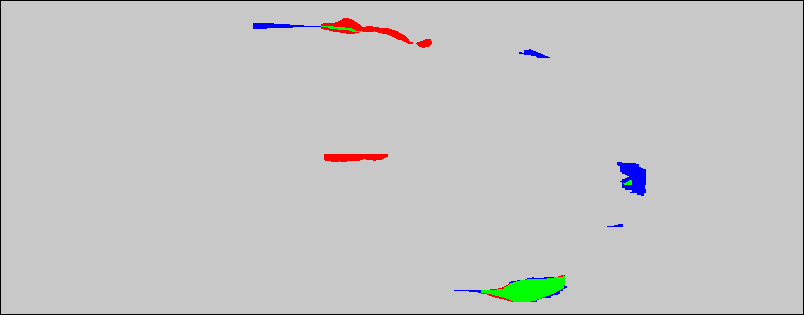}
\end{subfigure}\hfill
\begin{subfigure}[t]{\imgwidth}
    \includegraphics[width=\linewidth]{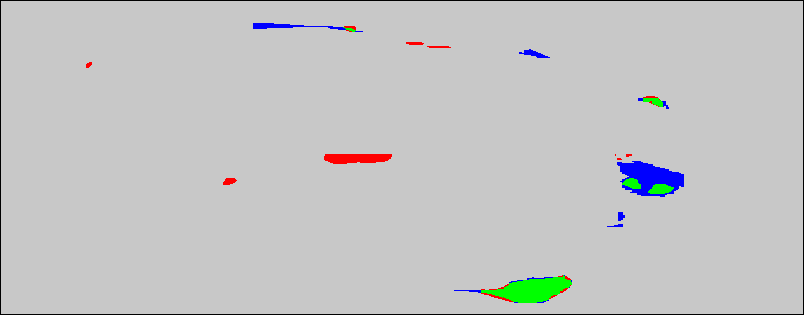}
\end{subfigure}\hfill
\begin{subfigure}[t]{\imgwidth}
    \includegraphics[width=\linewidth]{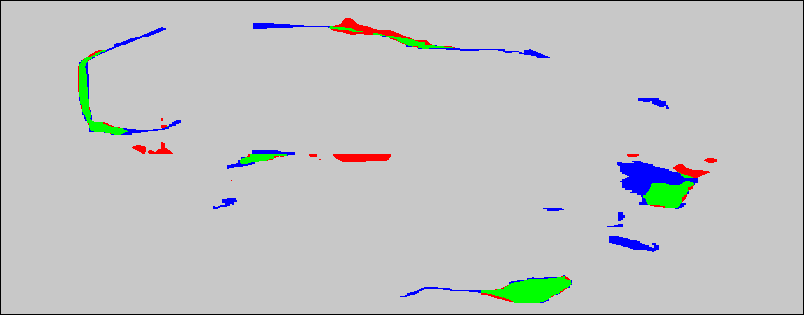}
\end{subfigure}\hfill
\begin{subfigure}[t]{\imgwidth}
    \includegraphics[width=\linewidth]{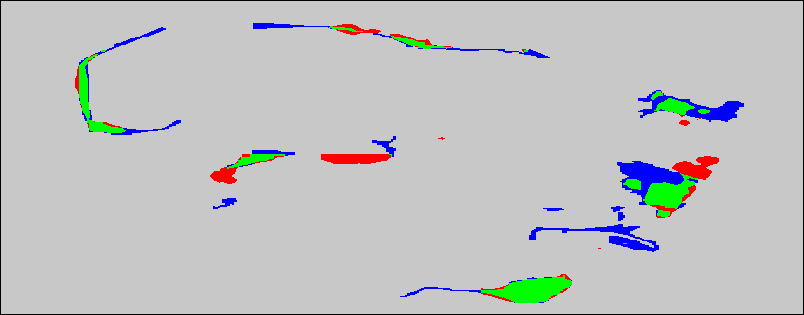}
\end{subfigure}\hfill
\begin{subfigure}[t]{\imgwidth}
    \includegraphics[width=\linewidth]{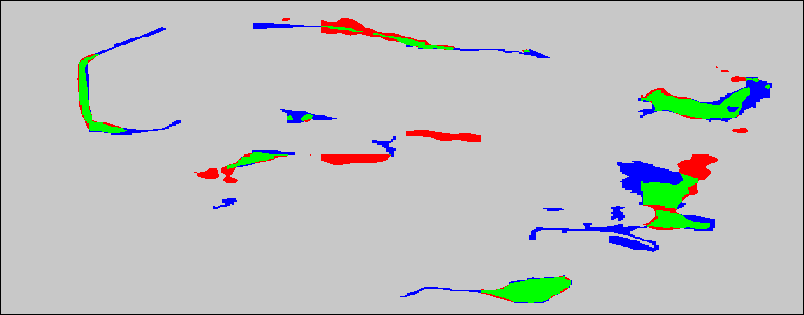}
\end{subfigure}\hfill
\begin{subfigure}[t]{\imgwidth}
    \includegraphics[width=\linewidth]{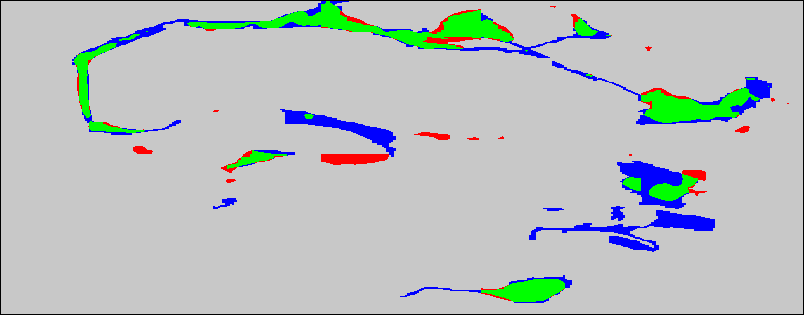}
\end{subfigure}\hfill
\begin{subfigure}[t]{\imgwidth}
    \includegraphics[width=\linewidth]{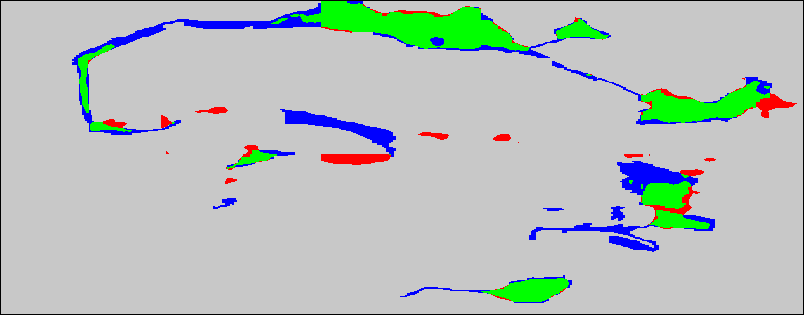}
\end{subfigure}\hfill
\begin{subfigure}[t]{\imgwidth}
    \includegraphics[width=\linewidth]{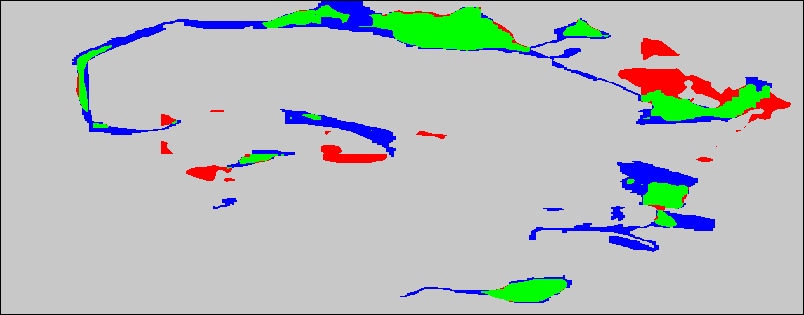}
\end{subfigure}\hfill
\\[2pt]

\raisebox{0.5\height}{\rowlabel{\scriptsize (c)}}\hspace{2pt}%
\begin{subfigure}[t]{\imgwidth}
    \includegraphics[width=\linewidth]{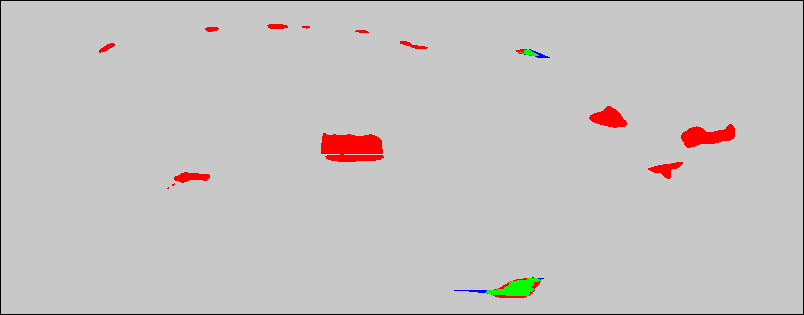}
\end{subfigure}\hfill
\begin{subfigure}[t]{\imgwidth}
    \includegraphics[width=\linewidth]{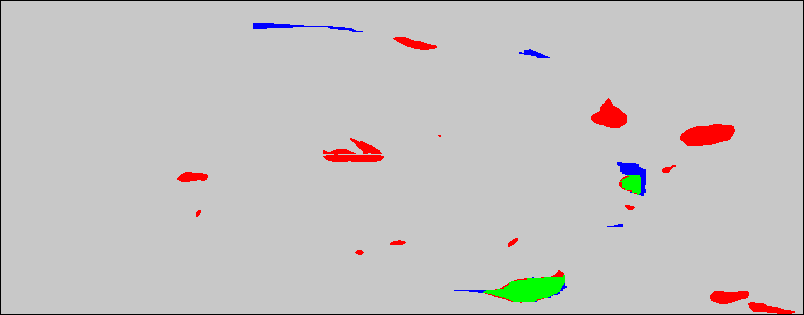}
\end{subfigure}\hfill
\begin{subfigure}[t]{\imgwidth}
    \includegraphics[width=\linewidth]{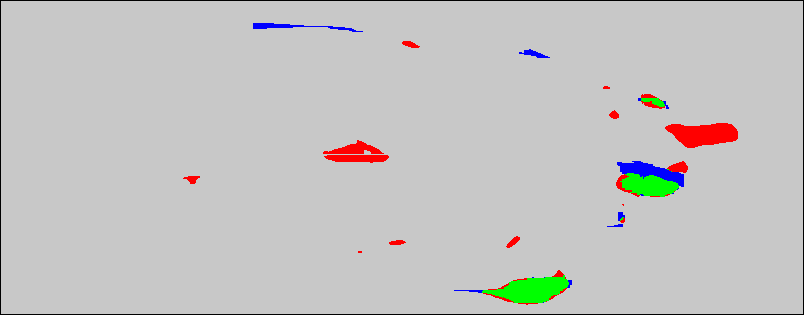}
\end{subfigure}\hfill
\begin{subfigure}[t]{\imgwidth}
    \includegraphics[width=\linewidth]{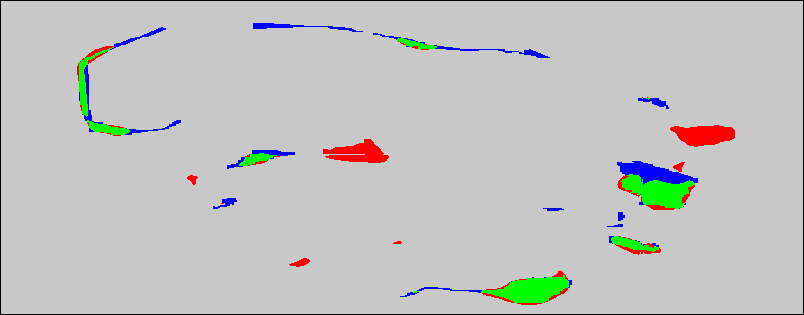}
\end{subfigure}\hfill
\begin{subfigure}[t]{\imgwidth}
    \includegraphics[width=\linewidth]{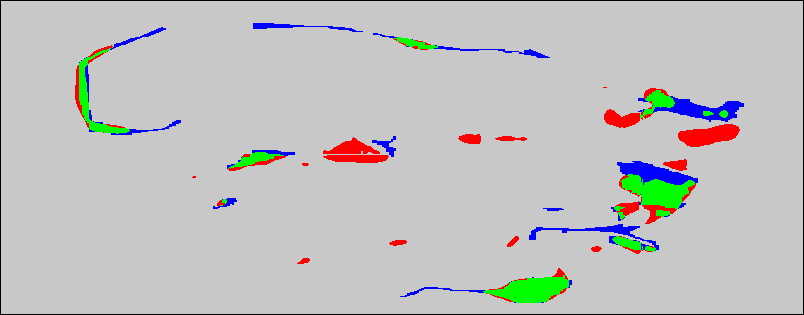}
\end{subfigure}\hfill
\begin{subfigure}[t]{\imgwidth}
    \includegraphics[width=\linewidth]{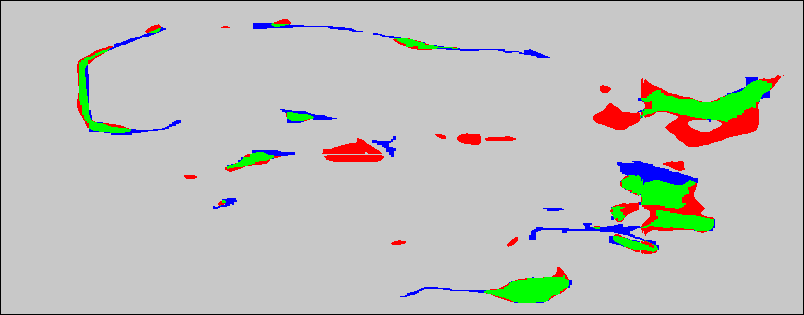}
\end{subfigure}\hfill
\begin{subfigure}[t]{\imgwidth}
    \includegraphics[width=\linewidth]{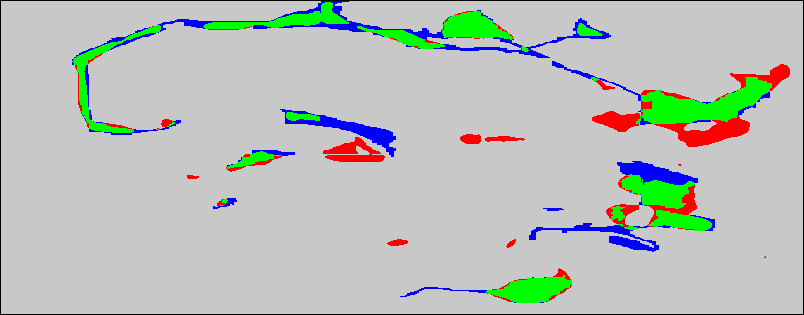}
\end{subfigure}\hfill
\begin{subfigure}[t]{\imgwidth}
    \includegraphics[width=\linewidth]{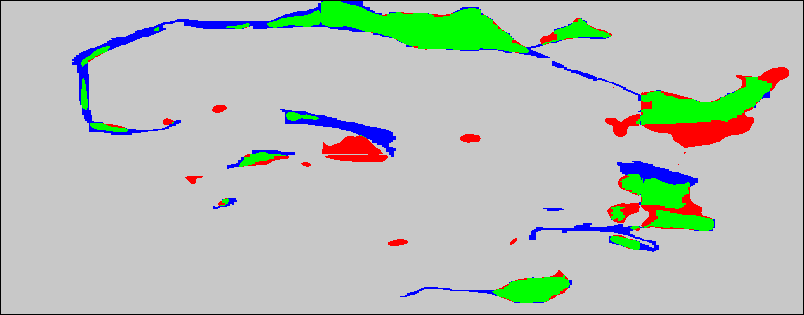}
\end{subfigure}\hfill
\begin{subfigure}[t]{\imgwidth}
    \includegraphics[width=\linewidth]{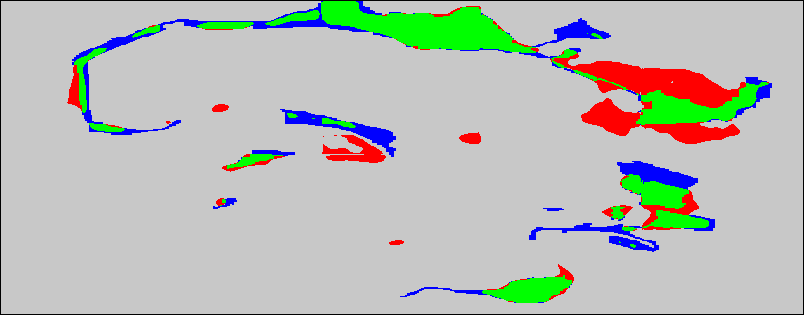}
\end{subfigure}\hfill
\\[2pt]

\raisebox{0.5\height}{\rowlabel{\scriptsize (d)}}\hspace{2pt}%
\begin{subfigure}[t]{\imgwidth}
    \includegraphics[width=\linewidth]{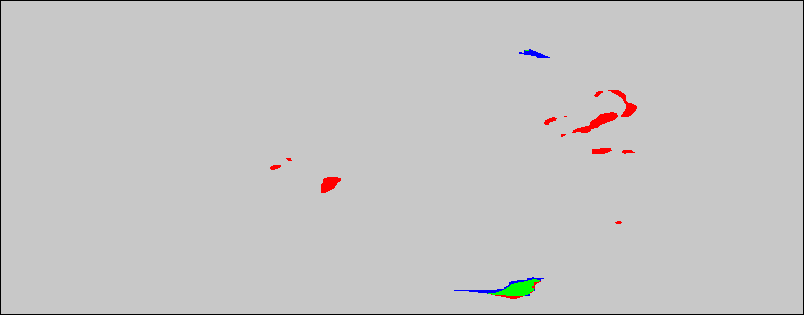}
\end{subfigure}\hfill
\begin{subfigure}[t]{\imgwidth}
    \includegraphics[width=\linewidth]{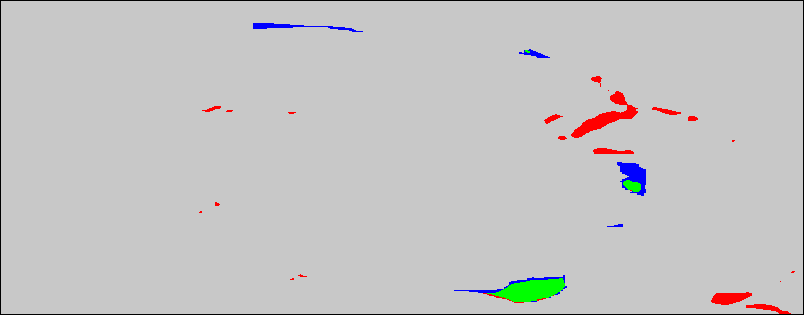}
\end{subfigure}\hfill
\begin{subfigure}[t]{\imgwidth}
    \includegraphics[width=\linewidth]{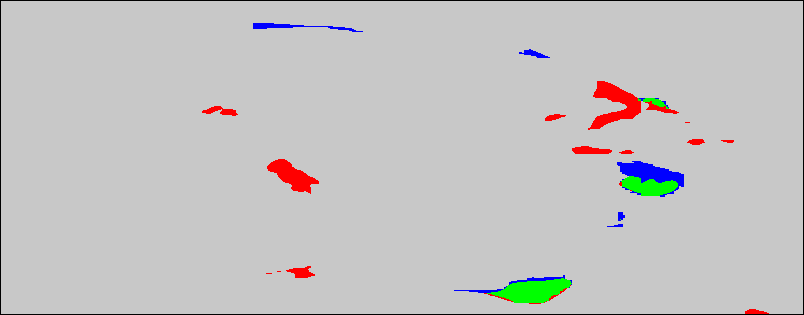}
\end{subfigure}\hfill
\begin{subfigure}[t]{\imgwidth}
    \includegraphics[width=\linewidth]{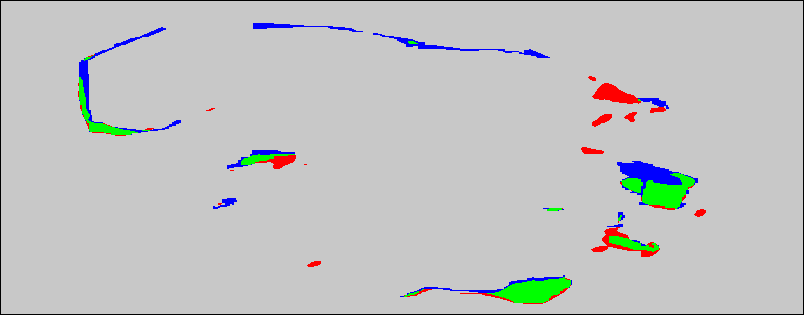}
\end{subfigure}\hfill
\begin{subfigure}[t]{\imgwidth}
    \includegraphics[width=\linewidth]{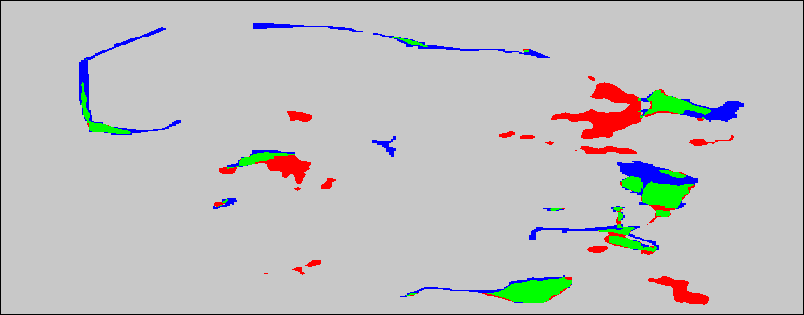}
\end{subfigure}\hfill
\begin{subfigure}[t]{\imgwidth}
    \includegraphics[width=\linewidth]{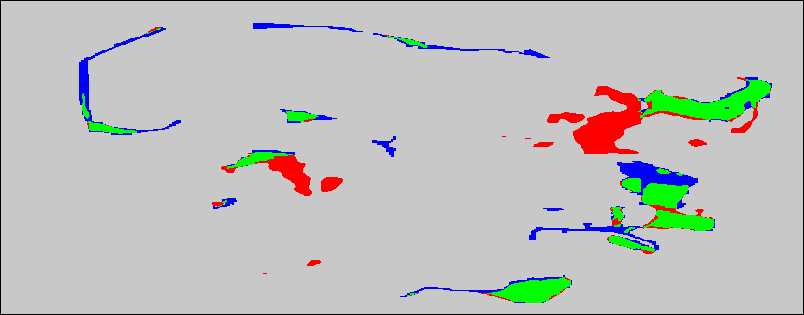}
\end{subfigure}\hfill
\begin{subfigure}[t]{\imgwidth}
    \includegraphics[width=\linewidth]{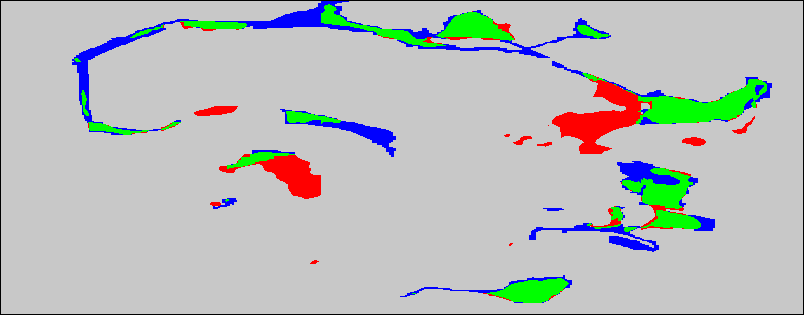}
\end{subfigure}\hfill
\begin{subfigure}[t]{\imgwidth}
    \includegraphics[width=\linewidth]{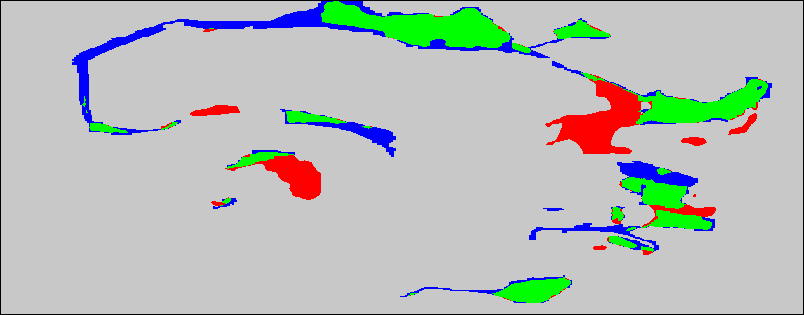}
\end{subfigure}\hfill
\begin{subfigure}[t]{\imgwidth}
    \includegraphics[width=\linewidth]{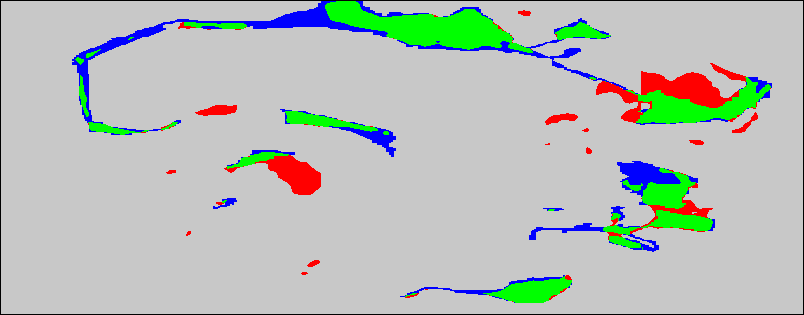}
\end{subfigure}\hfill
\\[2pt]

\raisebox{0.5\height}{\rowlabel{\scriptsize (e)}}\hspace{2pt}%
\begin{subfigure}[t]{\imgwidth}
    \includegraphics[width=\linewidth]{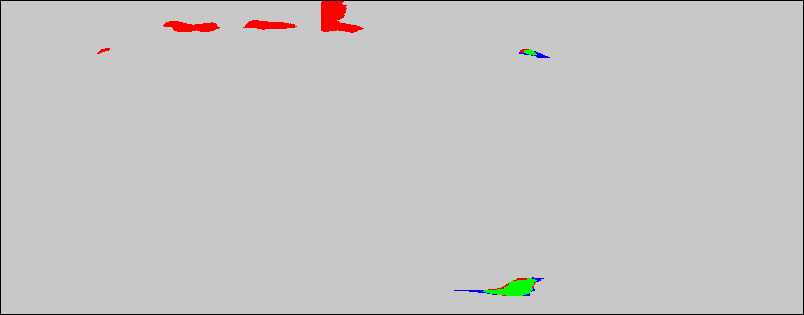}
\end{subfigure}\hfill
\begin{subfigure}[t]{\imgwidth}
    \includegraphics[width=\linewidth]{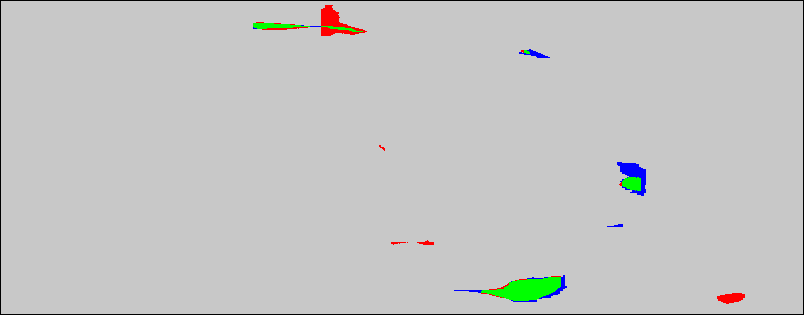}
\end{subfigure}\hfill
\begin{subfigure}[t]{\imgwidth}
    \includegraphics[width=\linewidth]{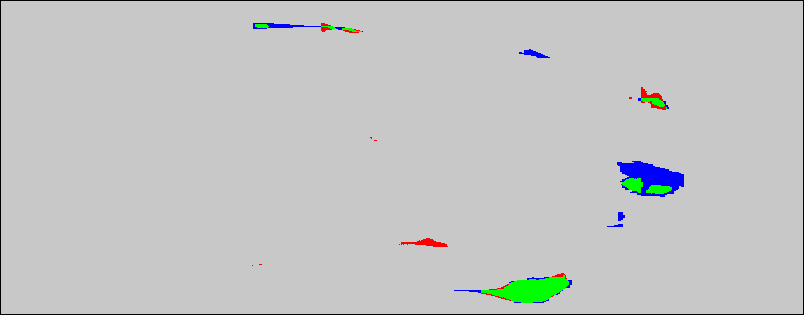}
\end{subfigure}\hfill
\begin{subfigure}[t]{\imgwidth}
    \includegraphics[width=\linewidth]{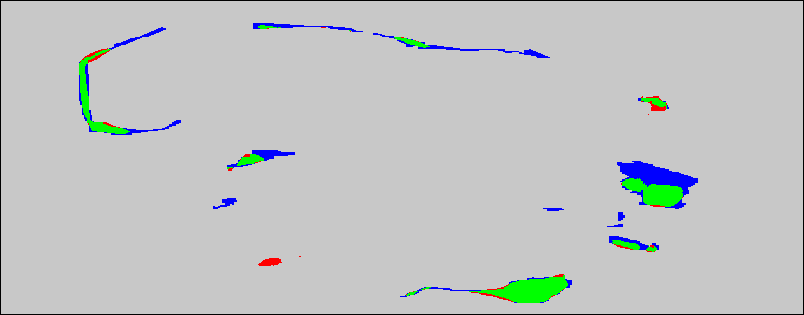}
\end{subfigure}\hfill
\begin{subfigure}[t]{\imgwidth}
    \includegraphics[width=\linewidth]{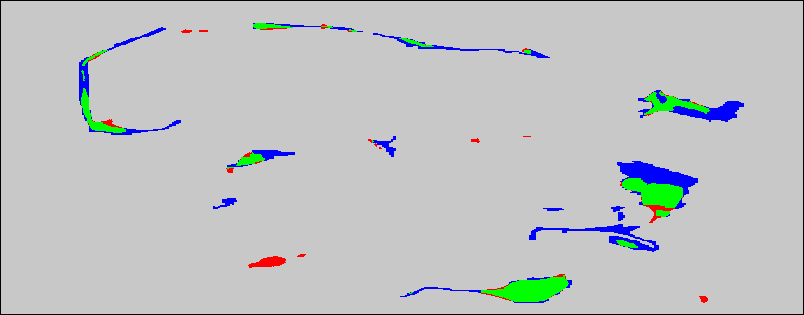}
\end{subfigure}\hfill
\begin{subfigure}[t]{\imgwidth}
    \includegraphics[width=\linewidth]{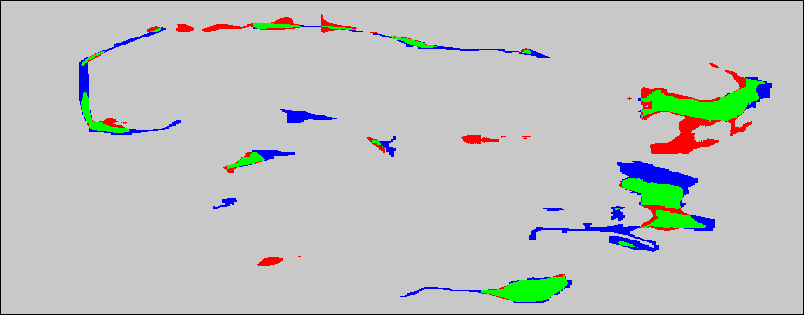}
\end{subfigure}\hfill
\begin{subfigure}[t]{\imgwidth}
    \includegraphics[width=\linewidth]{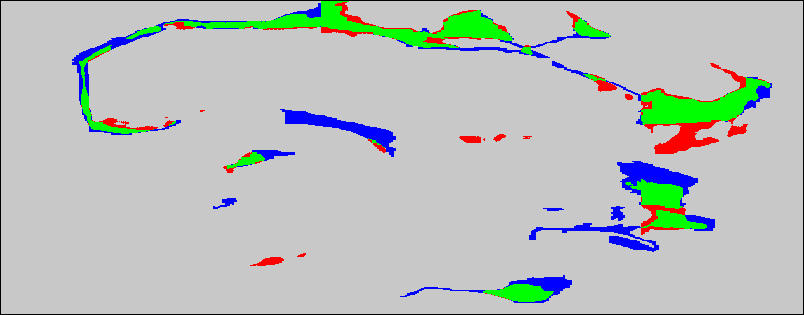}
\end{subfigure}\hfill
\begin{subfigure}[t]{\imgwidth}
    \includegraphics[width=\linewidth]{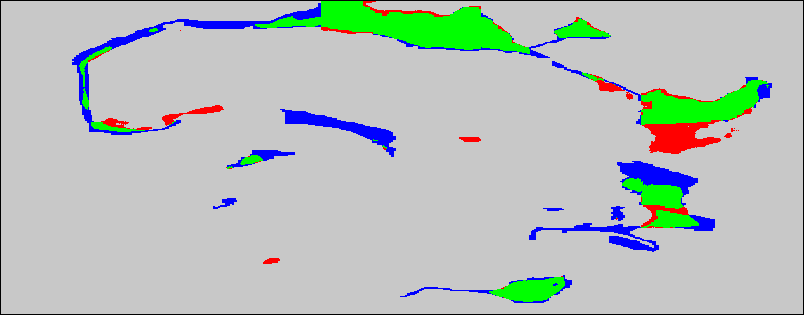}
\end{subfigure}\hfill
\begin{subfigure}[t]{\imgwidth}
    \includegraphics[width=\linewidth]{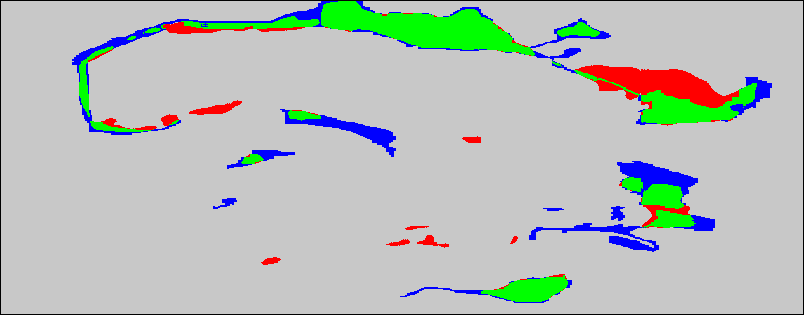}
\end{subfigure}\hfill
\\[2pt]

\raisebox{0.5\height}{\rowlabel{\scriptsize (f)}}\hspace{2pt}%
\begin{subfigure}[t]{\imgwidth}
    \includegraphics[width=\linewidth]{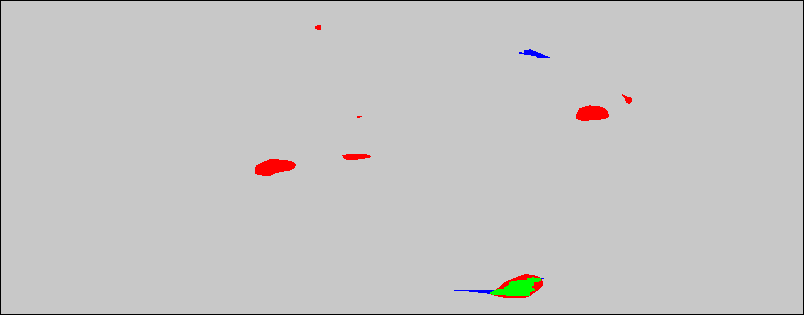}
\end{subfigure}\hfill
\begin{subfigure}[t]{\imgwidth}
    \includegraphics[width=\linewidth]{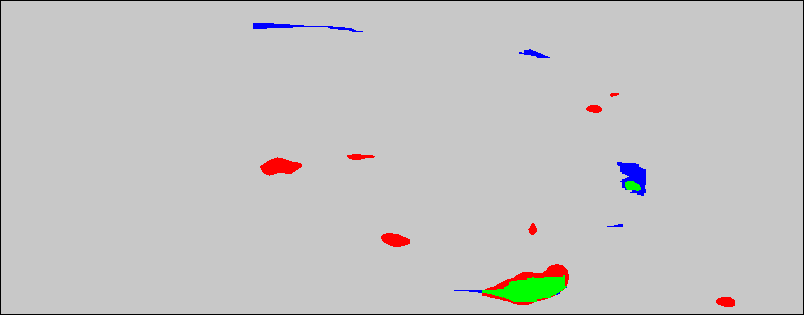}
\end{subfigure}\hfill
\begin{subfigure}[t]{\imgwidth}
    \includegraphics[width=\linewidth]{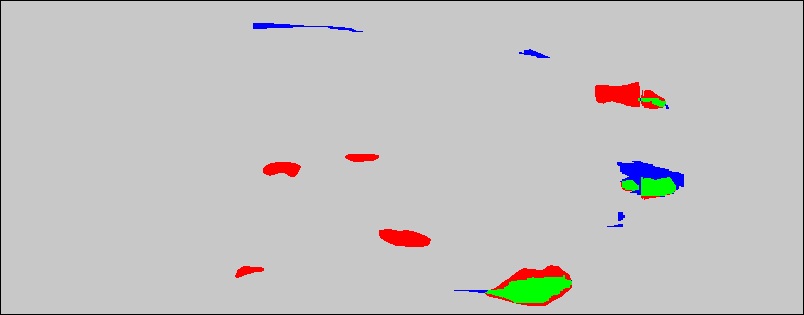}
\end{subfigure}\hfill
\begin{subfigure}[t]{\imgwidth}
    \includegraphics[width=\linewidth]{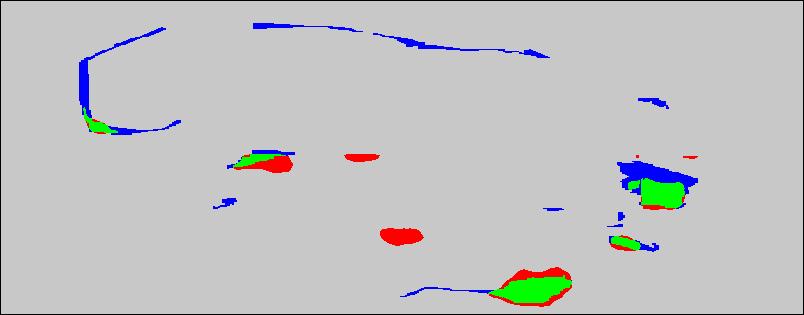}
\end{subfigure}\hfill
\begin{subfigure}[t]{\imgwidth}
    \includegraphics[width=\linewidth]{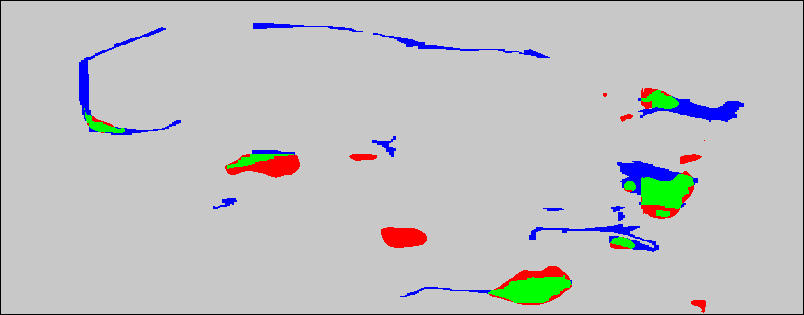}
\end{subfigure}\hfill
\begin{subfigure}[t]{\imgwidth}
    \includegraphics[width=\linewidth]{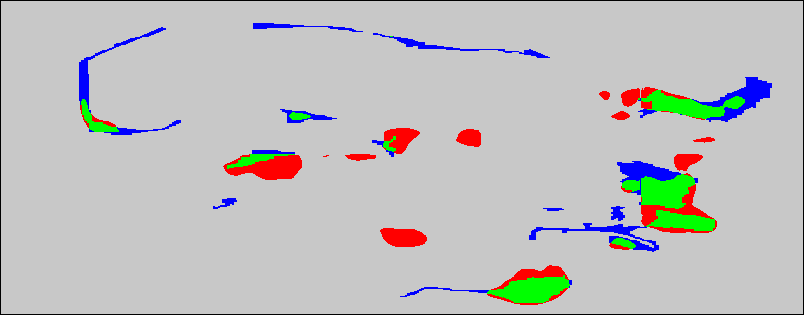}
\end{subfigure}\hfill
\begin{subfigure}[t]{\imgwidth}
    \includegraphics[width=\linewidth]{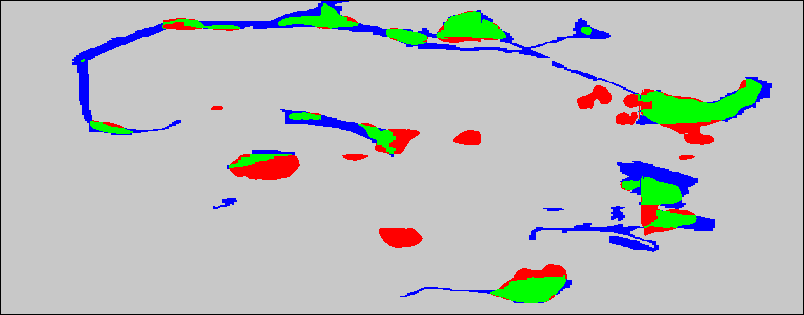}
\end{subfigure}\hfill
\begin{subfigure}[t]{\imgwidth}
    \includegraphics[width=\linewidth]{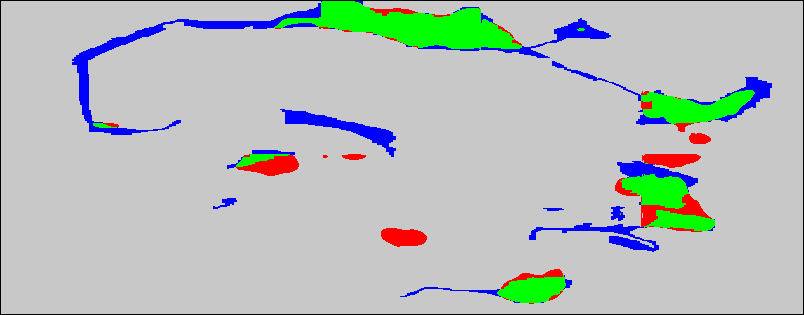}
\end{subfigure}\hfill
\begin{subfigure}[t]{\imgwidth}
    \includegraphics[width=\linewidth]{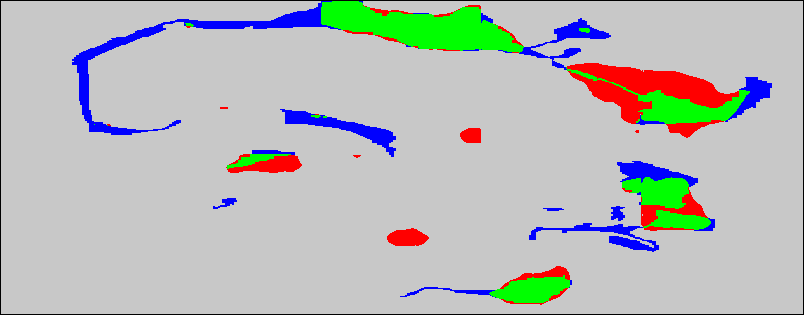}
\end{subfigure}\hfill
\\[2pt]

\raisebox{0.5\height}{\rowlabel{\scriptsize (g)}}\hspace{2pt}%
\begin{subfigure}[t]{\imgwidth}
    \includegraphics[width=\linewidth]{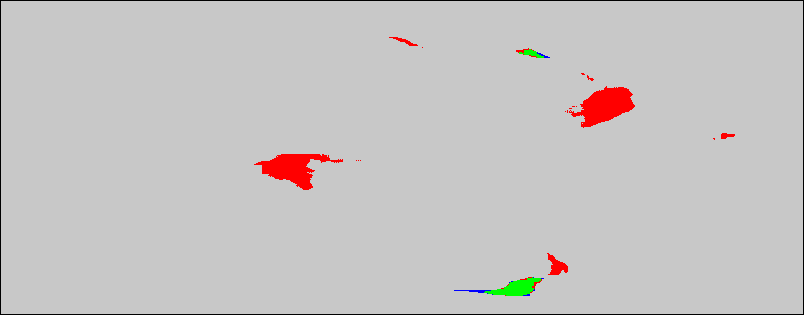}
\end{subfigure}\hfill
\begin{subfigure}[t]{\imgwidth}
    \includegraphics[width=\linewidth]{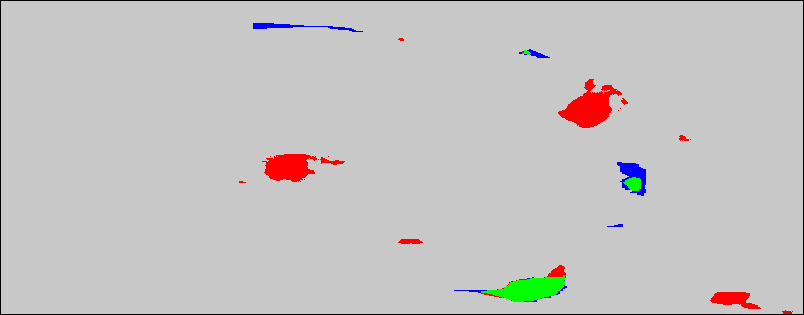}
\end{subfigure}\hfill
\begin{subfigure}[t]{\imgwidth}
    \includegraphics[width=\linewidth]{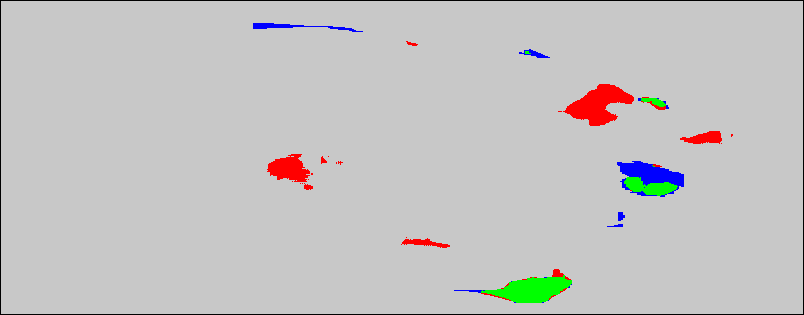}
\end{subfigure}\hfill
\begin{subfigure}[t]{\imgwidth}
    \includegraphics[width=\linewidth]{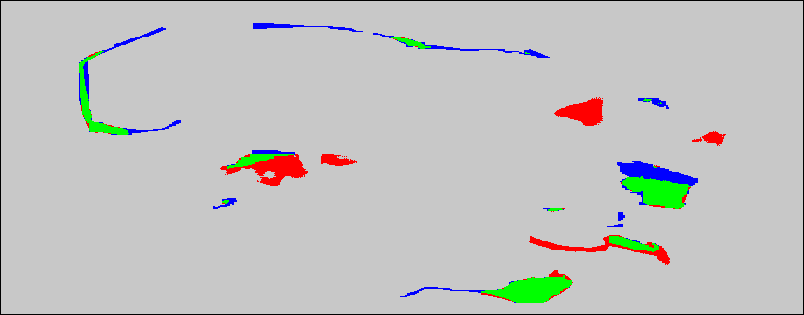}
\end{subfigure}\hfill
\begin{subfigure}[t]{\imgwidth}
    \includegraphics[width=\linewidth]{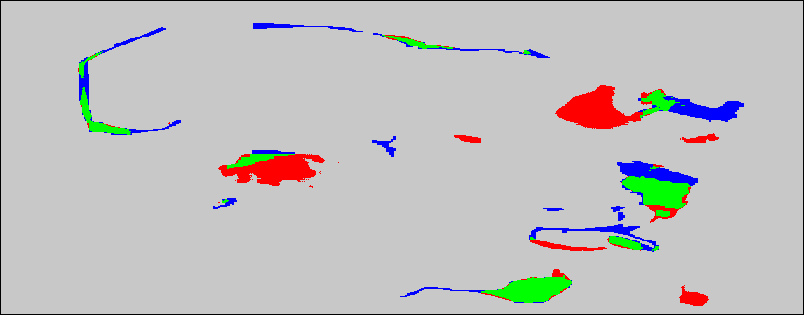}
\end{subfigure}\hfill
\begin{subfigure}[t]{\imgwidth}
    \includegraphics[width=\linewidth]{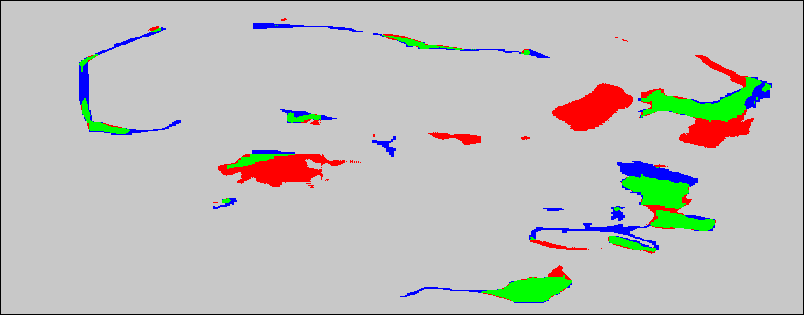}
\end{subfigure}\hfill
\begin{subfigure}[t]{\imgwidth}
    \includegraphics[width=\linewidth]{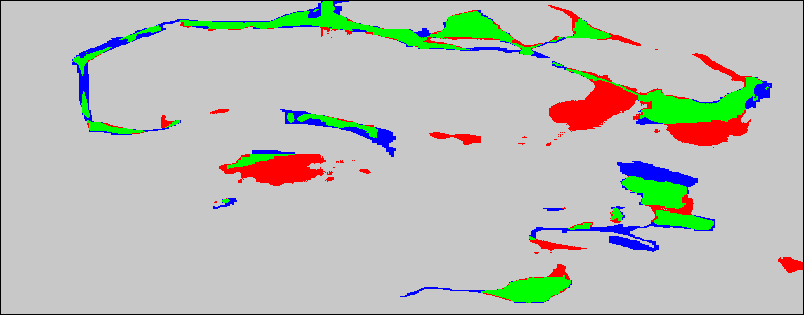}
\end{subfigure}\hfill
\begin{subfigure}[t]{\imgwidth}
    \includegraphics[width=\linewidth]{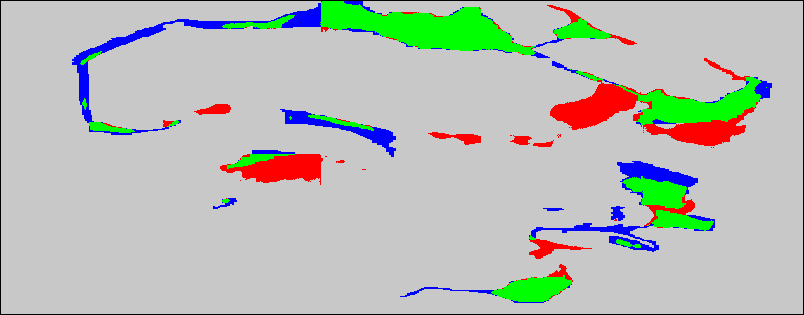}
\end{subfigure}\hfill
\begin{subfigure}[t]{\imgwidth}
    \includegraphics[width=\linewidth]{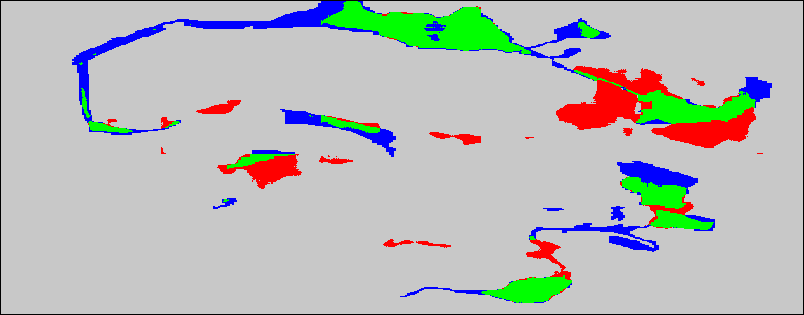}
\end{subfigure}\hfill
\\[2pt]

\raisebox{0.5\height}{\rowlabel{\scriptsize (h)}}\hspace{2pt}%
\begin{subfigure}[t]{\imgwidth}
    \includegraphics[width=\linewidth]{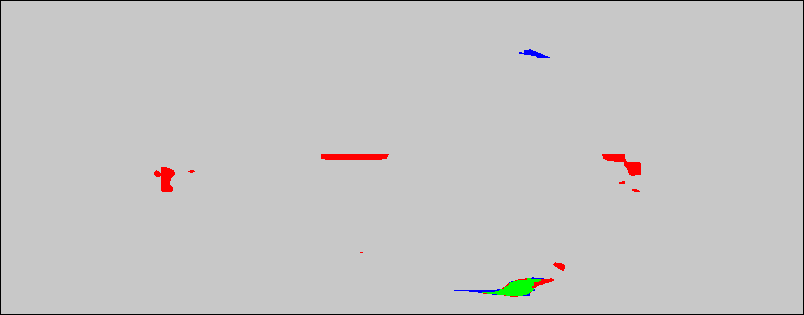}
\end{subfigure}\hfill
\begin{subfigure}[t]{\imgwidth}
    \includegraphics[width=\linewidth]{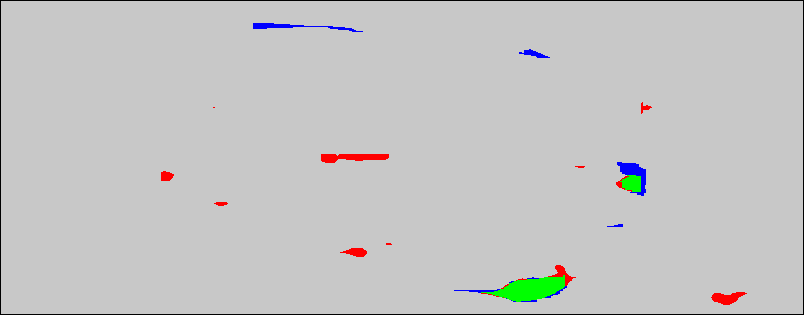}
\end{subfigure}\hfill
\begin{subfigure}[t]{\imgwidth}
    \includegraphics[width=\linewidth]{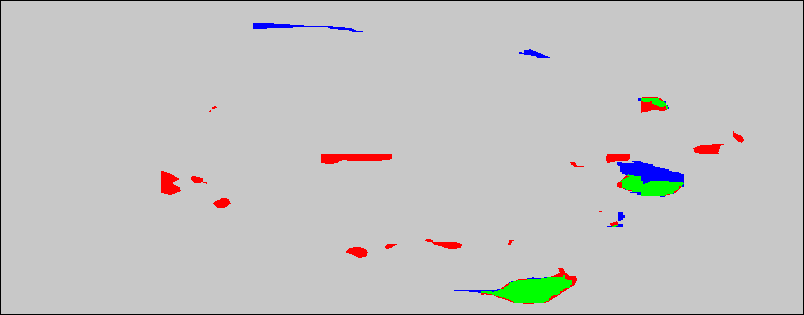}
\end{subfigure}\hfill
\begin{subfigure}[t]{\imgwidth}
    \includegraphics[width=\linewidth]{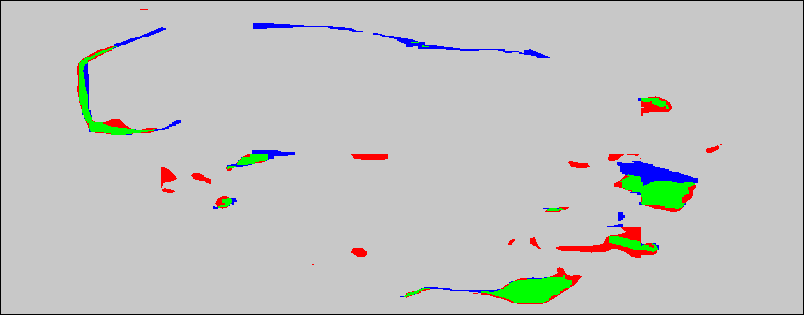}
\end{subfigure}\hfill
\begin{subfigure}[t]{\imgwidth}
    \includegraphics[width=\linewidth]{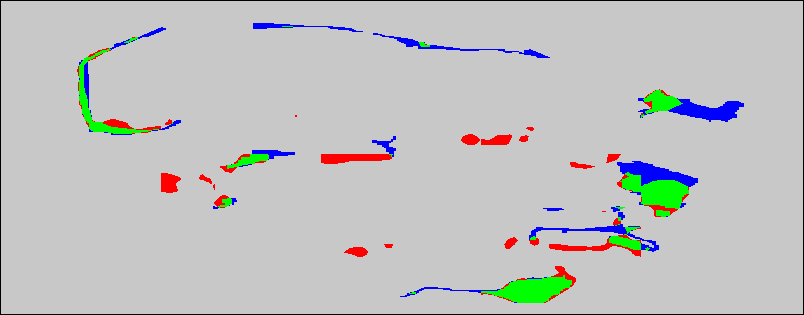}
\end{subfigure}\hfill
\begin{subfigure}[t]{\imgwidth}
    \includegraphics[width=\linewidth]{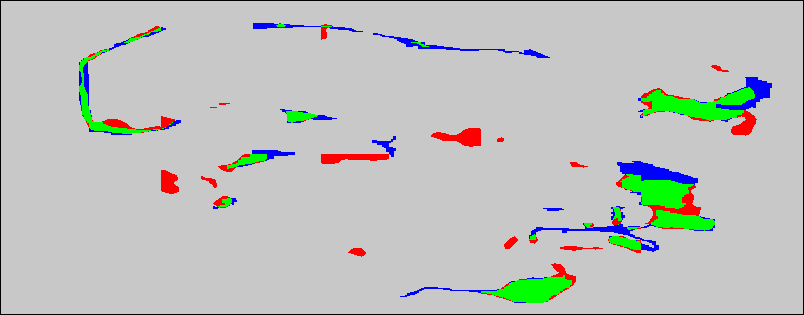}
\end{subfigure}\hfill
\begin{subfigure}[t]{\imgwidth}
    \includegraphics[width=\linewidth]{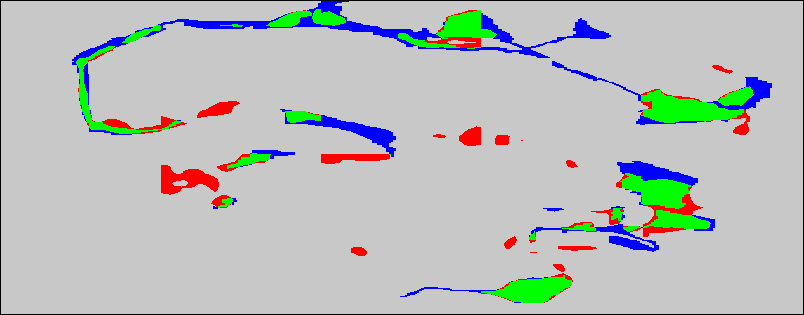}
\end{subfigure}\hfill
\begin{subfigure}[t]{\imgwidth}
    \includegraphics[width=\linewidth]{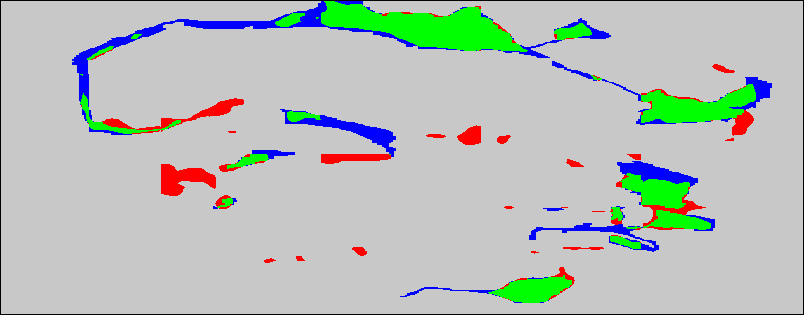}
\end{subfigure}\hfill
\begin{subfigure}[t]{\imgwidth}
    \includegraphics[width=\linewidth]{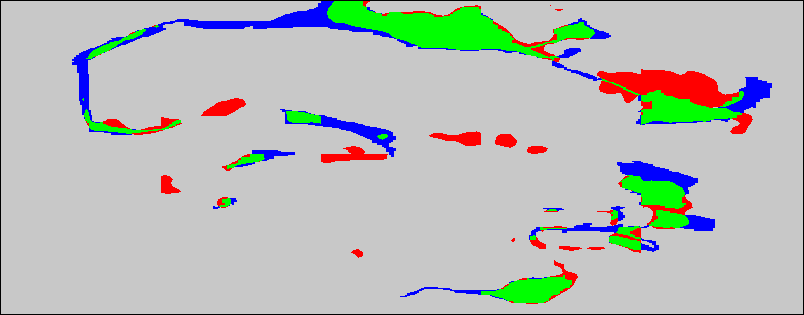}
\end{subfigure}\hfill
\\[2pt]

\raisebox{0.5\height}{\rowlabel{\scriptsize (i)}}\hspace{2pt}%
\begin{subfigure}[t]{\imgwidth}
    \includegraphics[width=\linewidth]{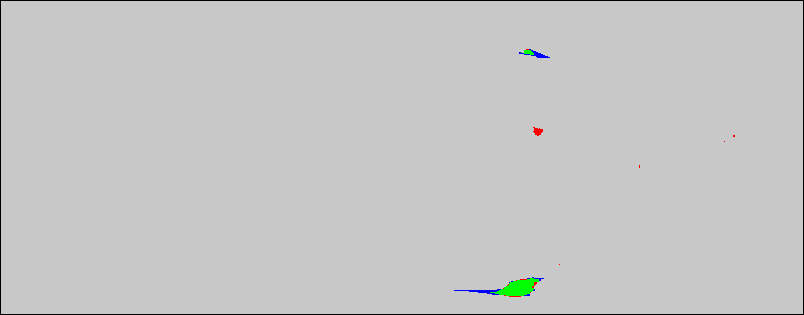}
\end{subfigure}\hfill
\begin{subfigure}[t]{\imgwidth}
    \includegraphics[width=\linewidth]{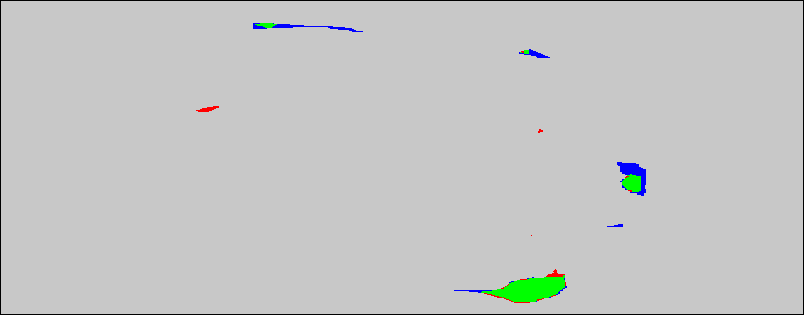}
\end{subfigure}\hfill
\begin{subfigure}[t]{\imgwidth}
    \includegraphics[width=\linewidth]{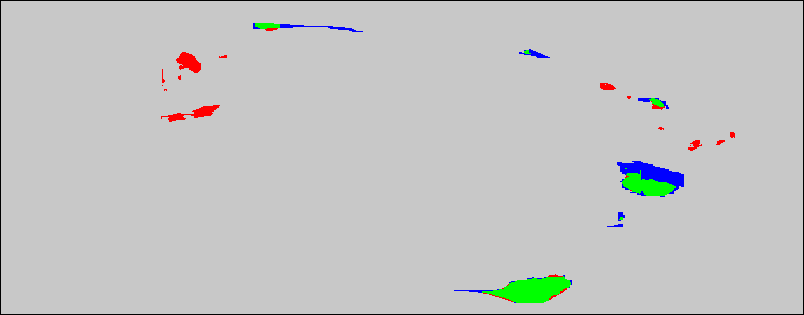}
\end{subfigure}\hfill
\begin{subfigure}[t]{\imgwidth}
    \includegraphics[width=\linewidth]{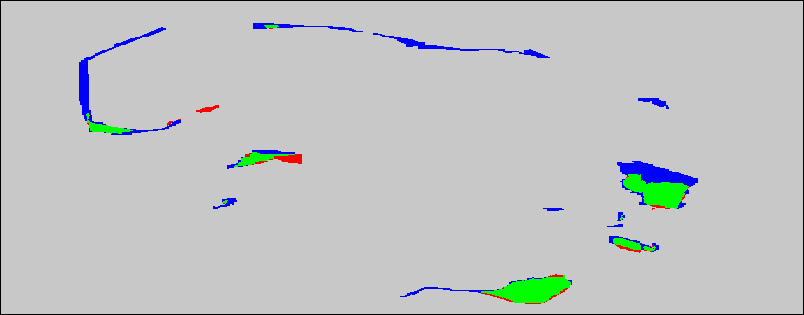}
\end{subfigure}\hfill
\begin{subfigure}[t]{\imgwidth}
    \includegraphics[width=\linewidth]{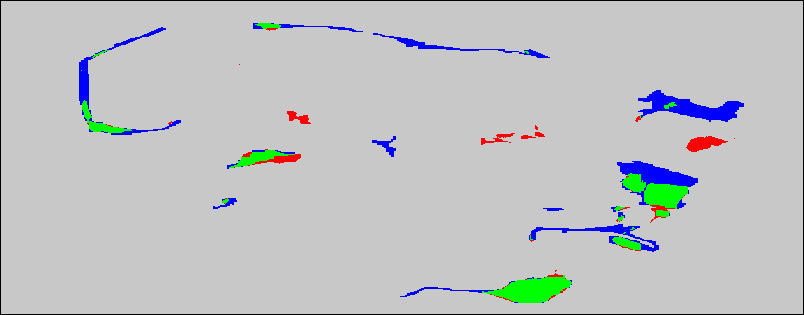}
\end{subfigure}\hfill
\begin{subfigure}[t]{\imgwidth}
    \includegraphics[width=\linewidth]{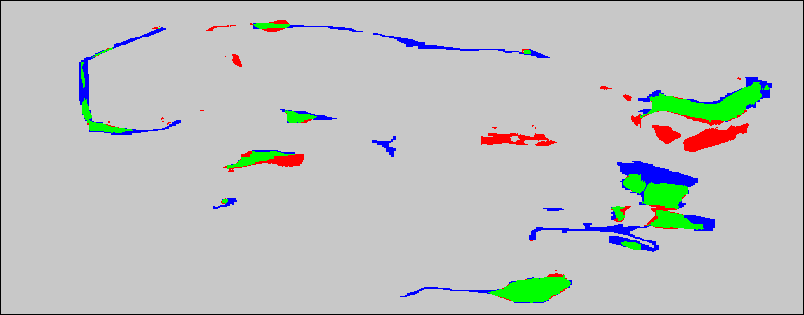}
\end{subfigure}\hfill
\begin{subfigure}[t]{\imgwidth}
    \includegraphics[width=\linewidth]{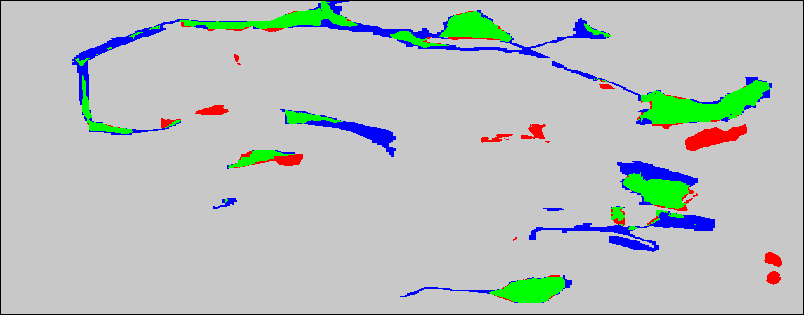}
\end{subfigure}\hfill
\begin{subfigure}[t]{\imgwidth}
    \includegraphics[width=\linewidth]{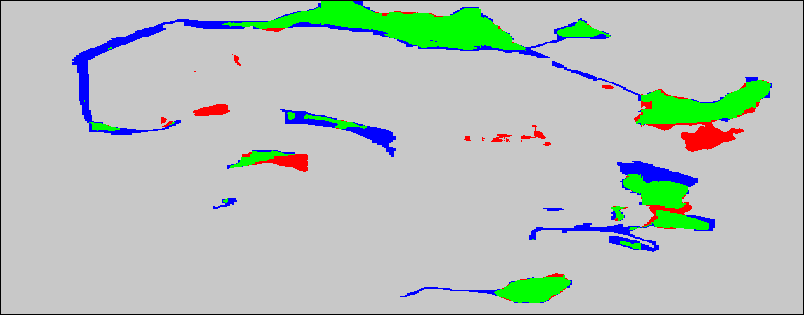}
\end{subfigure}\hfill
\begin{subfigure}[t]{\imgwidth}
    \includegraphics[width=\linewidth]{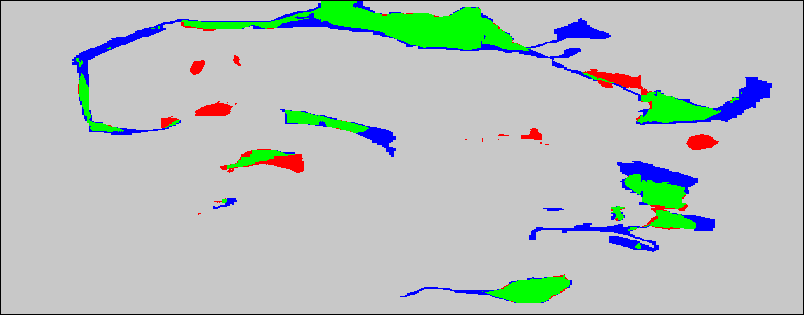}
\end{subfigure}\hfill
\\[2pt]

\raisebox{0.5\height}{\rowlabel{\scriptsize (j)}}\hspace{2pt}%
\begin{subfigure}[t]{\imgwidth}
    \includegraphics[width=\linewidth]{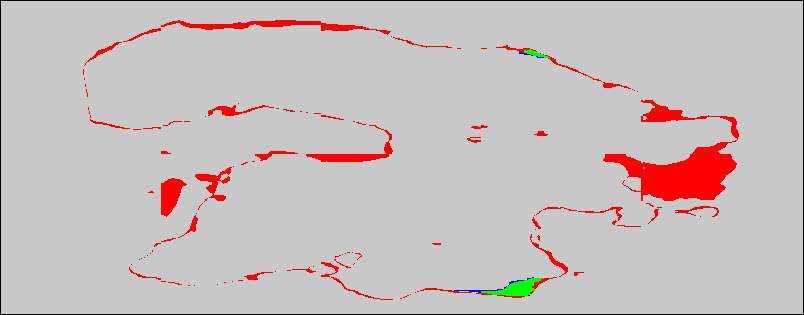}
\end{subfigure}\hfill
\begin{subfigure}[t]{\imgwidth}
    \includegraphics[width=\linewidth]{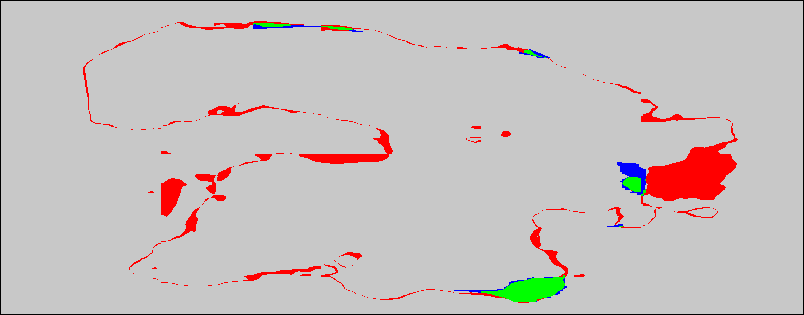}
\end{subfigure}\hfill
\begin{subfigure}[t]{\imgwidth}
    \includegraphics[width=\linewidth]{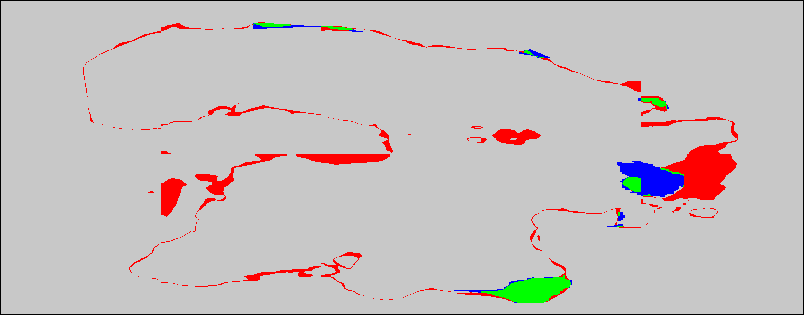}
\end{subfigure}\hfill
\begin{subfigure}[t]{\imgwidth}
    \includegraphics[width=\linewidth]{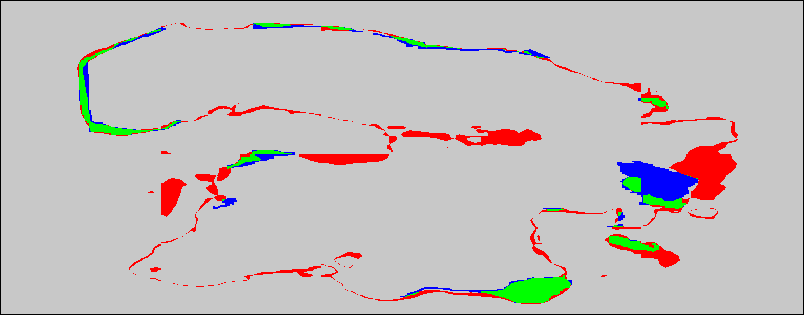}
\end{subfigure}\hfill
\begin{subfigure}[t]{\imgwidth}
    \includegraphics[width=\linewidth]{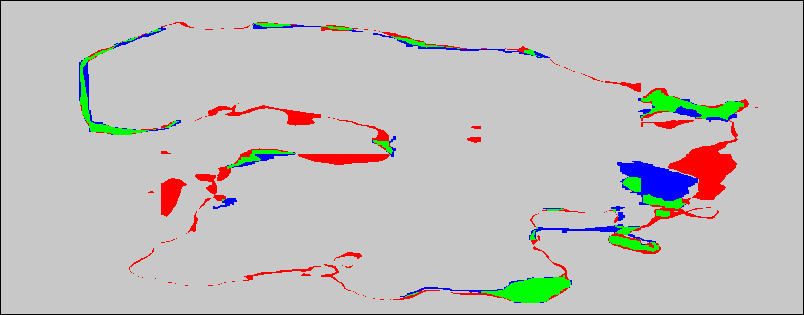}
\end{subfigure}\hfill
\begin{subfigure}[t]{\imgwidth}
    \includegraphics[width=\linewidth]{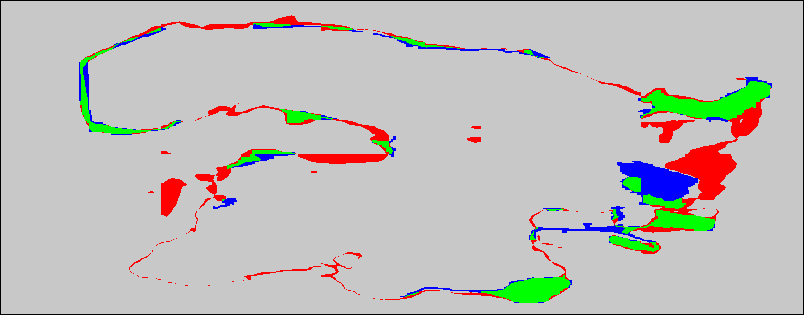}
\end{subfigure}\hfill
\begin{subfigure}[t]{\imgwidth}
    \includegraphics[width=\linewidth]{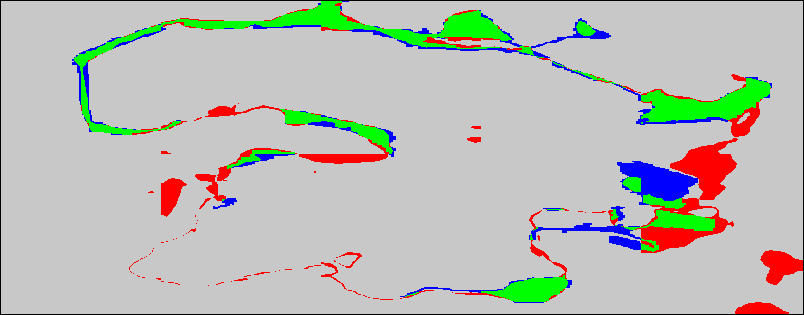}
\end{subfigure}\hfill
\begin{subfigure}[t]{\imgwidth}
    \includegraphics[width=\linewidth]{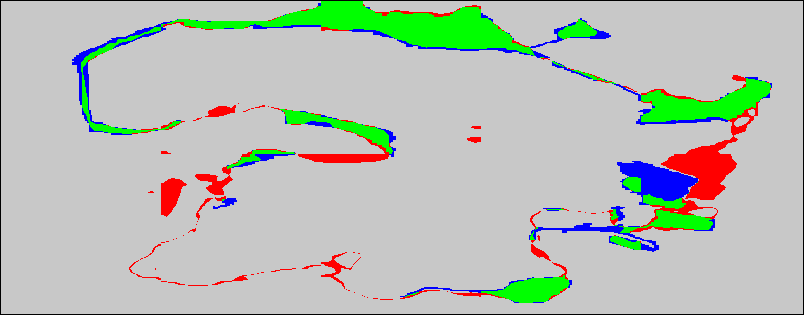}
\end{subfigure}\hfill
\begin{subfigure}[t]{\imgwidth}
    \includegraphics[width=\linewidth]{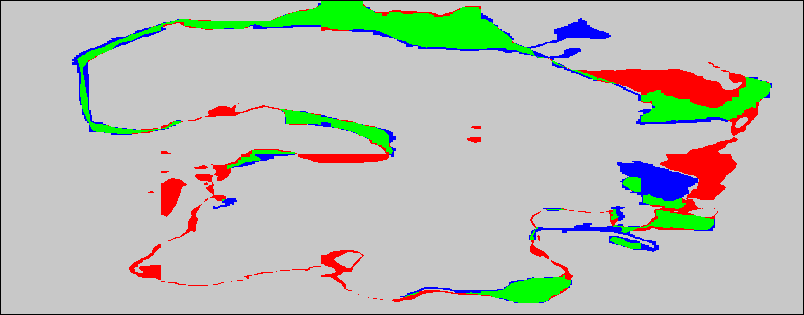}
\end{subfigure}\hfill
\\[2pt]

\raisebox{0.5\height}{\rowlabel{\scriptsize (k)}}\hspace{2pt}%
\begin{subfigure}[t]{\imgwidth}
    \includegraphics[width=\linewidth]{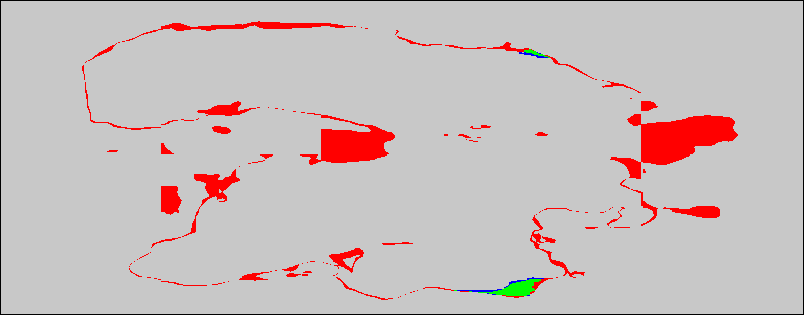}
    \caption{$2015\to 2016$}
\end{subfigure}\hfill
\begin{subfigure}[t]{\imgwidth}
    \includegraphics[width=\linewidth]{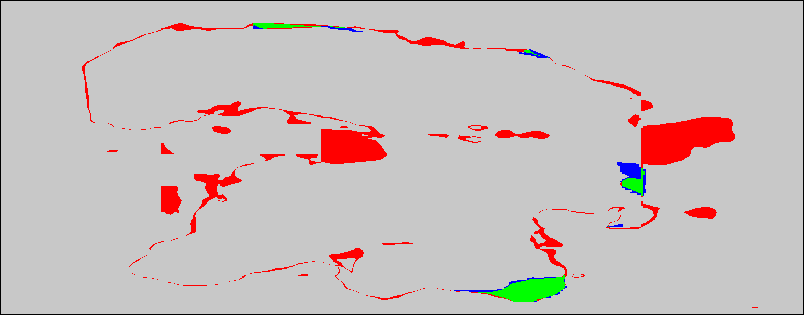}
    \caption{$2015\to 2017$}
\end{subfigure}\hfill
\begin{subfigure}[t]{\imgwidth}
    \includegraphics[width=\linewidth]{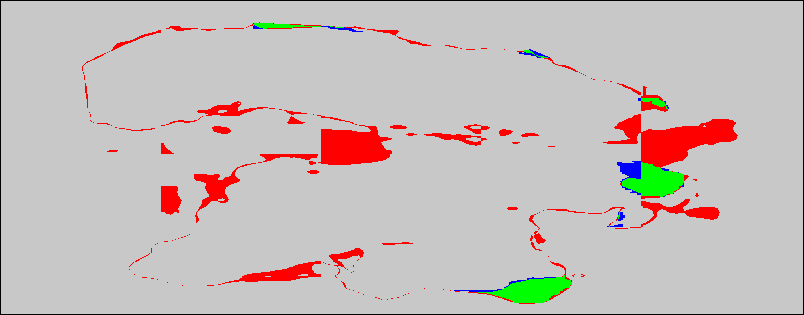}
    \caption{$2015\to 2018$}
\end{subfigure}\hfill
\begin{subfigure}[t]{\imgwidth}
    \includegraphics[width=\linewidth]{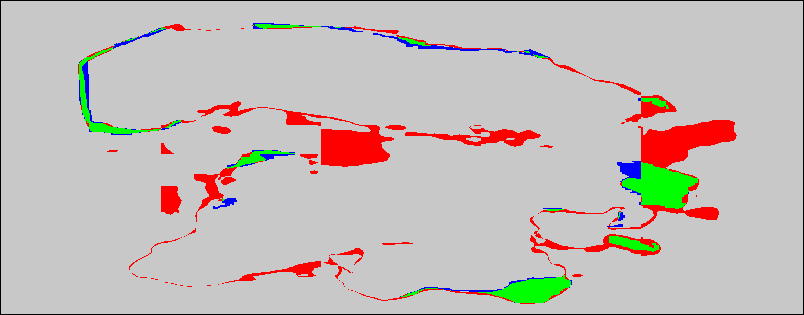}
    \caption{$2015\to 2019$}
\end{subfigure}\hfill
\begin{subfigure}[t]{\imgwidth}
    \includegraphics[width=\linewidth]{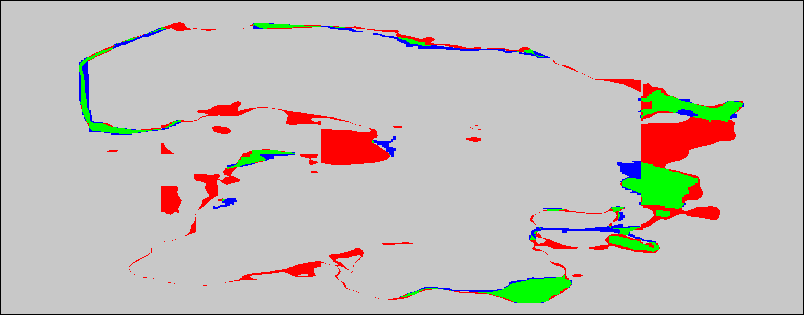}
    \caption{$2015\to 2020$}
\end{subfigure}\hfill
\begin{subfigure}[t]{\imgwidth}
    \includegraphics[width=\linewidth]{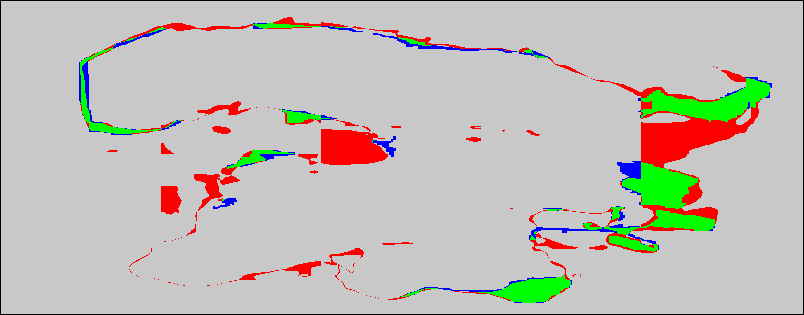}
    \caption{$2015\to 2021$}
\end{subfigure}\hfill
\begin{subfigure}[t]{\imgwidth}
    \includegraphics[width=\linewidth]{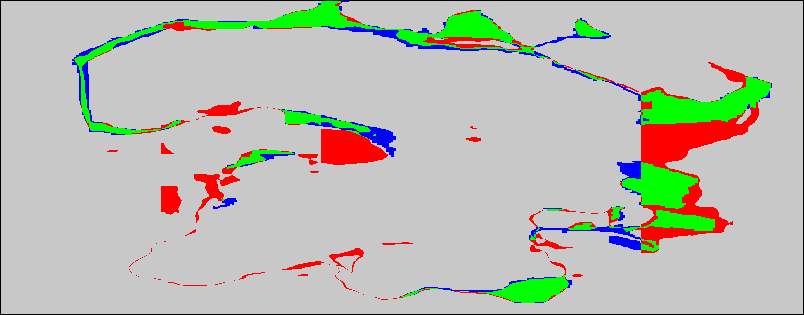}
    \caption{$2015\to 2022$}
\end{subfigure}\hfill
\begin{subfigure}[t]{\imgwidth}
    \includegraphics[width=\linewidth]{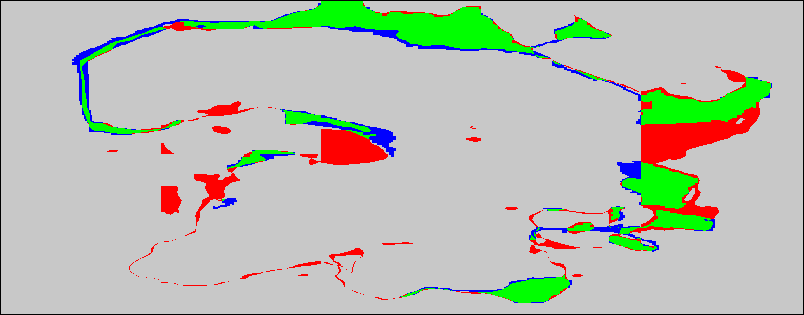}
    \caption{$2015\to 2023$}
\end{subfigure}\hfill
\begin{subfigure}[t]{\imgwidth}
    \includegraphics[width=\linewidth]{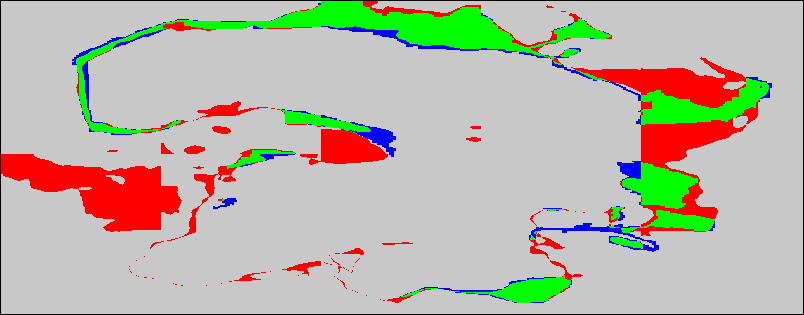}
    \caption{$2015\to 2024$}
\end{subfigure}\hfill
\\

\caption{Qualitative comparison for cross-temporal mining footprint monitoring, demonstrated by the site Björkdal Gold Mine, Sweden. The pixels of TP, TN, FP, and FN are indicated in green, grey, red, and blue, respectively. The first row (a) indicates the multitemporal images (2015$\to$ 2024 from left to right, only RGB bands are demonstrated for visualization), which are paired as inputs for change detection models. The second to the ninth rows demonstrate results obtained by change detection models (b) A2Net (c) BIT (d) CGNet (e) ChangeFormer (f) DMINet (g) ResUNet (h) TFI-GR (i) TinyCD, and the tenth and eleventh rows demonstrate the results captured by post-classification-based mining footprint mapping methods (j) Segformer (k) UperNet-SwinT. We only display mining footprint monitoring results with the starting year of 2015, due to the limited space in the paper. These models are selected as they have outperformed other methods.}
\label{fig:cross_temp_cd}
\end{figure*}

\begin{table*}
\caption{Quantitative comparison for multitemporal mining footprint mapping results, reported by different interval years. The overall F1 (OF1) is demonstrated in the last column by calculating the average F1 score from all samples. All the metrics are the higher the better. The best and second-best values are highlighted in bold and underlined format, respectively.}
\label{tab:multi_map_result}
\resizebox{\linewidth}{!}{%
\begin{tabular}{l|ccccccccccccc}
\toprule
Models & 2015 & 2016 & 2017 & 2018 & 2019 & 2020 & 2021 & 2022 & 2023 & 2024 & OF1 &GCA-TIoU &LCA-TIoU\\
\midrule
DeeplabV3 (MobileViTV2) & 0.8441 & 0.8326 & 0.8279 & 0.8399 & 0.8419 & 0.8541 & 0.8543 & 0.8410 & 0.8570 & 0.8439 & 0.8437 &0.7753 &	0.8048 \\
DeeplabV3P (ResNet-101) & 0.8510 & 0.8316 & 0.8205 & 0.8441 & 0.8401 & 0.8455 & 0.8475 & 0.8360 & 0.8397 & 0.8369 & 0.8394 &0.7659 	&0.7949 \\
DeeplabV3P (MobileNetV2) & 0.8342 & 0.8225 & 0.8120 & 0.8282 & 0.8265 & 0.8466 & 0.8328 & 0.8413 & 0.8326 & 0.8210 & 0.8299 &0.7732 	&0.8060 \\
LinkNet & 0.8346 & 0.8317 & 0.8170 & 0.8469 & 0.8436 & 0.8363 & 0.8511 & 0.8242 & 0.8318 & 0.8447 & 0.8362 &0.7693 	&0.8049 \\
Mask2Former (SwinT-B) & 0.8521 & 0.8444 & \underline{0.8435} & 0.8497 & 0.8416 & 0.8517 & 0.8498 & 0.8422 & \underline{0.8605} & 0.8440 & 0.8480 &0.7921 &0.8281 \\
UNet & 0.8514 & 0.8358 & 0.8411 & 0.8504 & 0.8462 & \underline{0.8593} & 0.8549 & \underline{0.8568} & 0.8552 & \underline{0.8535} & 0.8505 &0.7754  &0.8093  \\
UperNet (ConvNext-B5) & 0.8485 & 0.8368 & 0.8338 & \underline{0.8552} & 0.8433 & 0.8502 & \underline{0.8565} & 0.8426 & 0.8590 & 0.8481 & 0.8475 &0.7862 &	0.8177 \\ 
UperNet (SwinT-B) & \underline{0.8559} & \underline{0.8453} & \textbf{0.8448} & 0.8491 & \textbf{0.8525} & \textbf{0.8604} & \textbf{0.8641} & \textbf{0.8615} & \textbf{0.8637} & \textbf{0.8597} & \textbf{0.8558} & \textbf{0.8033} 	&\textbf{0.8400} \\
PSPNet & 0.8278 & 0.8313 & 0.8209 & 0.8402 & 0.8320 & 0.8447 & 0.8446 & 0.8341 & 0.8150 & 0.8372 & 0.8327 &0.7759 	& 0.8070  \\
SQNet & 0.8421 & 0.8179 & 0.8050 & 0.8402 & 0.8246 & 0.8403 & 0.8438 & 0.8188 & 0.8341 & 0.8305 & 0.8298 &0.7414 	&0.7691 \\
Segformer & \textbf{0.8593} & \textbf{0.8470} & 0.8393 & \textbf{0.8562} & \underline{0.8492} & 0.8555 & 0.8550 & 0.8476 & 0.8485 & 0.8529 & \underline{0.8511} &\underline{0.7947} &\underline{0.8286} \\
\bottomrule
\end{tabular}}
\end{table*}
\subsection{Results for Cross-temporal Change Detection}
\subsubsection{Change Detection Methods}
The performance patterns in Table \ref{tab:cross_temporal_results_cd_methods} and Fig. \ref{fig:cross_temp_cd} reveal distinct operational challenges in cross-temporal change detection, closely tied to the temporal gap between observations. For short intervals (1–3 years), the detection task is inherently difficult. Changes within this range are often subtle, reflecting early stages of anthropogenic activity, seasonal vegetation fluctuations, or gradual environmental processes that only partially alter land surface appearance. Such variations are easily masked by sensor noise, atmospheric differences, or illumination changes, making it challenging for models to consistently separate true structural change from background temporal variability. This explains why absolute performance is lower in the short term and why even top-performing models show a significant gap compared to their long-term scores.

In contrast, long intervals (4–9 years) typically encompass more pronounced and spatially extensive transformations—such as large-scale industrial expansion, infrastructure projects, or major land cover conversions—that are more robust to seasonal and sensor-induced noise. The greater magnitude of change over these periods narrows the performance gap between different methods, as the change signal becomes dominant and easier to detect regardless of architectural sophistication. However, this apparent improvement comes with an operational drawback: detecting changes only after they have accumulated over many years offers limited value for proactive monitoring or early intervention.


From a task-design perspective, this contrast highlights two fundamental challenges for the dataset and the problem setting. On the one hand, short-term change detection demands high temporal sensitivity that models must capture weak, localized variations while suppressing transient noise. This requires temporal consistency mechanisms and fine-grained spatial–temporal feature extraction. On the other hand, long-term change detection, while easier, offers less actionable insight — by the time changes become obvious, opportunities for prevention or mitigation may have passed.

For real-world applications such as monitoring mining expansion, tracking urban growth, or detecting environmental degradation, the ability to identify early-stage changes is critical. Missing these early signals could mean forfeiting the narrow window in which intervention is possible. The results here make clear that short-term monitoring is the more technically demanding and societally valuable aspect of the task, positioning it as a key frontier for methodological innovation.

\begin{table*}[htbp]
\caption{Quantitative comparison for cross-temporal change detection results generated by change detection models, reported by different interval years. The overall F1 (OF1) is demonstrated in the last column by calculating the average F1 score from all samples. The best and second-best values are highlighted in bold and underlined format, respectively.}
\label{tab:cross_temporal_results_cd_methods}
\resizebox{\linewidth}{!}{%
\begin{tabular}{l|cccccccccc}
\toprule
\multirow{2}{*}{Models} & \multicolumn{10}{c}{Interval Year(s)} \\
 & 1 & 2 & 3 & 4 & 5 & 6 & 7 & 8 & 9 & OF1 \\ \midrule
A2Net & 0.3092 & 0.3886 & 0.4285 & 0.4704 & 0.5105 & 0.5311 & 0.5626 & \textbf{0.5946} & \textbf{0.6109} & 0.4923 \\
AFCF3D & 0.2432 & 0.3500 & 0.3867 & 0.4197 & 0.4451 & 0.4559 & 0.4666 & 0.4788 & 0.4950 & 0.4218 \\
BIT & 0.3022 & 0.3999 & 0.4373 & 0.4669 & 0.4944 & 0.5205 & 0.5481 & 0.5562 & 0.5797 & 0.4809 \\
CGNet & 0.3198 & 0.4021 & 0.4119 & 0.4540 & 0.4837 & 0.5087 & 0.5325 & 0.5537 & 0.5533 & 0.4712 \\
ChangeFormer & 0.3132 & 0.4000 & 0.4371 & 0.4752 & 0.5199 & \underline{0.5421} & \underline{0.5738} & 0.5695 & 0.5765 & 0.4947 \\
DMINet & 0.3010 & 0.4042 & 0.4344 & 0.4722 & 0.5198 & 0.5381 & 0.5599 & 0.5793 & 0.5992 & 0.4943 \\
DTCDSCN & 0.2891 & 0.3738 & 0.4027 & 0.4335 & 0.4688 & 0.4870 & 0.4919 & 0.5081 & 0.5233 & 0.4457 \\
FC-EF & 0.2168 & 0.2770 & 0.3030 & 0.3201 & 0.3309 & 0.3203 & 0.3113 & 0.2988 & 0.3126 & 0.3046 \\
FCNPP & \underline{0.3293} & 0.4033 & 0.4310 & 0.4676 & 0.4921 & 0.5038 & 0.5308 & 0.5335 & 0.5574 & 0.4756 \\
HANet & 0.2656 & 0.3557 & 0.3988 & 0.4308 & 0.4754 & 0.5002 & 0.5230 & 0.5454 & 0.5698 & 0.4529 \\
HCGMNet & 0.2808 & 0.3778 & 0.4180 & 0.4640 & 0.4963 & 0.5283 & 0.5353 & 0.5575 & 0.5709 & 0.4733 \\
ICIFNet & 0.2840 & 0.3810 & 0.4148 & 0.4490 & 0.4888 & 0.5172 & 0.5312 & 0.5550 & 0.5921 & 0.4702 \\
MSPSNet & 0.2984 & 0.3913 & 0.4208 & 0.4578 & 0.5001 & 0.5358 & 0.5692 & 0.5699 & 0.5697 & 0.4837 \\
RDPNet & 0.2593 & 0.3551 & 0.3891 & 0.4233 & 0.4449 & 0.4553 & 0.4659 & 0.4727 & 0.5056 & 0.4232 \\
ResUnet & 0.3262 & 0.4064 & 0.4335 & 0.4744 & 0.5050 & 0.5220 & 0.5663 & 0.5762 & 0.5836 & 0.4903 \\
SNUNet & 0.2878 & 0.3776 & 0.4205 & 0.4662 & 0.4982 & 0.5176 & 0.5359 & 0.5539 & 0.5606 & 0.4738 \\
Siamunet\_Conc & 0.2471 & 0.3168 & 0.3336 & 0.3599 & 0.3757 & 0.3688 & 0.3799 & 0.3716 & 0.4182 & 0.3551 \\
SiamUnet\_Diff & 0.2079 & 0.2909 & 0.3156 & 0.3575 & 0.3912 & 0.4141 & 0.4230 & 0.4235 & 0.4468 & 0.3689 \\
TFI\_GR & 0.3132 & \underline{0.4124} & \textbf{0.4623} & \textbf{0.4995} & \textbf{0.5387} & \textbf{0.5594} & \textbf{0.5766} & 0.5867 & 0.6019 & \textbf{0.5119} \\
Tiny\_CD & \textbf{0.3381} & \textbf{0.4226} & \underline{0.4555} & \underline{0.4834} & \underline{0.5242} & 0.5399 & 0.5711 & \underline{0.5897} & \underline{0.6086} & \underline{0.5064} \\
\bottomrule
\end{tabular}}
\end{table*}

\subsubsection{Post-classification with Semantic Segmentation Methods}
We also construct a pipeline for semantic segmentation methods to enable them to detect the changes by a post-classification mechanism, which subtracts the bitemporal masks to obtain a change mask indicating the spatiotemporal variations between this period.
The results in Table \ref{tab:post_cls_seg_methods} show a clear and consistent trend: post-classification-based segmentation approaches yield noticeably lower performance across all temporal intervals when compared with dedicated change detection models. The OF1 scores remain below 0.34 for all tested networks, with Segformer achieving the highest value at 0.3317. While there is a gradual improvement in F1 as the interval between observations increases, the relative gains are limited, suggesting that the method’s capability to exploit temporal separation is constrained. This performance bottleneck can be attributed to the inherent design of post-classification pipelines, which segment each temporal snapshot independently and only later compute changes. Without an explicit mechanism to model temporal consistency, these models are prone to error propagation from per-year segmentation inaccuracies and often fail to capture subtle spatiotemporal variations—particularly in short-term intervals where change signals are weaker. These findings underscore the importance of integrating temporal dynamics directly into the model architecture for reliable change detection.
\begin{table*}[]
\caption{Quantitative comparison for cross-temporal change detection results generated by post-classification of semantic segmentation models, reported by different interval years. The overall F1 (OF1) is demonstrated in the last column by calculating the average F1 score from all samples. The best and second-best values are highlighted in bold and underlined format, respectively.}
\label{tab:post_cls_seg_methods}
\resizebox{\linewidth}{!}{%
\begin{tabular}{l|llllllllll}
\toprule
\multirow{2}{*}{Models} & \multicolumn{10}{c}{Interval Year(s)} \\
 & \multicolumn{1}{c}{1} & \multicolumn{1}{c}{2} & \multicolumn{1}{c}{3} & \multicolumn{1}{c}{4} & \multicolumn{1}{c}{5} & \multicolumn{1}{c}{6} & \multicolumn{1}{c}{7} & \multicolumn{1}{c}{8} & \multicolumn{1}{c}{9} & \multicolumn{1}{c}{OF1} \\ \midrule
DeeplabV3 (MobileViTV2)        & 0.0983 & 0.1800 & 0.2424 & 0.2922 & 0.3406 & 0.3796 & 0.4047 & 0.4428 & 0.4656 & 0.2893 \\
DeeplabV3P (MobileNetV2)       & 0.1156 & 0.2042 & 0.2629 & 0.3165 & 0.3649 & 0.4002 & 0.4237 & 0.4588 & 0.4867 & 0.3115 \\
DeeplabV3P (ResNet101)         & 0.1037 & 0.1880 & 0.2438 & 0.2962 & 0.3394 & 0.3929 & 0.4100 & 0.4355 & 0.4752 & 0.2921 \\
LinkNet                        & 0.0950 & 0.1753 & 0.2390 & 0.2892 & 0.3293 & 0.3618 & 0.3859 & 0.4251 & 0.4540 & 0.2795 \\
Mask2Former (SwinT-B)          & 0.1233 & 0.2119 & 0.2665 & 0.3182 & 0.3626 & 0.4052 & 0.4354 & 0.4717 & 0.5085 & 0.3185 \\
UNet                           & 0.1042 & 0.1847 & 0.2400 & 0.2891 & 0.3344 & 0.3809 & 0.4041 & 0.4330 & 0.4678 & 0.2890 \\
UperNet (ConvNext-B5)          & 0.1166 & 0.2096 & 0.2757 & 0.3266 & \textbf{0.3761} & \underline{0.4217} & \underline{0.4445} & \textbf{0.4853} & 0.5164 & 0.3251 \\
UperNet (SwinT-B)              & \underline{0.1261} & \underline{0.2177} & \underline{0.2767} & \underline{0.3287} & \underline{0.3741} & 0.4190 & 0.4436 & \underline{0.4791} & \underline{0.5168} & \underline{0.3297} \\
PSPNet                         & 0.0829 & 0.1524 & 0.2097 & 0.2573 & 0.2938 & 0.3314 & 0.3485 & 0.3772 & 0.4282 & 0.2490 \\
SQNet                          & 0.0947 & 0.1745 & 0.2350 & 0.2851 & 0.3264 & 0.3631 & 0.3885 & 0.4260 & 0.4712 & 0.2758 \\
Segformer                      & \textbf{0.1271} & \textbf{0.2185} & \textbf{0.2832} & \textbf{0.3319} & \textbf{0.3761} & \textbf{0.4272} & \textbf{0.4523} & 0.4765 & \textbf{0.5323} & \textbf{0.3317} \\
\bottomrule
\end{tabular}}
\end{table*}

\subsection{Discussion}
It is notable that, while slightly increasing overall between 2015 and 2024, the footprint of mining in Europe varies significantly in space and in time. Countries like Finland notably saw a notable increase in mining during the last decade, while Romania and Italy saw a significant reduction in mining activities. Interestingly, mining in Europe also endured temporal variations, such as a probable influence of the pandemic during the years 2020-2022. While it is not the scope of this manuscript to discuss the socio-political importance of EuroMineNet, we expect a diversity of users to make use of the granularity of the datasets. Indeed, the footprint of mining is vital to understand for environmental assessment, as mining significantly alters landscapes, causes deforestation, and harms biodiversity. Quantifying this footprint helps mitigate ecological damage and supports sustainable planning. Mining companies must measure and report their footprint to comply with regulations and maintain operational licenses. Analyzing the footprint improves resource efficiency, reducing waste and enhancing profitability. Understanding the footprint also helps address community concerns and fosters positive relations near mining sites. Transparent reporting builds investor confidence by meeting ESG criteria and supports ethical supply chains. Mining contributes to greenhouse gas emissions, so footprint assessment aids in climate change mitigation through carbon reduction strategies. Managing water use and preventing contamination is critical, as mining often strains local water resources. Knowledge of the mining footprint guides effective land rehabilitation after mining operations cease. Innovation driven by footprint awareness encourages cleaner technologies and sustainability leadership. Overall, managing the mining footprint balances environmental stewardship with business viability and social responsibility.
\section{Conclusion}
This study introduces EuroMineNet, the first large-scale, multispectral, multitemporal benchmark for mining footprint mapping and monitoring, addressing a long-standing gap in both mining-specific monitoring and the broader remote sensing community. Covering a decade of per-year Sentinel-2 observations for 133 mining sites across the European Union, EuroMineNet enables consistent year-by-year mapping and dynamic tracking of mining activity. We formalized two complementary tasks, multitemporal mining footprint mapping and cross-temporal change detection, alongside the proposed Change-Aware Temporal IoU (CA-TIoU) metrics, which promote temporally stable yet change-sensitive mapping. Beyond its immediate application to mining studies, EuroMineNet represents the first large-scale, multispectral, multitemporal dataset designed for change detection and land use and land cover (LULC) monitoring, offering an unparalleled resource for developing and benchmarking spatiotemporal methods. Experimental results highlight that while current deep learning models perform well for long-term change detection, consistent identification of fine-grained, short-term dynamics remains a challenge. This benchmark lays a strong foundation for advancing environmental monitoring, policy enforcement, and sustainable resource management.



\bibliographystyle{cas-model2-names}
\bibliography{reference}

\begin{thebibliography}{92}
\expandafter\ifx\csname natexlab\endcsname\relax\def\natexlab#1{#1}\fi
\providecommand{\url}[1]{\texttt{#1}}
\providecommand{\href}[2]{#2}
\providecommand{\path}[1]{#1}
\providecommand{\DOIprefix}{doi:}
\providecommand{\ArXivprefix}{arXiv:}
\providecommand{\URLprefix}{URL: }
\providecommand{\Pubmedprefix}{pmid:}
\providecommand{\doi}[1]{\href{http://dx.doi.org/#1}{\path{#1}}}
\providecommand{\Pubmed}[1]{\href{pmid:#1}{\path{#1}}}
\providecommand{\bibinfo}[2]{#2}
\ifx\xfnm\relax \def\xfnm[#1]{\unskip,\space#1}\fi
\bibitem[{Azadi et~al.(2020)Azadi, Northey, Ali and
  Edraki}]{azadi2020transparency}
\bibinfo{author}{Azadi, M.}, \bibinfo{author}{Northey, S.A.},
  \bibinfo{author}{Ali, S.H.}, \bibinfo{author}{Edraki, M.},
  \bibinfo{year}{2020}.
\newblock \bibinfo{title}{Transparency on greenhouse gas emissions from mining
  to enable climate change mitigation}.
\newblock \bibinfo{journal}{Nature Geoscience} \bibinfo{volume}{13},
  \bibinfo{pages}{100--104}.
\bibitem[{Bandara and Patel(2022)}]{bandara2022transformer}
\bibinfo{author}{Bandara, W.G.C.}, \bibinfo{author}{Patel, V.M.},
  \bibinfo{year}{2022}.
\newblock \bibinfo{title}{A transformer-based siamese network for change
  detection}, in: \bibinfo{booktitle}{IGARSS 2022-2022 IEEE International
  Geoscience and Remote Sensing Symposium}, \bibinfo{organization}{IEEE}. pp.
  \bibinfo{pages}{207--210}.
\bibitem[{Bruzzone and Prieto(2002)}]{bruzzone2002automatic}
\bibinfo{author}{Bruzzone, L.}, \bibinfo{author}{Prieto, D.F.},
  \bibinfo{year}{2002}.
\newblock \bibinfo{title}{Automatic analysis of the difference image for
  unsupervised change detection}.
\newblock \bibinfo{journal}{IEEE Transactions on Geoscience and Remote sensing}
  \bibinfo{volume}{38}, \bibinfo{pages}{1171--1182}.
\bibitem[{Camalan et~al.(2022)Camalan, Cui, Pauca, Alqahtani, Silman, Chan,
  Plemmons, Dethier, Fernandez and Lutz}]{camalan2022change}
\bibinfo{author}{Camalan, S.}, \bibinfo{author}{Cui, K.},
  \bibinfo{author}{Pauca, V.P.}, \bibinfo{author}{Alqahtani, S.},
  \bibinfo{author}{Silman, M.}, \bibinfo{author}{Chan, R.},
  \bibinfo{author}{Plemmons, R.J.}, \bibinfo{author}{Dethier, E.N.},
  \bibinfo{author}{Fernandez, L.E.}, \bibinfo{author}{Lutz, D.A.},
  \bibinfo{year}{2022}.
\newblock \bibinfo{title}{Change detection of amazonian alluvial gold mining
  using deep learning and sentinel-2 imagery}.
\newblock \bibinfo{journal}{Remote Sensing} \bibinfo{volume}{14},
  \bibinfo{pages}{1746}.
\bibitem[{Charou et~al.(2010)Charou, Stefouli, Dimitrakopoulos, Vasiliou and
  Mavrantza}]{charou2010using}
\bibinfo{author}{Charou, E.}, \bibinfo{author}{Stefouli, M.},
  \bibinfo{author}{Dimitrakopoulos, D.}, \bibinfo{author}{Vasiliou, E.},
  \bibinfo{author}{Mavrantza, O.}, \bibinfo{year}{2010}.
\newblock \bibinfo{title}{Using remote sensing to assess impact of mining
  activities on land and water resources}.
\newblock \bibinfo{journal}{Mine Water and the Environment}
  \bibinfo{volume}{29}, \bibinfo{pages}{45--52}.
\bibitem[{Chaurasia and Culurciello(2017)}]{chaurasia2017linknet}
\bibinfo{author}{Chaurasia, A.}, \bibinfo{author}{Culurciello, E.},
  \bibinfo{year}{2017}.
\newblock \bibinfo{title}{Linknet: Exploiting encoder representations for
  efficient semantic segmentation}, in: \bibinfo{booktitle}{2017 IEEE visual
  communications and image processing (VCIP)}, \bibinfo{organization}{IEEE}.
  pp. \bibinfo{pages}{1--4}.
\bibitem[{Chen et~al.(2022a)Chen, Pu, Yang, Tang and Xu}]{chen2022rdp}
\bibinfo{author}{Chen, H.}, \bibinfo{author}{Pu, F.}, \bibinfo{author}{Yang,
  R.}, \bibinfo{author}{Tang, R.}, \bibinfo{author}{Xu, X.},
  \bibinfo{year}{2022}a.
\newblock \bibinfo{title}{Rdp-net: Region detail preserving network for change
  detection}.
\newblock \bibinfo{journal}{IEEE Trans. Geosci. Remote Sens.}
  \bibinfo{volume}{60}, \bibinfo{pages}{1--10}.
\bibitem[{Chen et~al.(2021)Chen, Qi and Shi}]{chen2021remote}
\bibinfo{author}{Chen, H.}, \bibinfo{author}{Qi, Z.}, \bibinfo{author}{Shi,
  Z.}, \bibinfo{year}{2021}.
\newblock \bibinfo{title}{Remote sensing image change detection with
  transformers}.
\newblock \bibinfo{journal}{IEEE Transactions on Geoscience and Remote Sensing}
  \bibinfo{volume}{60}, \bibinfo{pages}{1--14}.
\bibitem[{Chen et~al.(2017)Chen, Papandreou, Schroff and
  Adam}]{chen2017rethinking}
\bibinfo{author}{Chen, L.C.}, \bibinfo{author}{Papandreou, G.},
  \bibinfo{author}{Schroff, F.}, \bibinfo{author}{Adam, H.},
  \bibinfo{year}{2017}.
\newblock \bibinfo{title}{Rethinking atrous convolution for semantic image
  segmentation}.
\newblock \bibinfo{journal}{arXiv preprint arXiv:1706.05587} .
\bibitem[{Chen et~al.(2018)Chen, Zhu, Papandreou, Schroff and
  Adam}]{chen2018encoder}
\bibinfo{author}{Chen, L.C.}, \bibinfo{author}{Zhu, Y.},
  \bibinfo{author}{Papandreou, G.}, \bibinfo{author}{Schroff, F.},
  \bibinfo{author}{Adam, H.}, \bibinfo{year}{2018}.
\newblock \bibinfo{title}{Encoder-decoder with atrous separable convolution for
  semantic image segmentation}, in: \bibinfo{booktitle}{Proceedings of the
  European conference on computer vision (ECCV)}, pp.
  \bibinfo{pages}{801--818}.
\bibitem[{Chen et~al.(2022b)Chen, Zheng, Niu and Plaza}]{chen2022open}
\bibinfo{author}{Chen, T.}, \bibinfo{author}{Zheng, X.}, \bibinfo{author}{Niu,
  R.}, \bibinfo{author}{Plaza, A.}, \bibinfo{year}{2022}b.
\newblock \bibinfo{title}{Open-pit mine area mapping with gaofen-2 satellite
  images using u-net+}.
\newblock \bibinfo{journal}{IEEE Journal of Selected Topics in Applied Earth
  Observations and Remote Sensing} \bibinfo{volume}{15},
  \bibinfo{pages}{3589--3599}.
\bibitem[{Chen et~al.(2022c)Chen, Zhou, Wang, Xu, He, Jin and
  Jin}]{chen2022egde}
\bibinfo{author}{Chen, Z.}, \bibinfo{author}{Zhou, Y.}, \bibinfo{author}{Wang,
  B.}, \bibinfo{author}{Xu, X.}, \bibinfo{author}{He, N.},
  \bibinfo{author}{Jin, S.}, \bibinfo{author}{Jin, S.}, \bibinfo{year}{2022}c.
\newblock \bibinfo{title}{Egde-net: A building change detection method for
  high-resolution remote sensing imagery based on edge guidance and
  differential enhancement}.
\newblock \bibinfo{journal}{ISPRS Journal of Photogrammetry and Remote Sensing}
  \bibinfo{volume}{191}, \bibinfo{pages}{203--222}.
\bibitem[{Cheng et~al.(2022)Cheng, Misra, Schwing, Kirillov and
  Girdhar}]{cheng2021mask2former}
\bibinfo{author}{Cheng, B.}, \bibinfo{author}{Misra, I.},
  \bibinfo{author}{Schwing, A.G.}, \bibinfo{author}{Kirillov, A.},
  \bibinfo{author}{Girdhar, R.}, \bibinfo{year}{2022}.
\newblock \bibinfo{title}{Masked-attention mask transformer for universal image
  segmentation}.
\bibitem[{Cheng et~al.(2024)Cheng, Huang, Li, Lyu, Xu, Zhao, Zhao and
  Xiang}]{cheng2024change}
\bibinfo{author}{Cheng, G.}, \bibinfo{author}{Huang, Y.}, \bibinfo{author}{Li,
  X.}, \bibinfo{author}{Lyu, S.}, \bibinfo{author}{Xu, Z.},
  \bibinfo{author}{Zhao, H.}, \bibinfo{author}{Zhao, Q.},
  \bibinfo{author}{Xiang, S.}, \bibinfo{year}{2024}.
\newblock \bibinfo{title}{Change detection methods for remote sensing in the
  last decade: A comprehensive review}.
\newblock \bibinfo{journal}{Remote Sensing} \bibinfo{volume}{16},
  \bibinfo{pages}{2355}.
\bibitem[{Codegoni et~al.(2023)Codegoni, Lombardi and
  Ferrari}]{codegoni2023tinycd}
\bibinfo{author}{Codegoni, A.}, \bibinfo{author}{Lombardi, G.},
  \bibinfo{author}{Ferrari, A.}, \bibinfo{year}{2023}.
\newblock \bibinfo{title}{Tinycd: A (not so) deep learning model for change
  detection}.
\newblock \bibinfo{journal}{Neural Computing and Applications}
  \bibinfo{volume}{35}, \bibinfo{pages}{8471--8486}.
\bibitem[{Cohen et~al.(2018)Cohen, Yang, Healey, Kennedy and
  Gorelick}]{cohen2018landtrendr}
\bibinfo{author}{Cohen, W.B.}, \bibinfo{author}{Yang, Z.},
  \bibinfo{author}{Healey, S.P.}, \bibinfo{author}{Kennedy, R.E.},
  \bibinfo{author}{Gorelick, N.}, \bibinfo{year}{2018}.
\newblock \bibinfo{title}{A landtrendr multispectral ensemble for forest
  disturbance detection}.
\newblock \bibinfo{journal}{Remote sensing of environment}
  \bibinfo{volume}{205}, \bibinfo{pages}{131--140}.
\bibitem[{Daudt et~al.(2018a)Daudt, Le~Saux and Boulch}]{daudt2018fully}
\bibinfo{author}{Daudt, R.C.}, \bibinfo{author}{Le~Saux, B.},
  \bibinfo{author}{Boulch, A.}, \bibinfo{year}{2018}a.
\newblock \bibinfo{title}{Fully convolutional siamese networks for change
  detection}, in: \bibinfo{booktitle}{2018 25th IEEE international conference
  on image processing (ICIP)}, \bibinfo{organization}{IEEE}. pp.
  \bibinfo{pages}{4063--4067}.
\bibitem[{Daudt et~al.(2018b)Daudt, Le~Saux, Boulch and
  Gousseau}]{daudt2018urban}
\bibinfo{author}{Daudt, R.C.}, \bibinfo{author}{Le~Saux, B.},
  \bibinfo{author}{Boulch, A.}, \bibinfo{author}{Gousseau, Y.},
  \bibinfo{year}{2018}b.
\newblock \bibinfo{title}{Urban change detection for multispectral earth
  observation using convolutional neural networks}, in:
  \bibinfo{booktitle}{IGARSS 2018-2018 IEEE International Geoscience and Remote
  Sensing Symposium}, \bibinfo{organization}{Ieee}. pp.
  \bibinfo{pages}{2115--2118}.
\bibitem[{Dehkordi et~al.(2024)Dehkordi, Nodeh, Dehkordi, Khorjestan,
  Ghaffarzadeh et~al.}]{dehkordi2024soil}
\bibinfo{author}{Dehkordi, M.M.}, \bibinfo{author}{Nodeh, Z.P.},
  \bibinfo{author}{Dehkordi, K.S.}, \bibinfo{author}{Khorjestan, R.R.},
  \bibinfo{author}{Ghaffarzadeh, M.}, et~al., \bibinfo{year}{2024}.
\newblock \bibinfo{title}{Soil, air, and water pollution from mining and
  industrial activities: Sources of pollution, environmental impacts, and
  prevention and control methods}.
\newblock \bibinfo{journal}{Results in Engineering} \bibinfo{volume}{23},
  \bibinfo{pages}{102729}.
\bibitem[{Dube et~al.(2024)Dube, Dube, Dalu, Gxokwe and
  Marambanyika}]{dube2024assessment}
\bibinfo{author}{Dube, T.}, \bibinfo{author}{Dube, T.}, \bibinfo{author}{Dalu,
  T.}, \bibinfo{author}{Gxokwe, S.}, \bibinfo{author}{Marambanyika, T.},
  \bibinfo{year}{2024}.
\newblock \bibinfo{title}{Assessment of land use and land cover, water nutrient
  and metal concentration related to illegal mining activities in an austral
  semi--arid river system: A remote sensing and multivariate analysis
  approach}.
\newblock \bibinfo{journal}{Science of The Total Environment}
  \bibinfo{volume}{907}, \bibinfo{pages}{167919}.
\bibitem[{Fang et~al.(2021)Fang, Li, Shao and Li}]{fang2021snunet}
\bibinfo{author}{Fang, S.}, \bibinfo{author}{Li, K.}, \bibinfo{author}{Shao,
  J.}, \bibinfo{author}{Li, Z.}, \bibinfo{year}{2021}.
\newblock \bibinfo{title}{Snunet-cd: A densely connected siamese network for
  change detection of vhr images}.
\newblock \bibinfo{journal}{IEEE Geoscience and Remote Sensing Letters}
  \bibinfo{volume}{19}, \bibinfo{pages}{1--5}.
\bibitem[{Feng et~al.(2023)Feng, Jiang, Xu and Zheng}]{feng2023change}
\bibinfo{author}{Feng, Y.}, \bibinfo{author}{Jiang, J.}, \bibinfo{author}{Xu,
  H.}, \bibinfo{author}{Zheng, J.}, \bibinfo{year}{2023}.
\newblock \bibinfo{title}{Change detection on remote sensing images using
  dual-branch multilevel intertemporal network}.
\newblock \bibinfo{journal}{IEEE Trans. Geosci. Remote Sens.}
  \bibinfo{volume}{61}, \bibinfo{pages}{1--15}.
\bibitem[{Feng et~al.(2022)Feng, Xu, Jiang, Liu and Zheng}]{feng2022icif}
\bibinfo{author}{Feng, Y.}, \bibinfo{author}{Xu, H.}, \bibinfo{author}{Jiang,
  J.}, \bibinfo{author}{Liu, H.}, \bibinfo{author}{Zheng, J.},
  \bibinfo{year}{2022}.
\newblock \bibinfo{title}{Icif-net: Intra-scale cross-interaction and
  inter-scale feature fusion network for bitemporal remote sensing images
  change detection}.
\newblock \bibinfo{journal}{IEEE Trans. Geosci. Remote Sens.}
  \bibinfo{volume}{60}, \bibinfo{pages}{1--13}.
\bibitem[{Firozjaei et~al.(2021)Firozjaei, Sedighi, Firozjaei, Kiavarz, Homaee,
  Arsanjani, Makki, Naimi and Alavipanah}]{firozjaei2021historical}
\bibinfo{author}{Firozjaei, M.K.}, \bibinfo{author}{Sedighi, A.},
  \bibinfo{author}{Firozjaei, H.K.}, \bibinfo{author}{Kiavarz, M.},
  \bibinfo{author}{Homaee, M.}, \bibinfo{author}{Arsanjani, J.J.},
  \bibinfo{author}{Makki, M.}, \bibinfo{author}{Naimi, B.},
  \bibinfo{author}{Alavipanah, S.K.}, \bibinfo{year}{2021}.
\newblock \bibinfo{title}{A historical and future impact assessment of mining
  activities on surface biophysical characteristics change: A remote
  sensing-based approach}.
\newblock \bibinfo{journal}{Ecological Indicators} \bibinfo{volume}{122},
  \bibinfo{pages}{107264}.
\bibitem[{Fu et~al.(2024)Fu, Zhu, Liu, Zhan, He, Shen, Zhao, Liu, Zhang, Liu
  et~al.}]{fu2024remote}
\bibinfo{author}{Fu, Y.}, \bibinfo{author}{Zhu, Z.}, \bibinfo{author}{Liu, L.},
  \bibinfo{author}{Zhan, W.}, \bibinfo{author}{He, T.}, \bibinfo{author}{Shen,
  H.}, \bibinfo{author}{Zhao, J.}, \bibinfo{author}{Liu, Y.},
  \bibinfo{author}{Zhang, H.}, \bibinfo{author}{Liu, Z.}, et~al.,
  \bibinfo{year}{2024}.
\newblock \bibinfo{title}{Remote sensing time series analysis: A review of data
  and applications}.
\newblock \bibinfo{journal}{Journal of Remote Sensing} \bibinfo{volume}{4},
  \bibinfo{pages}{0285}.
\bibitem[{Ghamisi et~al.(2021)Ghamisi, Shahi, Duan, Rasti, Lorenz, Booysen,
  Thiele, Contreras, Kirsch and Gloaguen}]{ghamisi2021potential}
\bibinfo{author}{Ghamisi, P.}, \bibinfo{author}{Shahi, K.R.},
  \bibinfo{author}{Duan, P.}, \bibinfo{author}{Rasti, B.},
  \bibinfo{author}{Lorenz, S.}, \bibinfo{author}{Booysen, R.},
  \bibinfo{author}{Thiele, S.}, \bibinfo{author}{Contreras, I.C.},
  \bibinfo{author}{Kirsch, M.}, \bibinfo{author}{Gloaguen, R.},
  \bibinfo{year}{2021}.
\newblock \bibinfo{title}{The potential of machine learning for a more
  responsible sourcing of critical raw materials}.
\newblock \bibinfo{journal}{IEEE Journal of Selected Topics in Applied Earth
  Observations and Remote Sensing} \bibinfo{volume}{14},
  \bibinfo{pages}{8971--8988}.
\bibitem[{Ghamisi et~al.(2025)Ghamisi, Yu, Marinoni, Gevaert, Persello,
  Selvakumaran, Girotto, Horton, Rufin, Hostert
  et~al.}]{ghamisi2025responsible}
\bibinfo{author}{Ghamisi, P.}, \bibinfo{author}{Yu, W.},
  \bibinfo{author}{Marinoni, A.}, \bibinfo{author}{Gevaert, C.M.},
  \bibinfo{author}{Persello, C.}, \bibinfo{author}{Selvakumaran, S.},
  \bibinfo{author}{Girotto, M.}, \bibinfo{author}{Horton, B.P.},
  \bibinfo{author}{Rufin, P.}, \bibinfo{author}{Hostert, P.}, et~al.,
  \bibinfo{year}{2025}.
\newblock \bibinfo{title}{Responsible artificial intelligence for earth
  observation: Achievable and realistic paths to serve the collective good}.
\newblock \bibinfo{journal}{IEEE Geoscience and Remote Sensing Magazine} .
\bibitem[{Giljum et~al.(2025)Giljum, Maus, Sonter, Luckeneder, Werner, Lutter,
  Gershenzon, Cole, Siqueira-Gay and Bebbington}]{giljum2025metal}
\bibinfo{author}{Giljum, S.}, \bibinfo{author}{Maus, V.},
  \bibinfo{author}{Sonter, L.}, \bibinfo{author}{Luckeneder, S.},
  \bibinfo{author}{Werner, T.}, \bibinfo{author}{Lutter, S.},
  \bibinfo{author}{Gershenzon, J.}, \bibinfo{author}{Cole, M.J.},
  \bibinfo{author}{Siqueira-Gay, J.}, \bibinfo{author}{Bebbington, A.},
  \bibinfo{year}{2025}.
\newblock \bibinfo{title}{Metal mining is a global driver of environmental
  change}.
\newblock \bibinfo{journal}{Nature Reviews Earth \& Environment} ,
  \bibinfo{pages}{1--15}.
\bibitem[{Gugger et~al.(2022)Gugger, Debut, Wolf, Schmid, Mueller, Mangrulkar,
  Sun and Bossan}]{accelerate}
\bibinfo{author}{Gugger, S.}, \bibinfo{author}{Debut, L.},
  \bibinfo{author}{Wolf, T.}, \bibinfo{author}{Schmid, P.},
  \bibinfo{author}{Mueller, Z.}, \bibinfo{author}{Mangrulkar, S.},
  \bibinfo{author}{Sun, M.}, \bibinfo{author}{Bossan, B.},
  \bibinfo{year}{2022}.
\newblock \bibinfo{title}{Accelerate: Training and inference at scale made
  simple, efficient and adaptable.}
\newblock
  \bibinfo{howpublished}{\url{https://github.com/huggingface/accelerate}}.
\bibitem[{Guo et~al.(2021)Guo, Zhang, Zhu, Zhong and Zhang}]{guo2021deep}
\bibinfo{author}{Guo, Q.}, \bibinfo{author}{Zhang, J.}, \bibinfo{author}{Zhu,
  S.}, \bibinfo{author}{Zhong, C.}, \bibinfo{author}{Zhang, Y.},
  \bibinfo{year}{2021}.
\newblock \bibinfo{title}{Deep multiscale siamese network with parallel
  convolutional structure and self-attention for change detection}.
\newblock \bibinfo{journal}{IEEE Trans. Geosci. Remote Sens.}
  \bibinfo{volume}{60}, \bibinfo{pages}{1--12}.
\bibitem[{Han et~al.(2023a)Han, Wu and Du}]{han2023hcgmnet}
\bibinfo{author}{Han, C.}, \bibinfo{author}{Wu, C.}, \bibinfo{author}{Du, B.},
  \bibinfo{year}{2023}a.
\newblock \bibinfo{title}{Hcgmnet: A hierarchical change guiding map network
  for change detection}, in: \bibinfo{booktitle}{IGARSS 2023-2023 IEEE
  International Geoscience and Remote Sensing Symposium},
  \bibinfo{organization}{IEEE}. pp. \bibinfo{pages}{5511--5514}.
\bibitem[{Han et~al.(2023b)Han, Wu, Guo, Hu and Chen}]{han2023hanet}
\bibinfo{author}{Han, C.}, \bibinfo{author}{Wu, C.}, \bibinfo{author}{Guo, H.},
  \bibinfo{author}{Hu, M.}, \bibinfo{author}{Chen, H.}, \bibinfo{year}{2023}b.
\newblock \bibinfo{title}{Hanet: A hierarchical attention network for change
  detection with bitemporal very-high-resolution remote sensing images}.
\newblock \bibinfo{journal}{IEEE Journal of Selected Topics in Applied Earth
  Observations and Remote Sensing} \bibinfo{volume}{16},
  \bibinfo{pages}{3867--3878}.
\bibitem[{Han et~al.(2023c)Han, Wu, Guo, Hu, Li and Chen}]{han2023change}
\bibinfo{author}{Han, C.}, \bibinfo{author}{Wu, C.}, \bibinfo{author}{Guo, H.},
  \bibinfo{author}{Hu, M.}, \bibinfo{author}{Li, J.}, \bibinfo{author}{Chen,
  H.}, \bibinfo{year}{2023}c.
\newblock \bibinfo{title}{Change guiding network: Incorporating change prior to
  guide change detection in remote sensing imagery}.
\newblock \bibinfo{journal}{IEEE Journal of Selected Topics in Applied Earth
  Observations and Remote Sensing} .
\bibitem[{He et~al.(2011)He, Wei, Shi, Zhang and Zhao}]{he2011detecting}
\bibinfo{author}{He, C.}, \bibinfo{author}{Wei, A.}, \bibinfo{author}{Shi, P.},
  \bibinfo{author}{Zhang, Q.}, \bibinfo{author}{Zhao, Y.},
  \bibinfo{year}{2011}.
\newblock \bibinfo{title}{Detecting land-use/land-cover change in rural--urban
  fringe areas using extended change-vector analysis}.
\newblock \bibinfo{journal}{International Journal of Applied Earth Observation
  and Geoinformation} \bibinfo{volume}{13}, \bibinfo{pages}{572--585}.
\bibitem[{He et~al.(2016)He, Zhang, Ren and Sun}]{he2016deep}
\bibinfo{author}{He, K.}, \bibinfo{author}{Zhang, X.}, \bibinfo{author}{Ren,
  S.}, \bibinfo{author}{Sun, J.}, \bibinfo{year}{2016}.
\newblock \bibinfo{title}{Deep residual learning for image recognition}, in:
  \bibinfo{booktitle}{Proceedings of the IEEE conference on computer vision and
  pattern recognition}, pp. \bibinfo{pages}{770--778}.
\bibitem[{Hu et~al.(2018)Hu, Dong et~al.}]{hu2018automatic}
\bibinfo{author}{Hu, Y.}, \bibinfo{author}{Dong, Y.}, et~al.,
  \bibinfo{year}{2018}.
\newblock \bibinfo{title}{An automatic approach for land-change detection and
  land updates based on integrated ndvi timing analysis and the cvaps method
  with gee support}.
\newblock \bibinfo{journal}{ISPRS journal of photogrammetry and remote sensing}
  \bibinfo{volume}{146}, \bibinfo{pages}{347--359}.
\bibitem[{Jab{\l}o{\'n}ska et~al.(2024)Jab{\l}o{\'n}ska, Maksymowicz,
  Tanajewski, Kaczan, Zi{\k{e}}ba and Wilgucki}]{jablonska2024minecam}
\bibinfo{author}{Jab{\l}o{\'n}ska, K.}, \bibinfo{author}{Maksymowicz, M.},
  \bibinfo{author}{Tanajewski, D.}, \bibinfo{author}{Kaczan, W.},
  \bibinfo{author}{Zi{\k{e}}ba, M.}, \bibinfo{author}{Wilgucki, M.},
  \bibinfo{year}{2024}.
\newblock \bibinfo{title}{Minecam: Application of combined remote sensing and
  machine learning for segmentation and change detection of mining areas
  enabling multi-purpose monitoring}.
\newblock \bibinfo{journal}{Remote Sensing} \bibinfo{volume}{16},
  \bibinfo{pages}{955}.
\bibitem[{Jain(2015)}]{jain2015environmental}
\bibinfo{author}{Jain, R.}, \bibinfo{year}{2015}.
\newblock \bibinfo{title}{Environmental impact of mining and mineral
  processing: management, monitoring, and auditing strategies}.
\newblock \bibinfo{publisher}{Butterworth-Heinemann}.
\bibitem[{Kumar and Gorai(2023)}]{kumar2023development}
\bibinfo{author}{Kumar, A.}, \bibinfo{author}{Gorai, A.K.},
  \bibinfo{year}{2023}.
\newblock \bibinfo{title}{Development of a deep convolutional neural network
  model for detection and delineation of coal mining regions}.
\newblock \bibinfo{journal}{Earth Science Informatics} \bibinfo{volume}{16},
  \bibinfo{pages}{1151--1171}.
\bibitem[{L{\`e}bre et~al.(2020)L{\`e}bre, Stringer, Svobodova, Owen, Kemp,
  C{\^o}te, Arratia-Solar and Valenta}]{lebre2020social}
\bibinfo{author}{L{\`e}bre, {\'E}.}, \bibinfo{author}{Stringer, M.},
  \bibinfo{author}{Svobodova, K.}, \bibinfo{author}{Owen, J.R.},
  \bibinfo{author}{Kemp, D.}, \bibinfo{author}{C{\^o}te, C.},
  \bibinfo{author}{Arratia-Solar, A.}, \bibinfo{author}{Valenta, R.K.},
  \bibinfo{year}{2020}.
\newblock \bibinfo{title}{The social and environmental complexities of
  extracting energy transition metals}.
\newblock \bibinfo{journal}{Nature communications} \bibinfo{volume}{11},
  \bibinfo{pages}{4823}.
\bibitem[{Leenstra et~al.(2021)Leenstra, Marcos, Bovolo and
  Tuia}]{leenstra2021self}
\bibinfo{author}{Leenstra, M.}, \bibinfo{author}{Marcos, D.},
  \bibinfo{author}{Bovolo, F.}, \bibinfo{author}{Tuia, D.},
  \bibinfo{year}{2021}.
\newblock \bibinfo{title}{Self-supervised pre-training enhances change
  detection in sentinel-2 imagery}, in: \bibinfo{booktitle}{International
  conference on pattern recognition}, \bibinfo{organization}{Springer}. pp.
  \bibinfo{pages}{578--590}.
\bibitem[{Lei et~al.(2019)Lei, Zhang, Lv, Li, Liu and Nandi}]{lei2019landslide}
\bibinfo{author}{Lei, T.}, \bibinfo{author}{Zhang, Y.}, \bibinfo{author}{Lv,
  Z.}, \bibinfo{author}{Li, S.}, \bibinfo{author}{Liu, S.},
  \bibinfo{author}{Nandi, A.K.}, \bibinfo{year}{2019}.
\newblock \bibinfo{title}{Landslide inventory mapping from bitemporal images
  using deep convolutional neural networks}.
\newblock \bibinfo{journal}{IEEE Geoscience and Remote Sensing Letters}
  \bibinfo{volume}{16}, \bibinfo{pages}{982--986}.
\bibitem[{Li et~al.(2025)Li, Wang, Werner, Chen and Chen}]{li2025machine}
\bibinfo{author}{Li, H.}, \bibinfo{author}{Wang, P.}, \bibinfo{author}{Werner,
  T.T.}, \bibinfo{author}{Chen, B.}, \bibinfo{author}{Chen, W.Q.},
  \bibinfo{year}{2025}.
\newblock \bibinfo{title}{Machine learning-enhanced monitoring of global copper
  mining areas}.
\newblock \bibinfo{journal}{Scientific Data} \bibinfo{volume}{12},
  \bibinfo{pages}{1120}.
\bibitem[{Li et~al.(2022a)Li, Xing, Du, Du, Zhang and Li}]{li2022change}
\bibinfo{author}{Li, J.}, \bibinfo{author}{Xing, J.}, \bibinfo{author}{Du, S.},
  \bibinfo{author}{Du, S.}, \bibinfo{author}{Zhang, C.}, \bibinfo{author}{Li,
  W.}, \bibinfo{year}{2022}a.
\newblock \bibinfo{title}{Change detection of open-pit mine based on siamese
  multiscale network}.
\newblock \bibinfo{journal}{IEEE Geoscience and Remote Sensing Letters}
  \bibinfo{volume}{20}, \bibinfo{pages}{1--5}.
\bibitem[{Li et~al.(2022b)Li, Zhong, Du and Du}]{li2022transunetcd}
\bibinfo{author}{Li, Q.}, \bibinfo{author}{Zhong, R.}, \bibinfo{author}{Du,
  X.}, \bibinfo{author}{Du, Y.}, \bibinfo{year}{2022}b.
\newblock \bibinfo{title}{Transunetcd: A hybrid transformer network for change
  detection in optical remote-sensing images}.
\newblock \bibinfo{journal}{IEEE Transactions on Geoscience and Remote Sensing}
  \bibinfo{volume}{60}, \bibinfo{pages}{1--19}.
\bibitem[{Li et~al.(2023)Li, Tang, Liu, Zhang, Dou, Wang and
  Zomaya}]{Li_2023_A2Net}
\bibinfo{author}{Li, Z.}, \bibinfo{author}{Tang, C.}, \bibinfo{author}{Liu,
  X.}, \bibinfo{author}{Zhang, W.}, \bibinfo{author}{Dou, J.},
  \bibinfo{author}{Wang, L.}, \bibinfo{author}{Zomaya, A.Y.},
  \bibinfo{year}{2023}.
\newblock \bibinfo{title}{Lightweight remote sensing change detection with
  progressive feature aggregation and supervised attention}.
\newblock \bibinfo{journal}{IEEE Trans. Geosci. Remote Sens.}
  \bibinfo{volume}{61}, \bibinfo{pages}{1--12}.
\newblock \DOIprefix\doi{10.1109/TGRS.2023.3241436}.
\bibitem[{Li et~al.(2022c)Li, Tang, Wang and Zomaya}]{li2022remote}
\bibinfo{author}{Li, Z.}, \bibinfo{author}{Tang, C.}, \bibinfo{author}{Wang,
  L.}, \bibinfo{author}{Zomaya, A.Y.}, \bibinfo{year}{2022}c.
\newblock \bibinfo{title}{Remote sensing change detection via temporal feature
  interaction and guided refinement}.
\newblock \bibinfo{journal}{IEEE Trans. Geosci. Remote Sens.}
  \bibinfo{volume}{60}, \bibinfo{pages}{1--11}.
\bibitem[{Liu et~al.(2020)Liu, Pang, Zhan, Zhang and Yang}]{liu2020building}
\bibinfo{author}{Liu, Y.}, \bibinfo{author}{Pang, C.}, \bibinfo{author}{Zhan,
  Z.}, \bibinfo{author}{Zhang, X.}, \bibinfo{author}{Yang, X.},
  \bibinfo{year}{2020}.
\newblock \bibinfo{title}{Building change detection for remote sensing images
  using a dual-task constrained deep siamese convolutional network model}.
\newblock \bibinfo{journal}{IEEE Geoscience and Remote Sensing Letters}
  \bibinfo{volume}{18}, \bibinfo{pages}{811--815}.
\bibitem[{Liu et~al.(2021a)Liu, Wang, Gojenko, Yu, Wei, Luo and
  Xiao}]{liu2021review}
\bibinfo{author}{Liu, Y.}, \bibinfo{author}{Wang, P.},
  \bibinfo{author}{Gojenko, B.}, \bibinfo{author}{Yu, J.},
  \bibinfo{author}{Wei, L.}, \bibinfo{author}{Luo, D.}, \bibinfo{author}{Xiao,
  T.}, \bibinfo{year}{2021}a.
\newblock \bibinfo{title}{A review of water pollution arising from agriculture
  and mining activities in central asia: Facts, causes and effects}.
\newblock \bibinfo{journal}{Environmental Pollution} \bibinfo{volume}{291},
  \bibinfo{pages}{118209}.
\bibitem[{Liu et~al.(2021b)Liu, Lin, Cao, Hu, Wei, Zhang, Lin and
  Guo}]{liu2021swin}
\bibinfo{author}{Liu, Z.}, \bibinfo{author}{Lin, Y.}, \bibinfo{author}{Cao,
  Y.}, \bibinfo{author}{Hu, H.}, \bibinfo{author}{Wei, Y.},
  \bibinfo{author}{Zhang, Z.}, \bibinfo{author}{Lin, S.}, \bibinfo{author}{Guo,
  B.}, \bibinfo{year}{2021}b.
\newblock \bibinfo{title}{Swin transformer: Hierarchical vision transformer
  using shifted windows}, in: \bibinfo{booktitle}{Proceedings of the IEEE/CVF
  international conference on computer vision}, pp.
  \bibinfo{pages}{10012--10022}.
\bibitem[{Maus et~al.(2020)Maus, Giljum, Gutschlhofer, da~Silva, Probst, Gass,
  Luckeneder, Lieber and McCallum}]{maus2020global}
\bibinfo{author}{Maus, V.}, \bibinfo{author}{Giljum, S.},
  \bibinfo{author}{Gutschlhofer, J.}, \bibinfo{author}{da~Silva, D.M.},
  \bibinfo{author}{Probst, M.}, \bibinfo{author}{Gass, S.L.},
  \bibinfo{author}{Luckeneder, S.}, \bibinfo{author}{Lieber, M.},
  \bibinfo{author}{McCallum, I.}, \bibinfo{year}{2020}.
\newblock \bibinfo{title}{A global-scale data set of mining areas}.
\newblock \bibinfo{journal}{Scientific data} \bibinfo{volume}{7},
  \bibinfo{pages}{289}.
\bibitem[{Maus et~al.(2022)Maus, Giljum, da~Silva, Gutschlhofer, da~Rosa,
  Luckeneder, Gass, Lieber and McCallum}]{maus2022update}
\bibinfo{author}{Maus, V.}, \bibinfo{author}{Giljum, S.},
  \bibinfo{author}{da~Silva, D.M.}, \bibinfo{author}{Gutschlhofer, J.},
  \bibinfo{author}{da~Rosa, R.P.}, \bibinfo{author}{Luckeneder, S.},
  \bibinfo{author}{Gass, S.L.}, \bibinfo{author}{Lieber, M.},
  \bibinfo{author}{McCallum, I.}, \bibinfo{year}{2022}.
\newblock \bibinfo{title}{An update on global mining land use}.
\newblock \bibinfo{journal}{Scientific data} \bibinfo{volume}{9},
  \bibinfo{pages}{433}.
\bibitem[{Mehta and Rastegari(2022)}]{mehta2022separable}
\bibinfo{author}{Mehta, S.}, \bibinfo{author}{Rastegari, M.},
  \bibinfo{year}{2022}.
\newblock \bibinfo{title}{Separable self-attention for mobile vision
  transformers}.
\newblock \bibinfo{journal}{arXiv preprint arXiv:2206.02680} .
\bibitem[{Ning et~al.(2024)Ning, Zhang, Zhang and Huang}]{ning2024multi}
\bibinfo{author}{Ning, X.}, \bibinfo{author}{Zhang, H.},
  \bibinfo{author}{Zhang, R.}, \bibinfo{author}{Huang, X.},
  \bibinfo{year}{2024}.
\newblock \bibinfo{title}{Multi-stage progressive change detection on high
  resolution remote sensing imagery}.
\newblock \bibinfo{journal}{ISPRS Journal of Photogrammetry and Remote Sensing}
  \bibinfo{volume}{207}, \bibinfo{pages}{231--244}.
\bibitem[{Padmanaban et~al.(2017)Padmanaban, Bhowmik and
  Cabral}]{padmanaban2017remote}
\bibinfo{author}{Padmanaban, R.}, \bibinfo{author}{Bhowmik, A.K.},
  \bibinfo{author}{Cabral, P.}, \bibinfo{year}{2017}.
\newblock \bibinfo{title}{A remote sensing approach to environmental monitoring
  in a reclaimed mine area}.
\newblock \bibinfo{journal}{ISPRS international journal of geo-information}
  \bibinfo{volume}{6}, \bibinfo{pages}{401}.
\bibitem[{Pan et~al.(2023)Pan, Lin, Zang, Long, Zhang, Xu and
  Jiang}]{pan2023new}
\bibinfo{author}{Pan, Y.}, \bibinfo{author}{Lin, H.}, \bibinfo{author}{Zang,
  Z.}, \bibinfo{author}{Long, J.}, \bibinfo{author}{Zhang, M.},
  \bibinfo{author}{Xu, X.}, \bibinfo{author}{Jiang, W.}, \bibinfo{year}{2023}.
\newblock \bibinfo{title}{A new change detection method for wetlands based on
  bi-temporal semantic reasoning unet++ in dongting lake, china}.
\newblock \bibinfo{journal}{Ecological Indicators} \bibinfo{volume}{155},
  \bibinfo{pages}{110997}.
\bibitem[{Pelletier et~al.(2024)Pelletier, Cardille, Wulder, White and
  Hermosilla}]{pelletier2024inter}
\bibinfo{author}{Pelletier, F.}, \bibinfo{author}{Cardille, J.A.},
  \bibinfo{author}{Wulder, M.A.}, \bibinfo{author}{White, J.C.},
  \bibinfo{author}{Hermosilla, T.}, \bibinfo{year}{2024}.
\newblock \bibinfo{title}{Inter-and intra-year forest change detection and
  monitoring of aboveground biomass dynamics using sentinel-2 and landsat}.
\newblock \bibinfo{journal}{Remote Sensing of Environment}
  \bibinfo{volume}{301}, \bibinfo{pages}{113931}.
\bibitem[{Peng et~al.(2025)Peng, Liu, Zhang, Guan, Li and
  Bruzzone}]{peng2025deep}
\bibinfo{author}{Peng, D.}, \bibinfo{author}{Liu, X.}, \bibinfo{author}{Zhang,
  Y.}, \bibinfo{author}{Guan, H.}, \bibinfo{author}{Li, Y.},
  \bibinfo{author}{Bruzzone, L.}, \bibinfo{year}{2025}.
\newblock \bibinfo{title}{Deep learning change detection techniques for optical
  remote sensing imagery: Status, perspectives and challenges}.
\newblock \bibinfo{journal}{International Journal of Applied Earth Observation
  and Geoinformation} \bibinfo{volume}{136}, \bibinfo{pages}{104282}.
\bibitem[{Ronneberger et~al.(2015)Ronneberger, Fischer and
  Brox}]{ronneberger2015u}
\bibinfo{author}{Ronneberger, O.}, \bibinfo{author}{Fischer, P.},
  \bibinfo{author}{Brox, T.}, \bibinfo{year}{2015}.
\newblock \bibinfo{title}{U-net: Convolutional networks for biomedical image
  segmentation}, in: \bibinfo{booktitle}{International Conference on Medical
  image computing and computer-assisted intervention},
  \bibinfo{organization}{Springer}. pp. \bibinfo{pages}{234--241}.
\bibitem[{Saleh et~al.(2024)Saleh, Weng, Holail, Hao and Xia}]{saleh2024dam}
\bibinfo{author}{Saleh, T.}, \bibinfo{author}{Weng, X.},
  \bibinfo{author}{Holail, S.}, \bibinfo{author}{Hao, C.},
  \bibinfo{author}{Xia, G.S.}, \bibinfo{year}{2024}.
\newblock \bibinfo{title}{Dam-net: Flood detection from sar imagery using
  differential attention metric-based vision transformers}.
\newblock \bibinfo{journal}{ISPRS Journal of Photogrammetry and Remote Sensing}
  \bibinfo{volume}{212}, \bibinfo{pages}{440--453}.
\bibitem[{Sandler et~al.(2018)Sandler, Howard, Zhu, Zhmoginov and
  Chen}]{sandler2018mobilenetv2}
\bibinfo{author}{Sandler, M.}, \bibinfo{author}{Howard, A.},
  \bibinfo{author}{Zhu, M.}, \bibinfo{author}{Zhmoginov, A.},
  \bibinfo{author}{Chen, L.C.}, \bibinfo{year}{2018}.
\newblock \bibinfo{title}{Mobilenetv2: Inverted residuals and linear
  bottlenecks}, in: \bibinfo{booktitle}{Proceedings of the IEEE conference on
  computer vision and pattern recognition}, pp. \bibinfo{pages}{4510--4520}.
\bibitem[{Saputra et~al.(2025)Saputra, Bhaswara, Nasution, Ern, Husna, Witra,
  Feliren, Owen, Kemp and Lechner}]{saputra2025multi}
\bibinfo{author}{Saputra, M.R.U.}, \bibinfo{author}{Bhaswara, I.D.},
  \bibinfo{author}{Nasution, B.I.}, \bibinfo{author}{Ern, M.A.L.},
  \bibinfo{author}{Husna, N.L.R.}, \bibinfo{author}{Witra, T.},
  \bibinfo{author}{Feliren, V.}, \bibinfo{author}{Owen, J.R.},
  \bibinfo{author}{Kemp, D.}, \bibinfo{author}{Lechner, A.M.},
  \bibinfo{year}{2025}.
\newblock \bibinfo{title}{Multi-modal deep learning approaches to semantic
  segmentation of mining footprints with multispectral satellite imagery}.
\newblock \bibinfo{journal}{Remote Sensing of Environment}
  \bibinfo{volume}{318}, \bibinfo{pages}{114584}.
\bibitem[{Sengupta(2021)}]{sengupta2021environmental}
\bibinfo{author}{Sengupta, M.}, \bibinfo{year}{2021}.
\newblock \bibinfo{title}{Environmental impacts of mining: monitoring,
  restoration, and control}.
\newblock \bibinfo{publisher}{CRC Press}.
\bibitem[{Siqueira-Gay et~al.(2020)Siqueira-Gay, Sonter and
  S{\'a}nchez}]{siqueira2020exploring}
\bibinfo{author}{Siqueira-Gay, J.}, \bibinfo{author}{Sonter, L.J.},
  \bibinfo{author}{S{\'a}nchez, L.E.}, \bibinfo{year}{2020}.
\newblock \bibinfo{title}{Exploring potential impacts of mining on forest loss
  and fragmentation within a biodiverse region of brazil's northeastern
  amazon}.
\newblock \bibinfo{journal}{Resources Policy} \bibinfo{volume}{67},
  \bibinfo{pages}{101662}.
\bibitem[{Sonter et~al.(2017)Sonter, Herrera, Barrett, Galford, Moran and
  Soares-Filho}]{sonter2017mining}
\bibinfo{author}{Sonter, L.J.}, \bibinfo{author}{Herrera, D.},
  \bibinfo{author}{Barrett, D.J.}, \bibinfo{author}{Galford, G.L.},
  \bibinfo{author}{Moran, C.J.}, \bibinfo{author}{Soares-Filho, B.S.},
  \bibinfo{year}{2017}.
\newblock \bibinfo{title}{Mining drives extensive deforestation in the
  brazilian amazon}.
\newblock \bibinfo{journal}{Nature communications} \bibinfo{volume}{8},
  \bibinfo{pages}{1013}.
\bibitem[{Sonter et~al.(2014)Sonter, Moran, Barrett and
  Soares-Filho}]{sonter2014processes}
\bibinfo{author}{Sonter, L.J.}, \bibinfo{author}{Moran, C.J.},
  \bibinfo{author}{Barrett, D.J.}, \bibinfo{author}{Soares-Filho, B.S.},
  \bibinfo{year}{2014}.
\newblock \bibinfo{title}{Processes of land use change in mining regions}.
\newblock \bibinfo{journal}{Journal of Cleaner Production}
  \bibinfo{volume}{84}, \bibinfo{pages}{494--501}.
\bibitem[{Sun et~al.(2024a)Sun, Zhou, Jia, Shao, Liu, Tao and
  Dai}]{sun2024impacts}
\bibinfo{author}{Sun, X.}, \bibinfo{author}{Zhou, Y.}, \bibinfo{author}{Jia,
  S.}, \bibinfo{author}{Shao, H.}, \bibinfo{author}{Liu, M.},
  \bibinfo{author}{Tao, S.}, \bibinfo{author}{Dai, X.}, \bibinfo{year}{2024}a.
\newblock \bibinfo{title}{Impacts of mining on vegetation phenology and
  sensitivity assessment of spectral vegetation indices to mining activities in
  arid/semi-arid areas}.
\newblock \bibinfo{journal}{Journal of Environmental Management}
  \bibinfo{volume}{356}, \bibinfo{pages}{120678}.
\bibitem[{Sun et~al.(2024b)Sun, Zhong, Wang and Zhang}]{sun2024identifying}
\bibinfo{author}{Sun, Z.}, \bibinfo{author}{Zhong, Y.}, \bibinfo{author}{Wang,
  X.}, \bibinfo{author}{Zhang, L.}, \bibinfo{year}{2024}b.
\newblock \bibinfo{title}{Identifying cropland non-agriculturalization with
  high representational consistency from bi-temporal high-resolution remote
  sensing images: From benchmark datasets to real-world application}.
\newblock \bibinfo{journal}{ISPRS Journal of Photogrammetry and Remote Sensing}
  \bibinfo{volume}{212}, \bibinfo{pages}{454--474}.
\bibitem[{Tang and Werner(2023)}]{tang2023global}
\bibinfo{author}{Tang, L.}, \bibinfo{author}{Werner, T.T.},
  \bibinfo{year}{2023}.
\newblock \bibinfo{title}{Global mining footprint mapped from high-resolution
  satellite imagery}.
\newblock \bibinfo{journal}{Communications Earth \& Environment}
  \bibinfo{volume}{4}, \bibinfo{pages}{134}.
\bibitem[{Treml et~al.(2016)Treml, Arjona-Medina, Unterthiner, Durgesh,
  Friedmann, Schuberth, Mayr, Heusel, Hofmarcher, Widrich
  et~al.}]{treml2016speeding}
\bibinfo{author}{Treml, M.}, \bibinfo{author}{Arjona-Medina, J.},
  \bibinfo{author}{Unterthiner, T.}, \bibinfo{author}{Durgesh, R.},
  \bibinfo{author}{Friedmann, F.}, \bibinfo{author}{Schuberth, P.},
  \bibinfo{author}{Mayr, A.}, \bibinfo{author}{Heusel, M.},
  \bibinfo{author}{Hofmarcher, M.}, \bibinfo{author}{Widrich, M.}, et~al.,
  \bibinfo{year}{2016}.
\newblock \bibinfo{title}{Speeding up semantic segmentation for autonomous
  driving} .
\bibitem[{Wang et~al.(2024)Wang, Qin, Zhang, He and Cohen}]{wang2024mapping}
\bibinfo{author}{Wang, Y.}, \bibinfo{author}{Qin, K.}, \bibinfo{author}{Zhang,
  Z.}, \bibinfo{author}{He, Q.}, \bibinfo{author}{Cohen, J.},
  \bibinfo{year}{2024}.
\newblock \bibinfo{title}{Mapping open-pit mining area in complex mining and
  mixed land cover zone using landsat imagery}.
\newblock \bibinfo{journal}{International Journal of Applied Earth Observation
  and Geoinformation} \bibinfo{volume}{129}, \bibinfo{pages}{103782}.
\bibitem[{Wang et~al.(2023)Wang, Chen, Zhu, Jia and Plaza}]{wang2023rseife}
\bibinfo{author}{Wang, Z.}, \bibinfo{author}{Chen, T.}, \bibinfo{author}{Zhu,
  D.}, \bibinfo{author}{Jia, K.}, \bibinfo{author}{Plaza, A.},
  \bibinfo{year}{2023}.
\newblock \bibinfo{title}{Rseife: A new remote sensing ecological index for
  simulating the land surface eco-environment}.
\newblock \bibinfo{journal}{Journal of Environmental Management}
  \bibinfo{volume}{326}, \bibinfo{pages}{116851}.
\bibitem[{Werner et~al.(2020)Werner, Mudd, Schipper, Huijbregts, Taneja and
  Northey}]{werner2020global}
\bibinfo{author}{Werner, T.T.}, \bibinfo{author}{Mudd, G.M.},
  \bibinfo{author}{Schipper, A.M.}, \bibinfo{author}{Huijbregts, M.A.},
  \bibinfo{author}{Taneja, L.}, \bibinfo{author}{Northey, S.A.},
  \bibinfo{year}{2020}.
\newblock \bibinfo{title}{Global-scale remote sensing of mine areas and
  analysis of factors explaining their extent}.
\newblock \bibinfo{journal}{Global Environmental Change} \bibinfo{volume}{60},
  \bibinfo{pages}{102007}.
\bibitem[{Wolf et~al.(2020)Wolf, Debut, Sanh, Chaumond, Delangue, Moi, Cistac,
  Rault, Louf, Funtowicz, Davison, Shleifer, von Platen, Ma, Jernite, Plu, Xu,
  Scao, Gugger, Drame, Lhoest and Rush}]{wolf-etal-2020-transformers}
\bibinfo{author}{Wolf, T.}, \bibinfo{author}{Debut, L.}, \bibinfo{author}{Sanh,
  V.}, \bibinfo{author}{Chaumond, J.}, \bibinfo{author}{Delangue, C.},
  \bibinfo{author}{Moi, A.}, \bibinfo{author}{Cistac, P.},
  \bibinfo{author}{Rault, T.}, \bibinfo{author}{Louf, R.},
  \bibinfo{author}{Funtowicz, M.}, \bibinfo{author}{Davison, J.},
  \bibinfo{author}{Shleifer, S.}, \bibinfo{author}{von Platen, P.},
  \bibinfo{author}{Ma, C.}, \bibinfo{author}{Jernite, Y.},
  \bibinfo{author}{Plu, J.}, \bibinfo{author}{Xu, C.}, \bibinfo{author}{Scao,
  T.L.}, \bibinfo{author}{Gugger, S.}, \bibinfo{author}{Drame, M.},
  \bibinfo{author}{Lhoest, Q.}, \bibinfo{author}{Rush, A.M.},
  \bibinfo{year}{2020}.
\newblock \bibinfo{title}{Transformers: State-of-the-art natural language
  processing}, in: \bibinfo{booktitle}{Proceedings of the 2020 Conference on
  Empirical Methods in Natural Language Processing: System Demonstrations},
  \bibinfo{publisher}{Association for Computational Linguistics},
  \bibinfo{address}{Online}. pp. \bibinfo{pages}{38--45}.
\newblock \URLprefix \url{https://www.aclweb.org/anthology/2020.emnlp-demos.6}.
\bibitem[{Wu et~al.(2017)Wu, Du, Cui and Zhang}]{wu2017post}
\bibinfo{author}{Wu, C.}, \bibinfo{author}{Du, B.}, \bibinfo{author}{Cui, X.},
  \bibinfo{author}{Zhang, L.}, \bibinfo{year}{2017}.
\newblock \bibinfo{title}{A post-classification change detection method based
  on iterative slow feature analysis and bayesian soft fusion}.
\newblock \bibinfo{journal}{Remote Sensing of Environment}
  \bibinfo{volume}{199}, \bibinfo{pages}{241--255}.
\bibitem[{Wu et~al.(2024)Wu, Zhang, Du, Chen, Wang and Zhong}]{wu2024unet}
\bibinfo{author}{Wu, C.}, \bibinfo{author}{Zhang, L.}, \bibinfo{author}{Du,
  B.}, \bibinfo{author}{Chen, H.}, \bibinfo{author}{Wang, J.},
  \bibinfo{author}{Zhong, H.}, \bibinfo{year}{2024}.
\newblock \bibinfo{title}{Unet-like remote sensing change detection: A review
  of current models and research directions}.
\newblock \bibinfo{journal}{IEEE Geoscience and Remote Sensing Magazine} .
\bibitem[{Xiao et~al.(2018)Xiao, Liu, Zhou, Jiang and Sun}]{xiao2018unified}
\bibinfo{author}{Xiao, T.}, \bibinfo{author}{Liu, Y.}, \bibinfo{author}{Zhou,
  B.}, \bibinfo{author}{Jiang, Y.}, \bibinfo{author}{Sun, J.},
  \bibinfo{year}{2018}.
\newblock \bibinfo{title}{Unified perceptual parsing for scene understanding},
  in: \bibinfo{booktitle}{Proceedings of the European conference on computer
  vision (ECCV)}, pp. \bibinfo{pages}{418--434}.
\bibitem[{Xie et~al.(2021)Xie, Wang, Yu, Anandkumar, Alvarez and
  Luo}]{xie2021segformer}
\bibinfo{author}{Xie, E.}, \bibinfo{author}{Wang, W.}, \bibinfo{author}{Yu,
  Z.}, \bibinfo{author}{Anandkumar, A.}, \bibinfo{author}{Alvarez, J.M.},
  \bibinfo{author}{Luo, P.}, \bibinfo{year}{2021}.
\newblock \bibinfo{title}{Segformer: Simple and efficient design for semantic
  segmentation with transformers}.
\newblock \bibinfo{journal}{Advances in neural information processing systems}
  \bibinfo{volume}{34}, \bibinfo{pages}{12077--12090}.
\bibitem[{Xie et~al.(2020)Xie, Pan, Luan, Yang and Xi}]{xie2020semantic}
\bibinfo{author}{Xie, H.}, \bibinfo{author}{Pan, Y.}, \bibinfo{author}{Luan,
  J.}, \bibinfo{author}{Yang, X.}, \bibinfo{author}{Xi, Y.},
  \bibinfo{year}{2020}.
\newblock \bibinfo{title}{Semantic segmentation of open pit mining area based
  on remote sensing shallow features and deep learning}, in:
  \bibinfo{booktitle}{International conference on Big Data Analytics for
  Cyber-Physical-Systems}, \bibinfo{organization}{Springer}. pp.
  \bibinfo{pages}{52--59}.
\bibitem[{Xie et~al.(2025)Xie, Li, Jiang, Yang, Qiao, Yuan and
  Nie}]{xie2025integrated}
\bibinfo{author}{Xie, Z.}, \bibinfo{author}{Li, K.}, \bibinfo{author}{Jiang,
  J.}, \bibinfo{author}{Yang, J.}, \bibinfo{author}{Qiao, X.},
  \bibinfo{author}{Yuan, D.}, \bibinfo{author}{Nie, C.}, \bibinfo{year}{2025}.
\newblock \bibinfo{title}{An integrated neighborhood and scale information
  network for open-pit mine change detection in high-resolution remote sensing
  images}.
\newblock \bibinfo{journal}{Computers \& Geosciences} \bibinfo{volume}{196},
  \bibinfo{pages}{105880}.
\bibitem[{Yang et~al.(2018)Yang, Li, Zipper, Shen, Miao and
  Donovan}]{yang2018identification}
\bibinfo{author}{Yang, Z.}, \bibinfo{author}{Li, J.}, \bibinfo{author}{Zipper,
  C.E.}, \bibinfo{author}{Shen, Y.}, \bibinfo{author}{Miao, H.},
  \bibinfo{author}{Donovan, P.F.}, \bibinfo{year}{2018}.
\newblock \bibinfo{title}{Identification of the disturbance and trajectory
  types in mining areas using multitemporal remote sensing images}.
\newblock \bibinfo{journal}{Science of the total environment}
  \bibinfo{volume}{644}, \bibinfo{pages}{916--927}.
\bibitem[{Ye et~al.(2023)Ye, Wang, Zhou, Lei, Fan and Qin}]{ye2023adjacent}
\bibinfo{author}{Ye, Y.}, \bibinfo{author}{Wang, M.}, \bibinfo{author}{Zhou,
  L.}, \bibinfo{author}{Lei, G.}, \bibinfo{author}{Fan, J.},
  \bibinfo{author}{Qin, Y.}, \bibinfo{year}{2023}.
\newblock \bibinfo{title}{Adjacent-level feature cross-fusion with 3-d cnn for
  remote sensing image change detection}.
\newblock \bibinfo{journal}{IEEE Transactions on Geoscience and Remote Sensing}
  \bibinfo{volume}{61}, \bibinfo{pages}{1--14}.
\bibitem[{Yu et~al.(2018)Yu, Xu, Xue, Li, Cheng, Liu, Porwal, Holden, Yang and
  Gong}]{yu2018monitoring}
\bibinfo{author}{Yu, L.}, \bibinfo{author}{Xu, Y.}, \bibinfo{author}{Xue, Y.},
  \bibinfo{author}{Li, X.}, \bibinfo{author}{Cheng, Y.}, \bibinfo{author}{Liu,
  X.}, \bibinfo{author}{Porwal, A.}, \bibinfo{author}{Holden, E.J.},
  \bibinfo{author}{Yang, J.}, \bibinfo{author}{Gong, P.}, \bibinfo{year}{2018}.
\newblock \bibinfo{title}{Monitoring surface mining belts using multiple remote
  sensing datasets: A global perspective}.
\newblock \bibinfo{journal}{Ore Geology Reviews} \bibinfo{volume}{101},
  \bibinfo{pages}{675--687}.
\bibitem[{Yu et~al.(2024a)Yu, Zhang, Das, Zhu and Ghamisi}]{yu2024maskcd}
\bibinfo{author}{Yu, W.}, \bibinfo{author}{Zhang, X.}, \bibinfo{author}{Das,
  S.}, \bibinfo{author}{Zhu, X.X.}, \bibinfo{author}{Ghamisi, P.},
  \bibinfo{year}{2024}a.
\newblock \bibinfo{title}{Maskcd: A remote sensing change detection network
  based on mask classification}.
\newblock \bibinfo{journal}{IEEE Transactions on Geoscience and Remote Sensing}
  \bibinfo{volume}{62}, \bibinfo{pages}{1--16}.
\bibitem[{Yu et~al.(2024b)Yu, Zhang, Gloaguen, Zhu and
  Ghamisi}]{yu2024minenetcd}
\bibinfo{author}{Yu, W.}, \bibinfo{author}{Zhang, X.},
  \bibinfo{author}{Gloaguen, R.}, \bibinfo{author}{Zhu, X.X.},
  \bibinfo{author}{Ghamisi, P.}, \bibinfo{year}{2024}b.
\newblock \bibinfo{title}{Minenetcd: A benchmark for global mining change
  detection on remote sensing imagery}.
\newblock \bibinfo{journal}{IEEE Transactions on Geoscience and Remote Sensing}
  .
\bibitem[{Yuan et~al.(2021)Yuan, Li, Si, Ren, Yang, Gong, Xia, Tong and
  Tong}]{9553995}
\bibinfo{author}{Yuan, L.}, \bibinfo{author}{Li, Y.}, \bibinfo{author}{Si, Y.},
  \bibinfo{author}{Ren, J.}, \bibinfo{author}{Yang, Y.}, \bibinfo{author}{Gong,
  Y.}, \bibinfo{author}{Xia, Y.}, \bibinfo{author}{Tong, Z.},
  \bibinfo{author}{Tong, L.}, \bibinfo{year}{2021}.
\newblock \bibinfo{title}{Multi-objects change detection based on res-unet},
  in: \bibinfo{booktitle}{2021 IEEE International Geoscience and Remote Sensing
  Symposium IGARSS}, pp. \bibinfo{pages}{4364--4367}.
\newblock \DOIprefix\doi{10.1109/IGARSS47720.2021.9553995}.
\bibitem[{Zhang et~al.(2012)Zhang, Wu, Zhang, Jiao and
  Li}]{zhang2012application}
\bibinfo{author}{Zhang, B.}, \bibinfo{author}{Wu, D.}, \bibinfo{author}{Zhang,
  L.}, \bibinfo{author}{Jiao, Q.}, \bibinfo{author}{Li, Q.},
  \bibinfo{year}{2012}.
\newblock \bibinfo{title}{Application of hyperspectral remote sensing for
  environment monitoring in mining areas}.
\newblock \bibinfo{journal}{Environmental Earth Sciences} \bibinfo{volume}{65},
  \bibinfo{pages}{649--658}.
\bibitem[{Zhang et~al.(2023a)Zhang, Li, Li, Zhang, Ran, Du, Guo and
  Hou}]{zhang2023assessing}
\bibinfo{author}{Zhang, C.}, \bibinfo{author}{Li, F.}, \bibinfo{author}{Li,
  J.}, \bibinfo{author}{Zhang, K.}, \bibinfo{author}{Ran, W.},
  \bibinfo{author}{Du, M.}, \bibinfo{author}{Guo, J.}, \bibinfo{author}{Hou,
  G.}, \bibinfo{year}{2023}a.
\newblock \bibinfo{title}{Assessing the effect, attribution, and potential of
  vegetation restoration in open-pit coal mines’ dumping sites during
  2003--2020 utilizing remote sensing}.
\newblock \bibinfo{journal}{Ecological Indicators} \bibinfo{volume}{155},
  \bibinfo{pages}{111003}.
\bibitem[{Zhang et~al.(2021)Zhang, He, Li, Xiao, Song, Lu and
  Wu}]{zhang2021continuous}
\bibinfo{author}{Zhang, M.}, \bibinfo{author}{He, T.}, \bibinfo{author}{Li,
  G.}, \bibinfo{author}{Xiao, W.}, \bibinfo{author}{Song, H.},
  \bibinfo{author}{Lu, D.}, \bibinfo{author}{Wu, C.}, \bibinfo{year}{2021}.
\newblock \bibinfo{title}{Continuous detection of surface-mining footprint in
  copper mine using google earth engine}.
\newblock \bibinfo{journal}{Remote Sensing} \bibinfo{volume}{13},
  \bibinfo{pages}{4273}.
\bibitem[{Zhang et~al.(2023b)Zhang, Yu, Pun and Shi}]{zhang2023cross}
\bibinfo{author}{Zhang, X.}, \bibinfo{author}{Yu, W.}, \bibinfo{author}{Pun,
  M.O.}, \bibinfo{author}{Shi, W.}, \bibinfo{year}{2023}b.
\newblock \bibinfo{title}{Cross-domain landslide mapping from large-scale
  remote sensing images using prototype-guided domain-aware progressive
  representation learning}.
\newblock \bibinfo{journal}{ISPRS Journal of Photogrammetry and Remote Sensing}
  \bibinfo{volume}{197}, \bibinfo{pages}{1--17}.
\bibitem[{Zhao et~al.(2017)Zhao, Shi, Qi, Wang and Jia}]{zhao2017pyramid}
\bibinfo{author}{Zhao, H.}, \bibinfo{author}{Shi, J.}, \bibinfo{author}{Qi,
  X.}, \bibinfo{author}{Wang, X.}, \bibinfo{author}{Jia, J.},
  \bibinfo{year}{2017}.
\newblock \bibinfo{title}{Pyramid scene parsing network}, in:
  \bibinfo{booktitle}{Proceedings of the IEEE conference on computer vision and
  pattern recognition}, pp. \bibinfo{pages}{2881--2890}.
\bibitem[{Zheng et~al.(2022)Zheng, Zhong, Tian, Ma and
  Zhang}]{zheng2022changemask}
\bibinfo{author}{Zheng, Z.}, \bibinfo{author}{Zhong, Y.},
  \bibinfo{author}{Tian, S.}, \bibinfo{author}{Ma, A.}, \bibinfo{author}{Zhang,
  L.}, \bibinfo{year}{2022}.
\newblock \bibinfo{title}{Changemask: Deep multi-task
  encoder-transformer-decoder architecture for semantic change detection}.
\newblock \bibinfo{journal}{ISPRS Journal of Photogrammetry and Remote Sensing}
  \bibinfo{volume}{183}, \bibinfo{pages}{228--239}.

\end{thebibliography}
\end{document}